\documentclass[10pt,twocolumn,letterpaper]{article}

\usepackage{3dv}
\usepackage{times}
\usepackage{epsfig}
\usepackage{graphicx}
\usepackage{amsmath}
\usepackage{amssymb}

\usepackage{subcaption}
\usepackage{multirow}
\usepackage{makecell}
\usepackage{empheq}
\usepackage{wasysym}
\usepackage[ruled,lined,longend,linesnumbered,noend]{algorithm2e}
\usepackage[noend]{algpseudocode}
\usepackage{ifthen}
\usepackage{adjustbox}

\SetCommentSty{mycommfont}
\SetKwComment{tcp}{$\triangleright$}{}
\usepackage{stackengine}
\usepackage[bookmarks=false]{hyperref}
\graphicspath{{figures/}}
\def\myref#1{{\color{blue}{#1}}}
\def\myblue#1{{\color{red}{#1}}}
\usepackage{array}
\usepackage{color}
\usepackage{xcolor}
\usepackage{ifthen}
\newtheorem{proposition}{Proposition}

\newcommand{\accerror}[2]{{#1}$\pm${#2}}

\newcommand{\sunderline}[1]{\underline{\smash{#1}}}

\hypersetup{colorlinks=true, linkcolor=red, filecolor=magenta, urlcolor=blue}
\def\etl{\textit{et~al.}}
\def\eg{\textit{e.g.}}
\def\ie{\textit{i.e.}}

\threedvfinalcopy 

\def\threedvPaperID{114} 

\ifthreedvfinal\fi
\begin{document}

\newif\ifsupp
\newif\ifmain
\newif\ifarxiv

\suppfalse
\mainfalse
\arxivtrue

\ifsupp
    \title{RANP: Resource Aware Neuron Pruning at Initialization for 3D CNNs\\
    -- Supplementary Material --}
\else
    \title{RANP: Resource Aware Neuron Pruning at Initialization for 3D CNNs}
\fi

\author{Zhiwei Xu$^{1,3}$ \quad Thalaiyasingam Ajanthan$^1$ \quad Vibhav Vineet$^2$ \quad Richard Hartley$^1$\\
$^1$Australian National University and Australian Centre for Robotic Vision\\
$^2$Microsoft Research, Redmond, USA\\
$^3$Data61, CSIRO, Australia\\
{\tt\small \{zhiwei.xu,thalaiyasingam.ajanthan,richard.hartley\}@anu.edu.au}\\
{\tt\small vibhav.vineet@microsoft.com}
}

\maketitle

\ifsupp
    \appendix
    \setcounter{equation}{11}
    \setcounter{figure}{5}
    \setcounter{table}{5}
\ifarxiv \subsection*{Appendix} \fi

We first provide the pseudocode of our RANP algorithm, then discuss our selection of MPMG-sum as vanilla NP, and justify our reweighting scheme against orthogonal initialization with more ablation experiments.

\section{Pseudocode of RANP Procedures}
In Alg.~\ref{alg:pseudo}, we provide the pseudocode of the pruning procedures of RANP.
In Alg.~\ref{alg:search}, we used a simple half-space method to automatically search for the max neuron sparsity with network feasibility.
Note that this searching cannot guarantee a small accuracy loss but merely to decide the maximum pruning capability.
The relation between pruning capability and accuracy was studied in the experimental section in the main paper and Table~\ref{tb:neuron_importance_full}.

\textbf{Loss Function and Metrics.}
Due to the page limitation, we provide loss functions and metrics used in our experiments.
Standard cross-entropy function was used as the loss function for ShapeNet and UCF101.
For BraTS'18, the weighted function in \cite{cross_validation} is

\vspace{-1.5em}
\begin{equation}
\label{eq:dice_loss}
\begin{aligned}
L = L_{ce} + \alpha L_{dice}
  = L_{ce} + \alpha \frac{1}{C} \sum_{i=1}^C \frac{2 |\mathbf{P}_i \cap \mathbf{G}_i|}{|\mathbf{P}| + |\mathbf{G}|}\ ,
\end{aligned}
\end{equation}
where $\alpha=0.25$ is an empiric weight for dice loss, $\mathbf{P}$ is prediction, $\mathbf{G}$ is ground truth, and $C$ is the number of classes.
Meanwhile, ShapeNet accuracy was measured by mean IoU over each part of object category \cite{mean_iou} while IoU by $|\mathbf{P} \cap \mathbf{G} | / |\mathbf{P} \cup \mathbf{G}|$ was adopted for BraTS'18.
For UCF101 classification, top-1 and top-5 recall rates were used.

\SetInd{0.2em}{0.6em}
\begin{algorithm*}
\KwIn{Dataset $\mathcal{D}=\{ (\mathbf{x}_i, \mathbf{y}_i) \}^S_{i=1}$ with $B$ samples per batch, neuron sparsity $\kappa$, resource importance $\{\tau_l\}$, coefficient $\lambda > 0$, and parameter masks $\mathbf{c}=\{c^l_{uv}\}$, where layer $l \in \mathcal{K} = \{1, ..., K\}$, and neuron $u \in \mathcal{N}_l = \{1, ..., N_l\}$.}
\KwOut{Binary neuron masks $\hat{\mathbf{c}}=\{\hat{c}^l_u\}$.}
\caption{Pruning Procedures of RANP-[f$\vert$m].}
\label{alg:pseudo}
\For{\textup{batch $t \in \{1, ..., \lfloor S/B \rfloor \}$}}{
  $\mathcal{D}^t \gets \{ (\mathbf{x}_i, \mathbf{y}_i) \}^{tB}_{i=(t-1)B+1}$ \tcp*[f]{mini-batch} \\
  $g^l_{uv} \gets \partial L(\mathbf{c} \odot \mathbf{w}; \mathcal{D}^t) / \partial c^l_{uv}$ \tcp*[f]{parameter mask gradient, Eq.~\myblue{2}} \\
  $g^l_{uv} \gets |g^l_{uv}|$, for MPMG \tcp*[f]{parameter mask importance, Eq.~\myblue{6}} \\
  $\nabla c^l_{uv} \overset{+}{\gets} g^l_{uv}, \forall u \in \mathcal{N}_l, \forall v \in \mathcal{N}_{l-1}$ \tcp*[f]{gradient accumulation} \\
}
$\nabla c^l_{uv} \gets \nabla c^l_{uv} / \lfloor S/B \rfloor, \forall u \in \mathcal{N}_l, \forall v \in \mathcal{N}_{l-1}, \forall l \in \mathcal{K}$ \tcp*[f]{average on mini-batch} \\
$s^l_u \gets |\sum_{v=1}^{N_{l-1}} \nabla c^l_{uv}|, \forall u \in \mathcal{N}_l, \forall l \in \mathcal{K}$ \tcp*[f]{vanilla neuron importance, Eq.~\myblue{8}} \\
$\Bar{s}^l \gets \sum_{u \in \mathcal{N}_l} s^l_u / N_l, \forall l \in \mathcal{K} $ \tcp*[f]{mean neuron importance, Eq.~\myblue{9}} \\
$\tilde{s}^l_u \gets ( \max_{j \in \mathcal{K}} \Bar{s}^j / \Bar{s}^l) s^l_u, \forall u \in \mathcal{N}_l, \forall l \in \mathcal{K} $  \tcp*[f]{weighting, Eq.~\myblue{9}} \\
$\hat{s}^l_u \gets ( 1 + \lambda e^{-\tau_l} / \sum_{j \in \mathcal{K}} e^{-\tau_j} ) \tilde{s}^l_u, \forall u \in \mathcal{N}_l, \forall l \in \mathcal{K} $  \tcp*[f]{reweighting, Eq.~\myblue{10}} \\
$\{\ddot{s}_u\} \gets \text{SortDescending} \left( \{ \hat{s}^l_u \} \right), \forall u \in \mathcal{N}_l, \forall l \in \mathcal{K}$ \tcp*[f]{sorting in descending} \\

$\hat{c}^l_{u} \gets 1[\hat{s}^l_u - \ddot{s}_{\kappa} \geq 0], \forall u \in \mathcal{N}_l, \forall l \in \mathcal{K}$ \tcp*[f]{binary neuron mask, Eq.~\myblue{11}} \\
\end{algorithm*}

\SetInd{0.2em}{0.6em}
\begin{algorithm}[!t]
\KwIn{Dataset $
\mathcal{D}$, layerwise resource usage $\boldsymbol{\tau}$ w.r.t. FLOPs or memory, coefficient $\lambda>0$, lower and upper sparsity $\kappa_{min}$ and $\kappa_{max}$, threshold $\delta=1e-4$.
``feasible network" means not all neurons are removed in each layer.}
\KwOut{Max neuron sparsity $\kappa^{*}$.}
\caption{Auto-Search for Max Neuron Sparsity
.}
\label{alg:search}
Initilize $\kappa_{min} \gets 0, \kappa_{max} \gets 1$ \\
\While{\textup{($\kappa_{max} - \kappa_{min} > \delta$)}}{
  $\kappa = 0.5 \left( \kappa_{min} + \kappa_{max} \right)$ \\
  $y = \text{NeuronPruning}(\mathcal{D}, \boldsymbol{\tau}, \lambda, \kappa)$ \tcp*[f]{Alg.~\myblue{\ref{alg:pseudo}}}\\
  \If{$y == 0(\text{feasible network})$}{
    $\kappa_{min} \gets \kappa$
  }\Else{
    $\kappa_{max} \gets \kappa$
  }
}
$\kappa^{*} = \kappa$
\end{algorithm}

\section{Impacts of the Activation Function}

\begin{figure}[!h]
\begin{center}
  \begin{subfigure}[b]{0.22\textwidth}
  \centering
  \includegraphics[width=\textwidth]{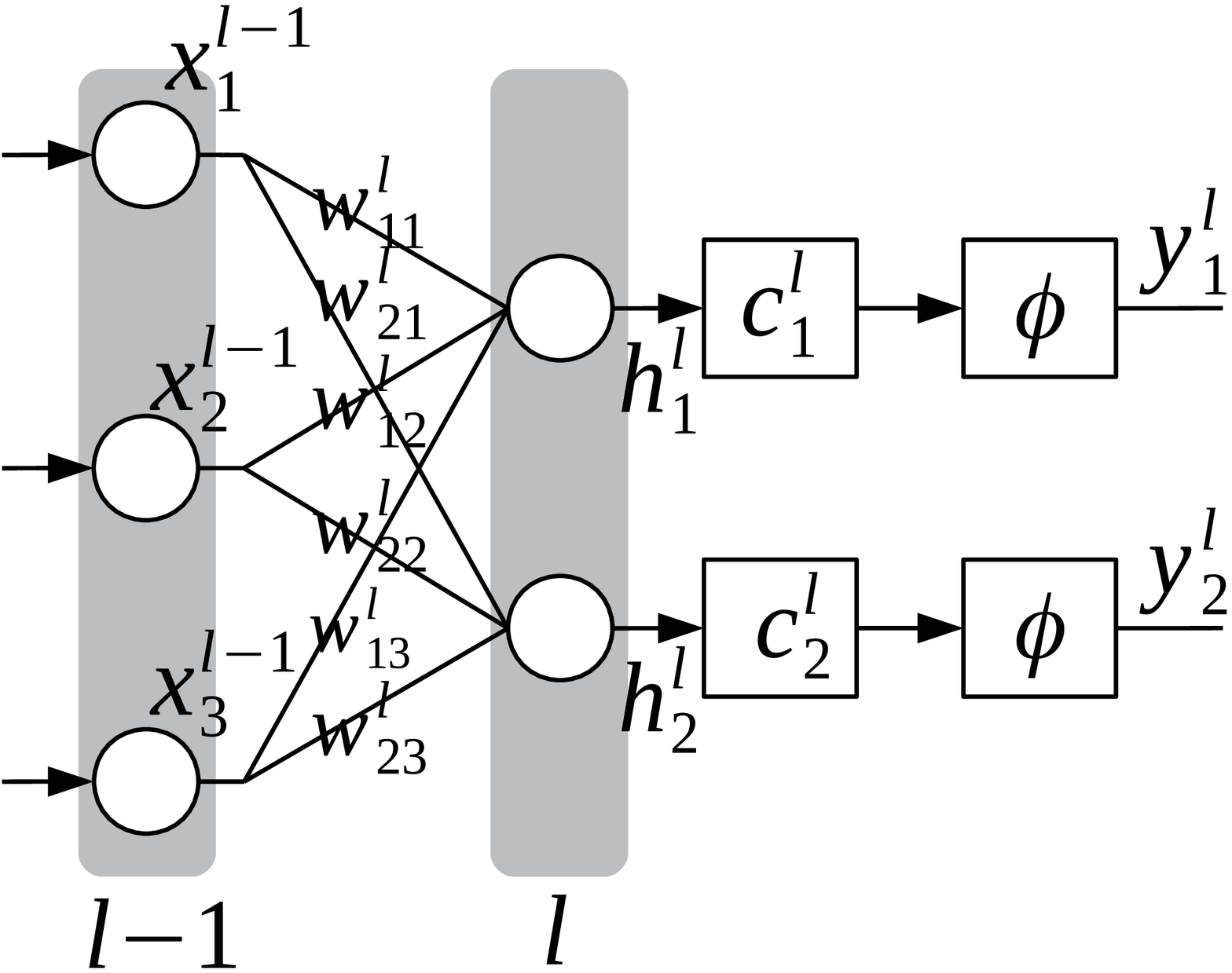}
  \caption{Pre-activations}
  \label{fig:pre_activation}
  \end{subfigure}
  ~
  \begin{subfigure}[b]{0.22\textwidth}
  \centering
  \includegraphics[width=\textwidth]{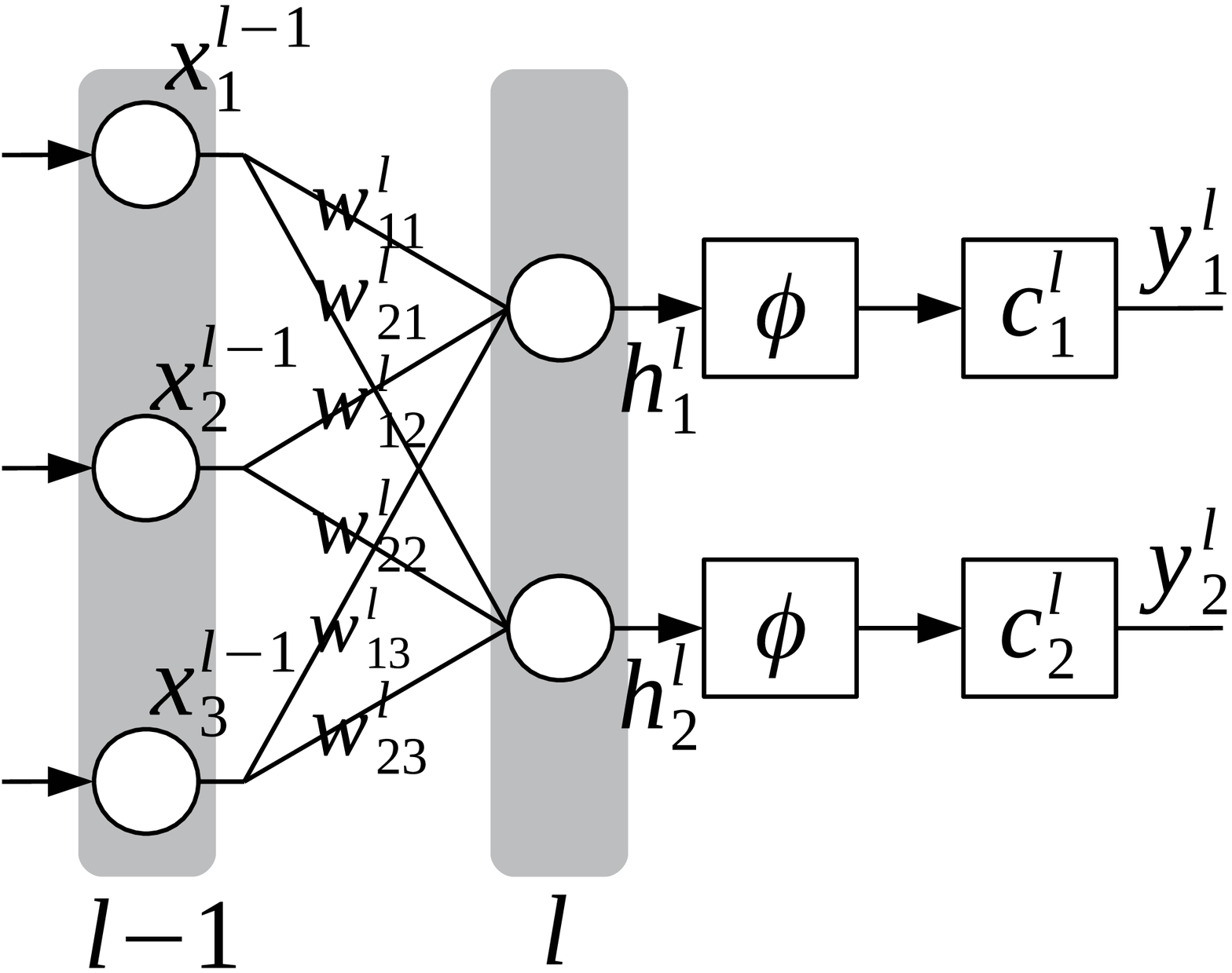}
  \caption{Post-activations}
  \label{fig:post_activation}
  \end{subfigure}
\caption{Pre-activations and post-activations, where $\mathbf{x}$ are layer inputs, $\mathbf{w}$ are weights, $\mathbf{c}$ are neuron masks, $\phi(\cdot)$ is an activation function, $\mathbf{h}$ are hidden values, and $\mathbf{y}$ are outputs.}
\end{center}
\vspace{-5mm}
\end{figure}

In the following, we first establish the relation between MPMG and MNMG for calculating neuron importance given a homogeneous activation function $\phi(\cdot)$ that includes but not limited to ReLU used in the 3D CNNs.
Then we analyze the impact of such an activation function on the calculation of neuron importance by derivating the mask gradients on post-activations and pre-activations illustrated in Figs.~\myblue{\ref{fig:post_activation}} and \myblue{\ref{fig:pre_activation}} respectively.

\begin{proposition}
For a network activation function $\phi(w)$: $\mathbb{R} \rightarrow \mathbb{R}$ being a homogeneous function of degree 1 satisfying $\phi(cw)=c\phi(w), \forall c \geq 0$, the neuron mask gradient equals the sum of parameter mask gradients of this neuron.
\end{proposition}

\noindent \textit{Proof:} Given a neuron mask $c_1$ before the activation function $\phi(\cdot)$ in Fig.~\myblue{\ref{fig:pre_activation}} and the output of the 1st neuron as $y^l_1$, we have

\begin{equation}
\label{eq:loss_1}
\begin{aligned}
y^l_1
  &= \phi(c^l_1 \odot h^l_1) \\
  &= \phi \left( c^l_1 \odot \left( x^{l-1}_1 w^l_{11} + x^{l-1}_2 w^l_{12} + x^{l-1}_3 w^l_{13} \right) \right) \\
  &= \phi \left( c^l_1 x^{l-1}_1 w^l_{11} + c^l_1 x^{l-1}_2 w^l_{12} + c^l_1 x^{l-1}_3 w^l_{13} \right)\ .
\end{aligned}
\end{equation}
The gradient of loss $L$ over the neuron mask $c^l_1$ is

\begin{equation}
\label{eq:neuron_mask_1}
\begin{aligned}
\frac{\partial L}{\partial c^l_1}
&=\frac{\partial L}{\partial y^l_1} \frac{\partial y^l_1}{\partial c^l_1} \\
&= \frac{\partial L}{\partial y^l_1} \left( x^{l-1}_1 w^l_{11} + x^{l-1}_2 w^l_{12} + x^{l-1}_3 w^l_{13} \right)\ .
\end{aligned}
\end{equation}
Meanwhile, if setting masks on weights of this neuron directly, we can obtain

\begin{equation}
\label{eq:loss_2}
\begin{aligned}
y^l_1
= \phi(c^l_{11} x^{l-1}_1 w^l_{11} + c^l_{12} x^{l-1}_2 w^l_{12} + c^l_{13} x^{l-1}_3 w^l_{13})\ ,
\end{aligned}
\end{equation}
then the gradient of weight mask, \eg, $c^l_{11}$, from loss is

\begin{equation}
\label{eq:weight_mask_1}
\begin{aligned}
\frac{\partial L}{c^l_{11}}
= \frac{\partial L}{\partial y^l_1} \frac{\partial y^l_1}{\partial c^l_{11}}
= \frac{\partial L}{\partial y^l_1} x^{l-1}_1 w^l_{11}\ .
\end{aligned}
\end{equation}
Similarly,

\begin{equation}
\label{eq:weight_mask_sum_1}
\begin{aligned}
&\frac{\partial L}{\partial c^l_{11}}
+ \frac{\partial L}{\partial c^l_{12}}
+ \frac{\partial L}{\partial c^l_{13}} \\
= &\frac{\partial L}{\partial y^l_1} \left( x^{l-1}_1 w^l_{11} + x^{l-1}_2 w^l_{12} + x^{l-1}_3 w^l_{13} \right)\ .
\end{aligned}
\end{equation}

Clearly, Eq.~\myblue{\ref{eq:neuron_mask_1}} equals Eq.~\myblue{\ref{eq:weight_mask_sum_1}}.
Hence, the neuron mask gradients can be calculated by parameter mask gradients.
To this end, the proof is done.

Furthermore, given such a homogeneous activation function in Prop.~\myblue{1}, the importance of a post-activation equals the importance of its pre-activation.
In more detail, for post-activations in Fig.~\myblue{\ref{fig:post_activation}}, output $y^l_1$ is

\begin{equation}
\begin{aligned}
y^l_1
&= c^l_1 \odot \phi(h^l_1) \\
&= c^l_1 \odot \phi \left(x^{l-1}_1 w^l_{11} + x^{l-1}_2 w^l_{12} + x^{l-1}_3 w^l_{13} \right)\ .
\end{aligned}
\end{equation}
Since the activation function satisfies $c\phi(w)=\phi(cw)$,

\vspace{-5mm}
\begin{equation}
\label{eq:loss_3}
\begin{aligned}
y^l_1
= \phi (c^l_1 x^{l-1}_1 w^l_{11} + c^l_1 x^{l-1}_2 w^l_{12} + c^l_1 x^{l-1}_3 w^l_{13})\ .
\end{aligned}
\end{equation}
The neuron importance determined by neuron mask $c^l_1$ is

\vspace{-3mm}
\begin{equation}
\label{eq:neuron_mask_2}
\begin{aligned}
\frac{\partial L}{\partial c^l_1}
&= \frac{\partial L}{\partial y^l_1} \frac{\partial y^l_1}{\partial c^l_1} \\
&= \frac{\partial L}{\partial y^l_1} \left( x^{l-1}_1 w^l_{11} + x^{l-1}_2 w^l_{12} + x^{l-1}_3 w^l_{13} \right)\ .
\end{aligned}
\end{equation}

Clearly, Eq.~\myblue{\ref{eq:neuron_mask_2}} equals Eq.~\myblue{\ref{eq:neuron_mask_1}}.
Now, the importance of pre-activations and post-activations is the same given such a homogeneous activation function.

\section{Resource Aware Reweighting Scheme}
As described in Sec.~\myblue{4.2} in the main paper, the reweighting of RANP is conducted by first balancing the layer-wise distribution of neuron importance and then adopting resource importance $\tau_{l}$ for layer $l \in \mathcal{K}$ to further reduce resources.
Since FLOPs and memory are the main resources of 3D CNNs, $\tau_l$ is defined by FLOPs or memory as follows.

Generally, given input dimension of the $l$th layer $(x_{\text{in}},x_h,x_w,x_d)$\footnote{The dimension order follows that of PyTorch.}, neuron dimension $(f_{\text{out}},f_{\text{in}},f_h,f_w,f_d)$, and output dimension $(y_{\text{in}},y_h,y_w,y_d)$ with $x_{\text{in}}=f_{\text{in}}$ and $f_{\text{out}}=y_{\text{in}}$, the resource importance in terms of FLOPs or memory is defined by

\vspace{-5mm}
\begin{subequations}
\begin{align}
\text{FLOPs:} \enskip \tau_l
= &\left[ \left( f_h f_w f_d + f_h f_w f_d - 1 \right) f_{\text{in}}\right. \nonumber \\
&\left. + f_{\text{in}} - 1 + 1\vert_{\text{bias}} \right] y_{\text{in}} y_h y_w y_d \nonumber \\
= &\left( 2 f_h f_w f_d f_{\text{in}} - 1 + 1\vert_{\text{bias}} \right) y_{\text{in}} y_h y_w y_d, \\
\text{Memory:} \enskip \tau_l &= y_{\text{in}} y_h y_w y_d,
\end{align}
\end{subequations}
where $(f_h f_w f_d)$ is the number of operations of multiplications of filter\footnote{Here, we refer a 3D filter with dimension $(f_h, f_w, f_d)$. } and layer input, $(f_h f_w f_d - 1)$ is for additions of values from the multiplications, $(f_{\text{in}})$ is for multiplications over all $f_{\text{in}}$ filters, $(f_{\text{in}} - 1)$ is for additions of values from all these multiplications, $(1\vert_{\text{bias}})$ is for an addition when the neuron has a bias, and $(y_{\text{in}} y_h y_w y_d)$ is for all elements of the layer output.

\section{More Ablation Study}

In this section, we add more experimental results for the analysis of selecting MPMG-sum as vanilla NP, Glorot initialization for network initialization compared with orthogonal initialization \cite{signal_propagation} to handle the imbalanced layer-wise distribution of neuron importance, and visualization of neuron distribution by RANP for BraTS'18 in addition to that for ShapeNet in the main paper.

Figures in this sections are for 3D-UNets on ShapeNet and BraTS'18 because 3D-UNets used in our experiments typically clarify the neuron imbalance and memory issues and are clear for illustration with a limited number of layers, \ie, 15 layers, while MobileNetV2 and I3D have more than 55 layers but many are not typical 3D convolutional layers with 3$^3$ kernel size filters.

\subsection{MPMG-sum as Vanilla Neuron Pruning} \label{sec:generalization}

\begin{table*}[t]
\centering
\caption{More results of vanilla NP in addition to Table~\myblue{1} in the main paper.
\textbf{Main resource consumption} (GFLOPs and memory) are considered but not parameters whose resource consumption is much smaller than memory.
Among the neuron pruning methods, we marked bold \textbf{the best} and underlined \sunderline{the second best}.
Overall, we selected MPMG-sum as vanilla NP and the corresponding neuron sparsity for large resource reductions with small accuracy loss.}
\label{tb:neuron_importance_full}
\resizebox{\textwidth}{!}{\begin{tabular}{l|llrrrrcccccc}
  \hline
  \multicolumn{1}{c}{Dataset}
  & \multicolumn{1}{c}{Model}
  & \multicolumn{1}{c}{Manner}
  & \multicolumn{1}{c}{Sparsity(\%)}
  & \multicolumn{1}{c}{Param(MB)}
  & \multicolumn{1}{c}{GFLOPs}
  & \multicolumn{1}{c}{Memory(MB)}
  & \multicolumn{6}{c}{Metrics(\%)} \\
  \hline
  & & & & & & & \multicolumn{6}{c}{mIoU} \\
  \multirow{7}{*}{ShapeNet}
  & \multirow{7}{*}{3D-UNet}
  & Full\cite{3dunet} & 0
  & 62.26
  & 237.85
  & 997.00 & \multicolumn{6}{c}{\accerror{83.79}{0.21}} \\
  &&MPMG-mean & 68.10
  & 5.08
  & 110.14
  & 819.97 & \multicolumn{6}{c}{\accerror{83.33}{0.18}} \\
  &&MPMG-max & 70.24
  & 4.54
  & 107.38
  & 809.88 & \multicolumn{6}{c}{\textbf{\accerror{83.79}{0.10}}} \\
  &&MPMG-sum & 78.24
  & 2.54
  & \textbf{55.69}
  & \textbf{557.32} & \multicolumn{6}{c}{\accerror{83.26}{0.14}} \\
  &&MNMG-mean & 63.03
  & 4.23
  & 112.95
  & 834.98 & \multicolumn{6}{c}{\accerror{83.46}{0.13}} \\
  &&MNMG-max & 73.93
  & 3.67
  & 103.57
  & 796.44 & \multicolumn{6}{c}{\accerror{83.51}{0.08}} \\
  &&MNMG-sum & 66.93
  & 4.29
  & \sunderline{100.34}
  & \sunderline{783.14} & \multicolumn{6}{c}{\sunderline{\accerror{83.65}{0.02}}} \\
  \hline
  & & & & & & & \multicolumn{2}{c}{ET}
  & \multicolumn{2}{c}{TC}
  & \multicolumn{2}{c}{WT} \\
  \multirow{7}{*}{BraTS'18}
  & \multirow{7}{*}{3D-UNet}
  & Full\cite{3dunet} & 0
  & 15.57
  & 478.13
  & 3628.00
  & \multicolumn{2}{c}{72.96$\pm$0.60}
  & \multicolumn{2}{c}{73.51$\pm$1.54}
  & \multicolumn{2}{c}{86.79$\pm$0.35} \\
  && MPMG-mean & 65.64
  & 1.48
  & 226.86
  & 3038.27
  & \multicolumn{2}{c}{\sunderline{73.51$\pm$0.82}}
  & \multicolumn{2}{c}{\sunderline{73.28$\pm$1.14}}
  & \multicolumn{2}{c}{\sunderline{87.15$\pm$0.43}} \\
  && MPMG-max & 75.78
  & 0.83
  & 189.43
  & 2812.53
  & \multicolumn{2}{c}{\textbf{73.67$\pm$0.98}}
  & \multicolumn{2}{c}{72.73$\pm$1.70}
  & \multicolumn{2}{c}{86.44$\pm$0.71} \\
  && MPMG-sum & 78.17
  & 0.55
  & \sunderline{104.50}
  & \sunderline{1936.44}
  & \multicolumn{2}{c}{71.94$\pm$1.68}
  & \multicolumn{2}{c}{69.39$\pm$2.29}
  & \multicolumn{2}{c}{84.68$\pm$0.78} \\
  && MNMG-mean & 63.85
  & 1.08
  & 176.76
  & 2790.64
  & \multicolumn{2}{c}{73.35$\pm$0.70}
  & \multicolumn{2}{c}{\textbf{73.38$\pm$0.94}}
  & \multicolumn{2}{c}{\textbf{87.21$\pm$0.38}} \\
  &&MNMG-max & 80.05
  & 0.59
  & 169.99
  & 2676.05
  & \multicolumn{2}{c}{72.52$\pm$1.91}
  & \multicolumn{2}{c}{72.40$\pm$1.74}
  & \multicolumn{2}{c}{84.63$\pm$0.60} \\
  &&MNMG-sum & 81.32
  & 0.35
  & \textbf{73.50}
  & \textbf{1933.20}
  & \multicolumn{2}{c}{64.48$\pm$1.10}
  & \multicolumn{2}{c}{68.47$\pm$1.59}
  & \multicolumn{2}{c}{80.71$\pm$1.07} \\
  \hline
  &&&&&&
  & \multicolumn{3}{c}{\hspace{10mm}Top-1}
  & \multicolumn{3}{c}{\hspace{-12mm}Top-5} \\
  \multirow{14}{*}{UCF101}
  & \multirow{7}{*}{MobileNetV2}
  & Full\cite{mobilenetv2} & 0
  & 9.47
  & 0.58
  & 157.47
  & \multicolumn{3}{c}{\hspace{10mm}
  \accerror{47.08}{0.72}}
  & \multicolumn{3}{c}{\hspace{-12mm}
  \accerror{76.68}{0.50}} \\
  && MPMG-mean & 26.31
  & 4.39
  & 0.55
  & 156.00
  & \multicolumn{3}{c}{\hspace{10mm}
  \accerror{2.98}{0.14}}\footnotemark & \multicolumn{3}{c}{\hspace{-12mm}
  \accerror{14.04}{0.14}} \\
  && MPMG-max & 29.48
  & 3.96
  & 0.54
  & 155.38
  & \multicolumn{3}{c}{\hspace{10mm}
  \accerror{3.49}{0.12}} & \multicolumn{3}{c}{\hspace{-12mm}
  \accerror{13.64}{0.10}} \\
  && MPMG-sum & 33.15
  & 6.35
  & 0.55
  & 155.17
  & \multicolumn{3}{c}{\hspace{10mm}
  \textbf{\accerror{46.32}{0.79}}}
  & \multicolumn{3}{c}{\hspace{-12mm}
  \textbf{\accerror{75.42}{0.60}}} \\
  && MNMG-mean & 38.91
  & 2.79
  & \sunderline{0.50}
  & \sunderline{147.69}
  & \multicolumn{3}{c}{\hspace{10mm}
  \sunderline{\accerror{29.13}{0.92}}} & \multicolumn{3}{c}{\hspace{-12mm}
  \sunderline{\accerror{62.93}{1.37}}} \\
  && MNMG-max & 50.33
  & 2.59
  & 0.53
  & 153.45
  & \multicolumn{3}{c}{\hspace{10mm}
  \accerror{2.84}{0.06}} & \multicolumn{3}{c}{\hspace{-12mm}
  \accerror{13.40}{0.23}} \\
  && MNMG-sum & 39.89
  & 4.66
  & \textbf{0.43}
  & \textbf{120.01}
  & \multicolumn{3}{c}{\hspace{10mm}
  \accerror{1.03}{0.00}} & \multicolumn{3}{c}{\hspace{-12mm}
  \accerror{5.76}{0.00}} \\
  \cline{2-13}
  & \multirow{7}{*}{I3D}
  & Full\cite{i3d} & 0
  & 47.27
  & 27.88
  & 201.28
  & \multicolumn{3}{c}{\hspace{10mm}
  \accerror{51.58}{1.86}}
  & \multicolumn{3}{c}{\hspace{-12mm}
  \accerror{77.35}{0.63}} \\
  && MPMG-mean & 16.47
  & 31.57
  & 26.50
  & 196.51
  & \multicolumn{3}{c}{\hspace{10mm}
  \sunderline{\accerror{51.88}{2.00}}} & \multicolumn{3}{c}{\hspace{-12mm}
  \accerror{77.98}{1.46}} \\
  && MPMG-max & 19.83
  & 30.06
  & 26.31 
  & 195.62
  & \multicolumn{3}{c}{\hspace{10mm}
  \textbf{\accerror{52.44}{1.25}}} & \multicolumn{3}{c}{\hspace{-12mm}
  \textbf{\accerror{78.08}{1.27}}} \\
  && MPMG-sum & 25.32
  & 29.93
  & 25.76
  & 192.42
  & \multicolumn{3}{c}{\hspace{10mm}
  \accerror{51.57}{1.46}}
  & \multicolumn{3}{c}{\hspace{-12mm}
  \sunderline{\accerror{78.07}{1.34}}} \\
  && MNMG-mean & 35.36
  & 16.69
  & \textbf{15.37}
  & \textbf{124.85}
  & \multicolumn{3}{c}{\hspace{10mm}
  \accerror{49.26}{0.96}} & \multicolumn{3}{c}{\hspace{-12mm}
  \accerror{75.70}{1.49}} \\
  && MNMG-max & 40.27
  & 17.86
  & 23.73
  & 184.77
  & \multicolumn{3}{c}{\hspace{10mm}
  \accerror{44.90}{1.19}} & \multicolumn{3}{c}{\hspace{-12mm}
  \accerror{74.43}{1.26}} \\
  && MNMG-sum & 32.87
  & 20.00
  & \sunderline{16.03}
  & \sunderline{125.17}
  & \multicolumn{3}{c}{\hspace{10mm}
  \accerror{46.90}{1.26}} & \multicolumn{3}{c}{\hspace{-12mm}
  \accerror{74.02}{1.25}} \\
  \hline
\end{tabular}}
\vspace{-3mm}
\end{table*}
\footnotetext[3]{For MobileNetV2 pruned by MPMG-mean, MPMG-max, MNMG-max, and MNMG-sum, the accuracy is very low because 1) the neuron sparsity here is the extreme (largest) value, a larger one will make network infeasible by removing whole layer(s) and 2) the distribution of neuron importance is rather imbalanced possibly caused by the high mixture of 1$^3$ kernels and 3$^3$ in MobileNetV2.

In the pruned networks, we observe that, for MPMG-mean, MPMG-max, and MNMG-max, the last convolutional layer has only 1 neuron retained; for MNMG-sum, 2 convolutional layers have only 1 neuron retained.
Note that, this imbalance issue can be greatly alleviated by the reweighting of our RANP, while we select MPMG-sum as vanilla NP merely according to the results in Table~\ref{tb:neuron_importance_full}.}

In Sec.~\myblue{5.2} in the main paper, we select MPMG-sum as vanilla neuron pruning for the trade-off between computational resources and accuracy.
To give a comprehensive study of this selection, we demonstrate detailed results of mean, max, and sum operations of MPMG and MNMG in Table~\myblue{\ref{tb:neuron_importance_full}}.
Note that we relax the sum operation in Eq.~\myblue{8} in the main paper to mean, max, and sum.

In Table~\myblue{\ref{tb:neuron_importance_full}}, we aim at obtaining the maximum neuron sparsity due to the target of reducing the computational resources at an extreme sparsity level with minimal accuracy loss.
Vividly, for \textit{ShapeNet}, MPMG-sum achieves the largest maximum neuron sparsity 78.24\% among all with only ${\sim}$0.53\% accuracy loss.
Differently, for \textit{BraTS'18}, MNMG-sum has the largest maximum neuron sparsity 81.32\%; however, the accuracy loss can reach up to ${\sim}$8.48\%.
In contrast, while MPMG-sum has the second-largest maximum neuron sparsity 78.17\%, the accuracy loss is much smaller than MNMG-sum.
For \textit{UCF101}, it is surprising that many manners have low accuracy.
As we analyse the reason in the footnote in Table~\ref{tb:neuron_importance_full}, with the extreme neuron sparsity, some layers of the pruned networks have only 1 neuron retained, losing sufficient features for learning, and thus, leading to low accuracy.

Hence, considering the comprehensive performance of reducing resources and maintaining the accuracy, MPMG-sum is selected as vanilla NP.
Note that any neuron sparsity greater than the maximum neuron sparsity will make the pruned network infeasible by pruning the whole layer(s).
\vspace{-1mm}

\subsection{Initialization for Neuron Imbalance}
The imbalanced layer-wise distribution of neuron importance hinders pruning at a high sparsity level due to the pruning of the whole layer(s).
For 2D classification tasks in \cite{signal_propagation}, orthogonal initialization is used to effectively solve this problem for balancing the importance of parameters; but it does not improve our neuron pruning results in 3D tasks and even leads to a poor pruning capability with a lower maximum neuron sparsity than Glorot initialization~\cite{xavier}.
This is briefly mentioned in Sec.~\myblue{4.1} in the main paper.
Here, we compare the resource reducing capability using Glorot initialization and orthogonal initialization.

\begin{table}[t]
\centering
\caption{Impact of parameter initialization on neuron pruning.
``ort": orthogonal initialization;
``xn": Glorot initialization;
``f": FLOPs.
``Sparsity" is the least max neuron sparsity among all manners to ensure the network feasibility.
RANP-f with Glorot initialization achieves the least FLOPs and memory consumption.}
\label{tb:orthogonal}
\setlength{\tabcolsep}{2pt}
\resizebox{0.48\textwidth}{!}{
\begin{tabular}{llrrrr}
  \hline
  \multicolumn{1}{c}{Dataset(Model)}
  & Manner & Sparsity(\%) & Param(MB) & GFLOPs & Mem(MB) \\
  \hline
  \multirow{5}{*}{\makecell[l]{ShapeNet\\ (3D-UNet)}}
  & Full\cite{3dunet} & 0
  & 62.26
  & 237.85
  & 997.00 \\
  & Vanilla-ort & 70.53
  & 4.40
  & 72.65
  & 630.00 \\
  & Vanilla-xn & 70.53
  & 4.56
  & 73.22
  & 618.35 \\
  & RANP-f-ort & 70.53
  & 5.40
  & 21.73
  & 366.29 \\
  & RANP-f-xn & 70.53
  & 5.52
  & \textbf{15.06}
  & \textbf{328.66} \\
  \hline
  \multirow{5}{*}{\makecell[l]{BraTS'18\\ (3D-UNet)}}
  & Full\cite{3dunet} & 0
  & 15.57
  & 478.13
  & 3628.00 \\
  & Vanilla-ort\cite{signal_propagation} & 72.20
  & 0.95
  & 159.91
  & 2240.33 \\
  & Vanilla-xn & 72.20
  & 0.92
  & 130.28
  & 2109.19 \\
  & RANP-f-ort & 72.20
  & 1.24
  & 33.28
  & 967.56 \\
  & RANP-f-xn & 72.20
  & 1.29
  & \textbf{23.31}
  & \textbf{850.56} \\
  \hline
  \multirow{5}{*}{\makecell[l]{UCF101\\ (MobileNetV2)}}
  & Full\cite{mobilenetv2} & 0
  & 9.47
  & 0.58
  & 157.47 \\
  & Vanilla-ort\cite{signal_propagation} & 30.21
  & 6.80
  & 0.56
  & 155.71 \\
  & Vanilla-xn & 30.21
  & 6.77
  & 0.55
  & 155.48 \\
  & RANP-f-ort & 30.21
  & 5.12
  & 0.32
  & 105.88 \\
  & RANP-f-xn & 30.21
  & 5.19
  & \textbf{0.28}
  & \textbf{94.50} \\
  \hline
  \multirow{5}{*}{\makecell[l]{UCF101\\ (I3D)}}
  & Full\cite{i3d} & 0
  & 47.27
  & 27.88
  & 201.28 \\
  & Vanilla-ort\cite{signal_propagation} & 24.24
  & 30.56
  & 25.83
  & 192.70 \\
  & Vanilla-xn & 24.24
  & 30.64
  & 25.85
  & 192.88 \\
  & RANP-f-ort & 24.24
  & 27.39
  & 15.94
  & 144.10 \\
  & RANP-f-xn & 24.24
  & 27.38
  & \textbf{14.63}
  & \textbf{133.80} \\
  \hline
\end{tabular}}
\vspace{-3mm}
\end{table}

\begin{figure*}[!t]
\begin{center}
  \begin{subfigure}[b]{0.23\textwidth}
  \centering
  \includegraphics[width=\textwidth]{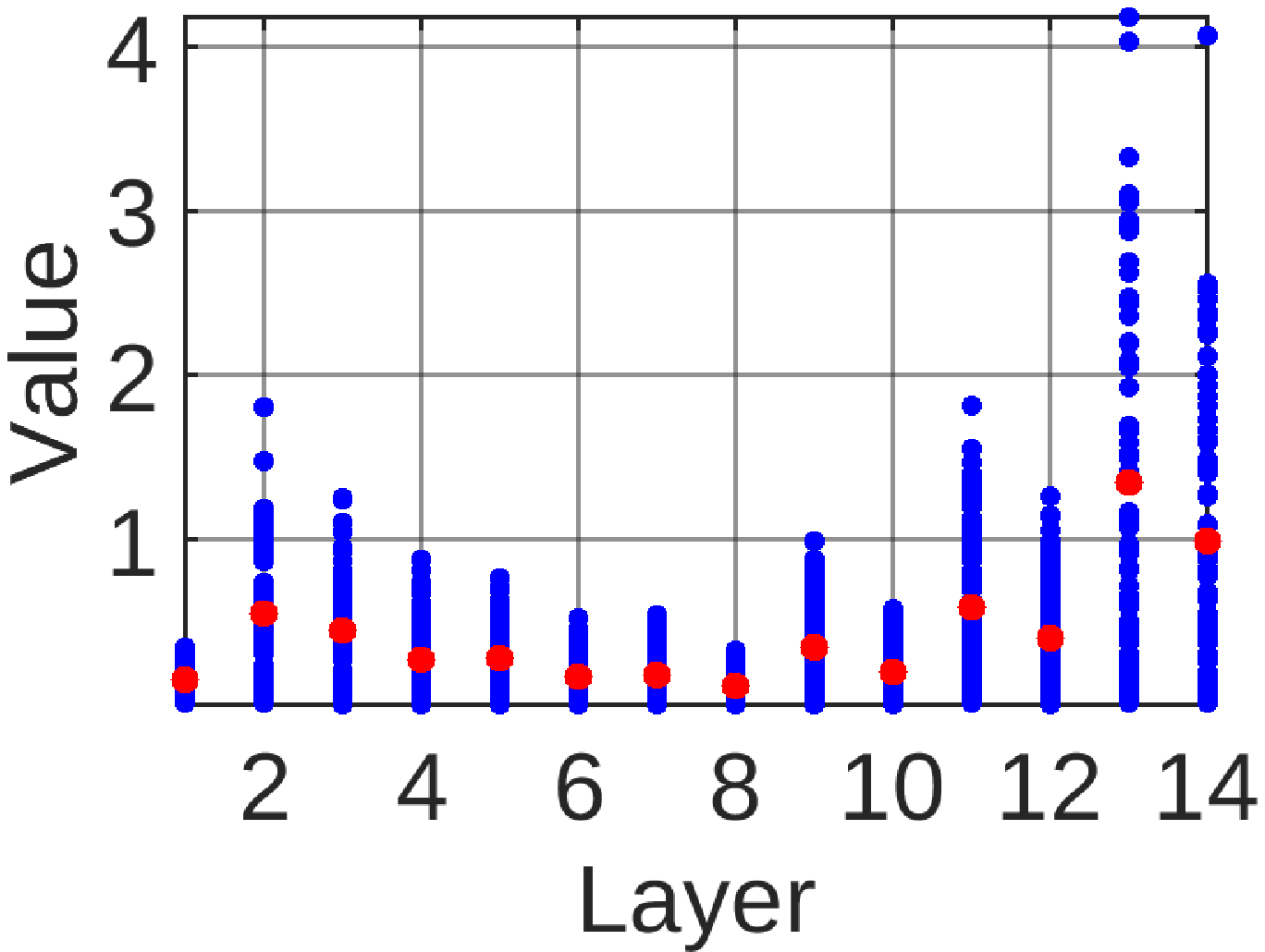}
  \caption{\scriptsize{ShapeNet, Vanilla NP-ort}}
  \label{fig:shapenet_vanilla_ort_neuron_value}
  \end{subfigure}
  ~
  \begin{subfigure}[b]{0.23\textwidth}
  \centering
  \includegraphics[width=\textwidth]{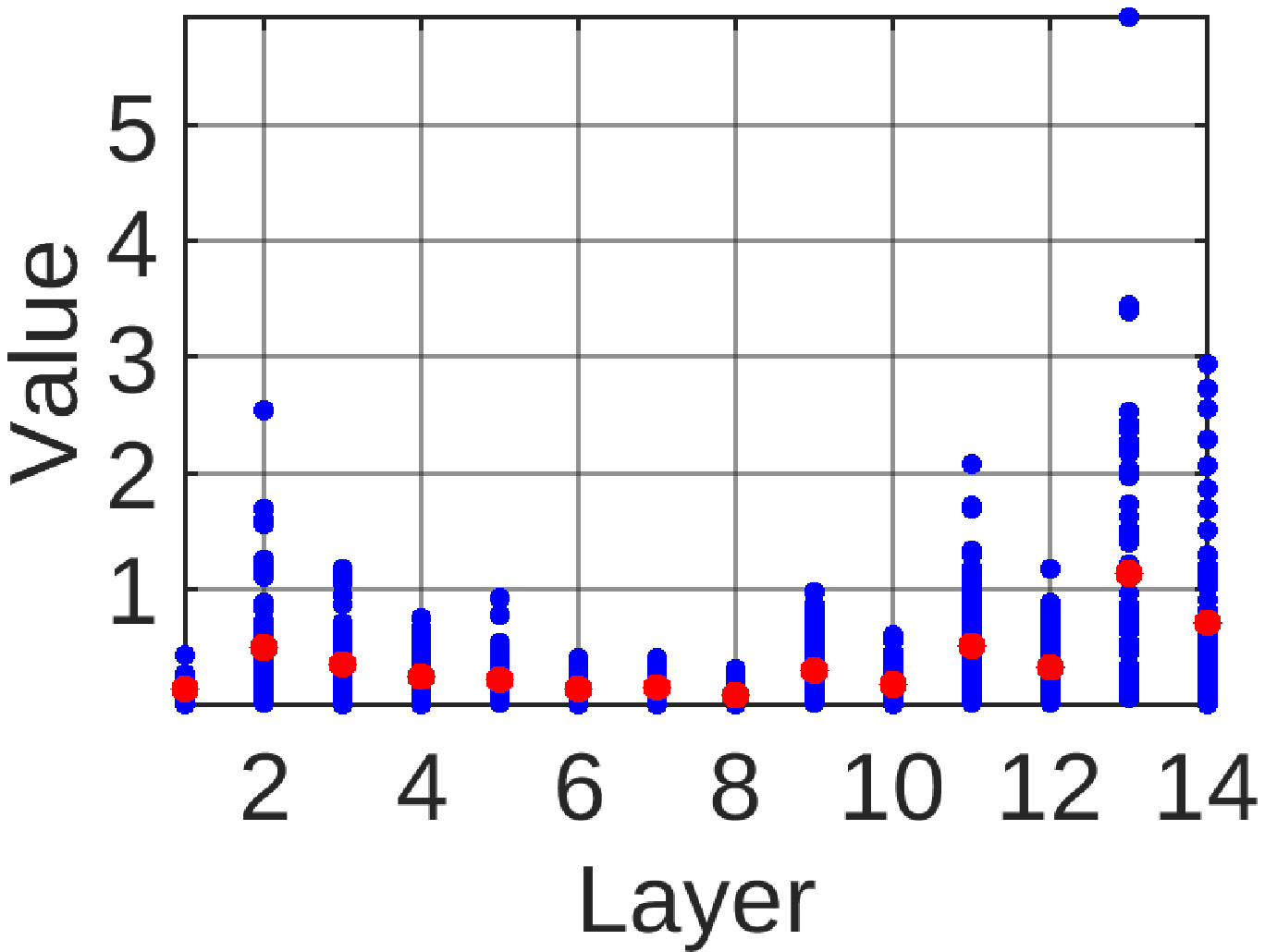}
  \caption{\scriptsize{ShapeNet, Vanilla NP-xn}}
  \label{fig:shapenet_vanilla_xn_neuron_value}
  \end{subfigure}
  ~
  \begin{subfigure}[b]{0.23\textwidth}
  \centering
  \includegraphics[width=\textwidth]{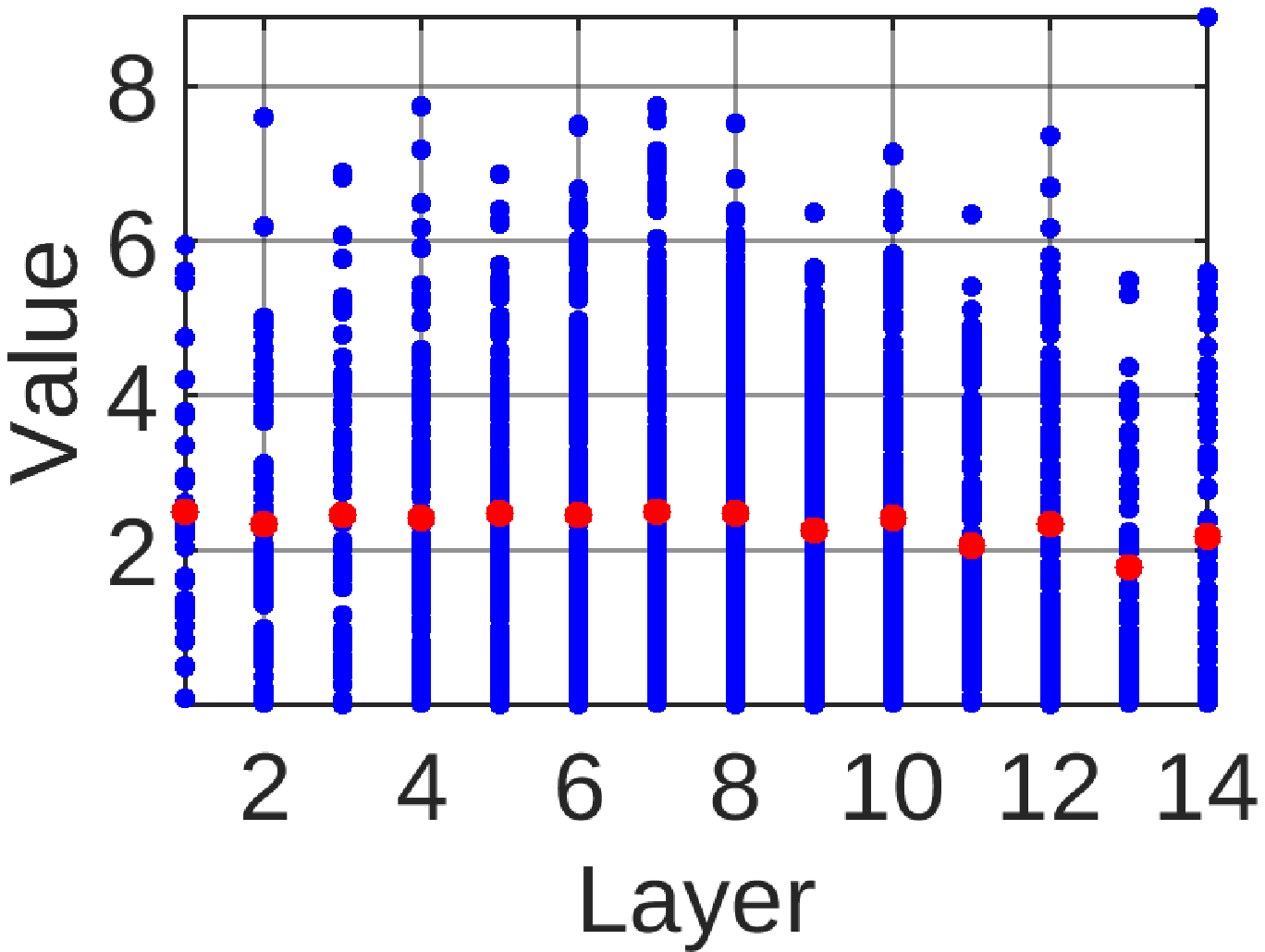}
  \caption{\scriptsize{ShapeNet, RANP-f-ort}}
  \label{fig:shapenet_ranp_ort_neuron_value}
  \end{subfigure}
  ~
  \begin{subfigure}[b]{0.23\textwidth}
  \centering
  \includegraphics[width=\textwidth]{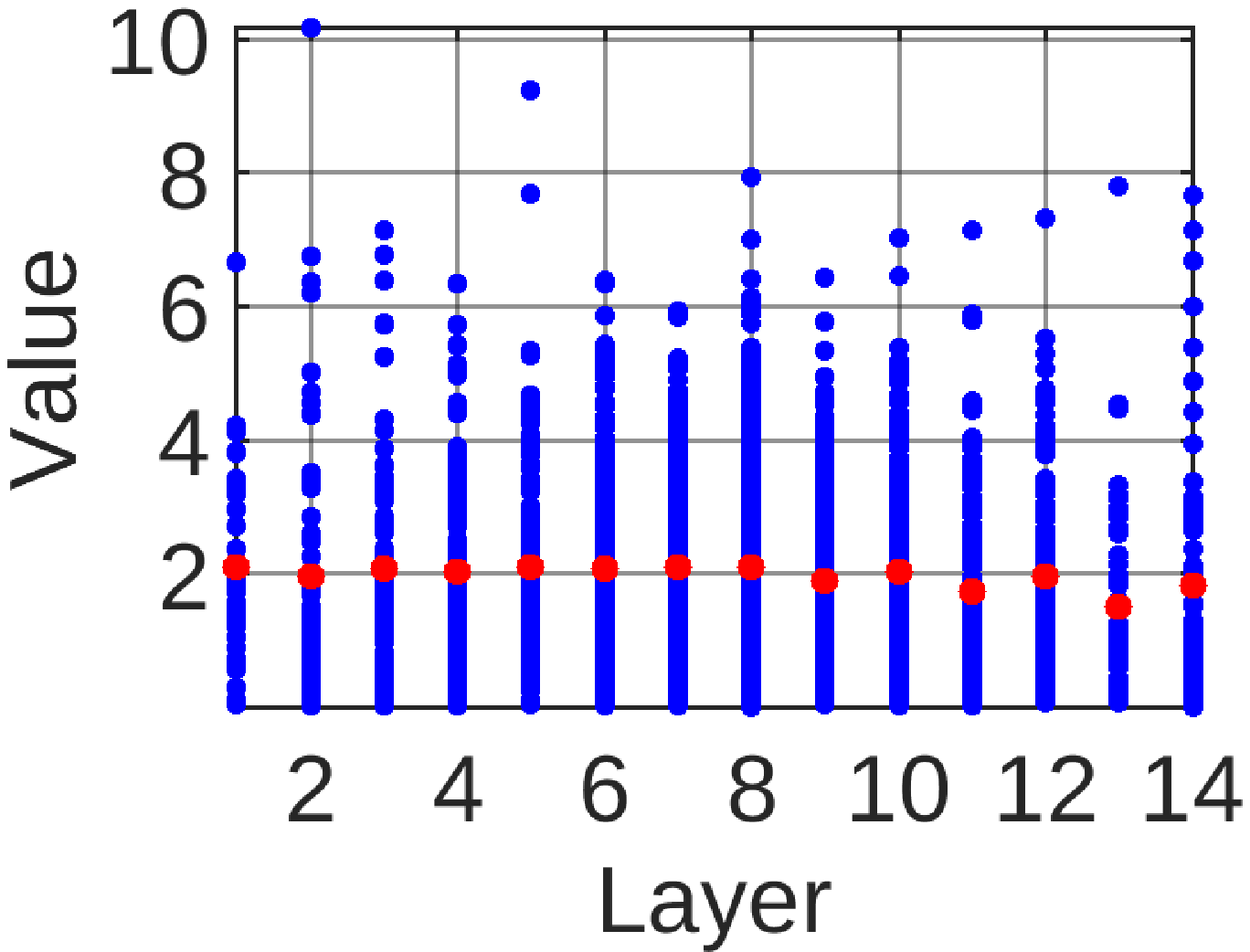}
  \caption{\scriptsize{ShapeNet, RANP-f-xn}}
  \label{fig:shapenet_ranp_xn_neuron_value}
  \end{subfigure}
  \\
  \begin{subfigure}[b]{0.23\textwidth}
  \centering
  \includegraphics[width=\textwidth]{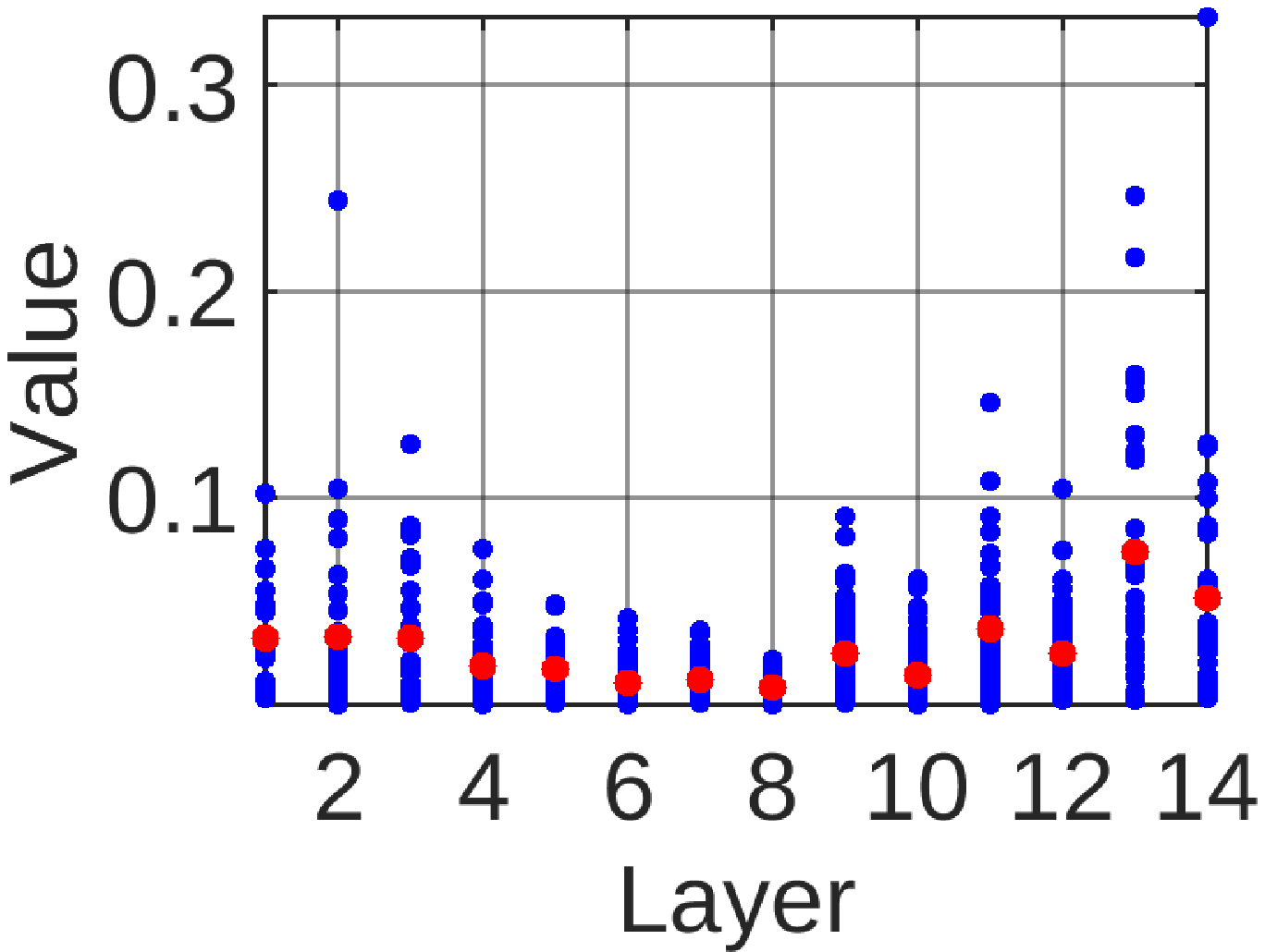}
  \caption{\scriptsize{BraTS'18, Vanilla NP-ort}}
  \label{fig:brats_vanilla_ort_neuron_value}
  \end{subfigure}
  ~
  \begin{subfigure}[b]{0.23\textwidth}
  \centering
  \includegraphics[width=\textwidth]{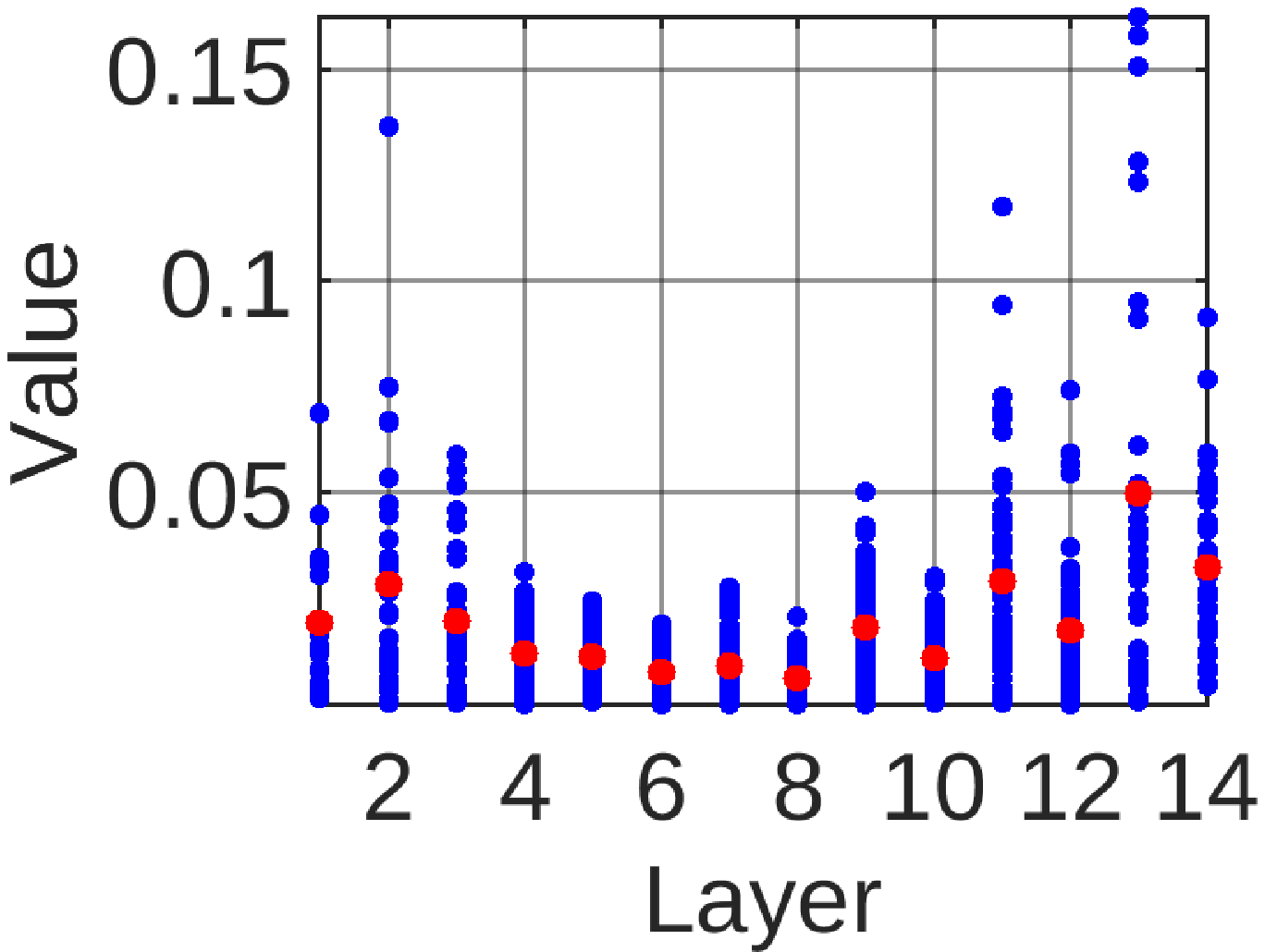}
  \caption{\scriptsize{BraTS'18, Vanilla NP-xn}}
  \label{fig:brats_vanilla_xn_neuron_value}
  \end{subfigure}
  ~
  \begin{subfigure}[b]{0.23\textwidth}
  \centering
  \includegraphics[width=\textwidth]{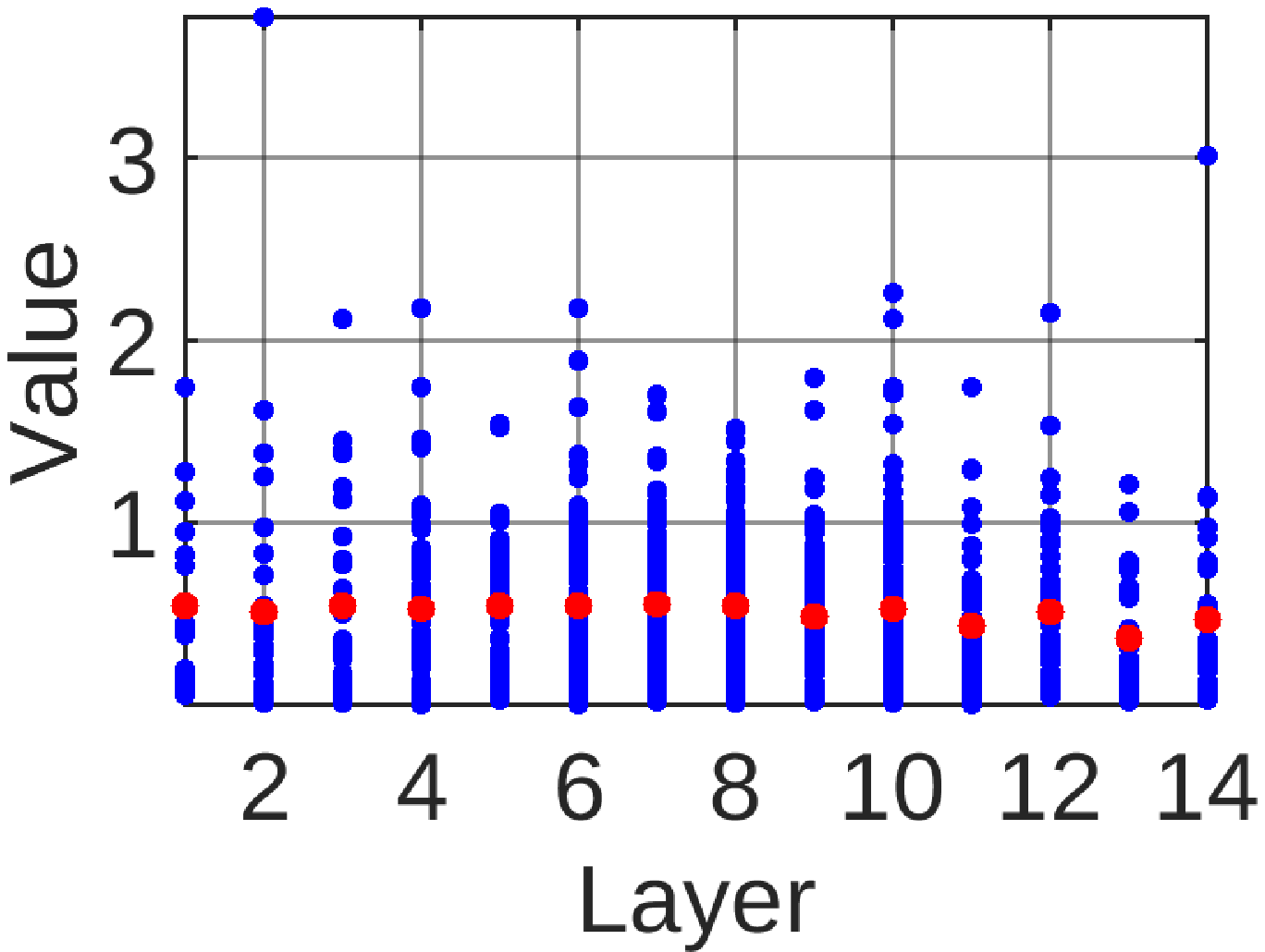}
  \caption{\scriptsize{BraTS'18, RANP-f-ort}}
  \label{fig:brats_ranp_ort_neuron_value}
  \end{subfigure}
  ~
  \begin{subfigure}[b]{0.23\textwidth}
  \centering
  \includegraphics[width=\textwidth]{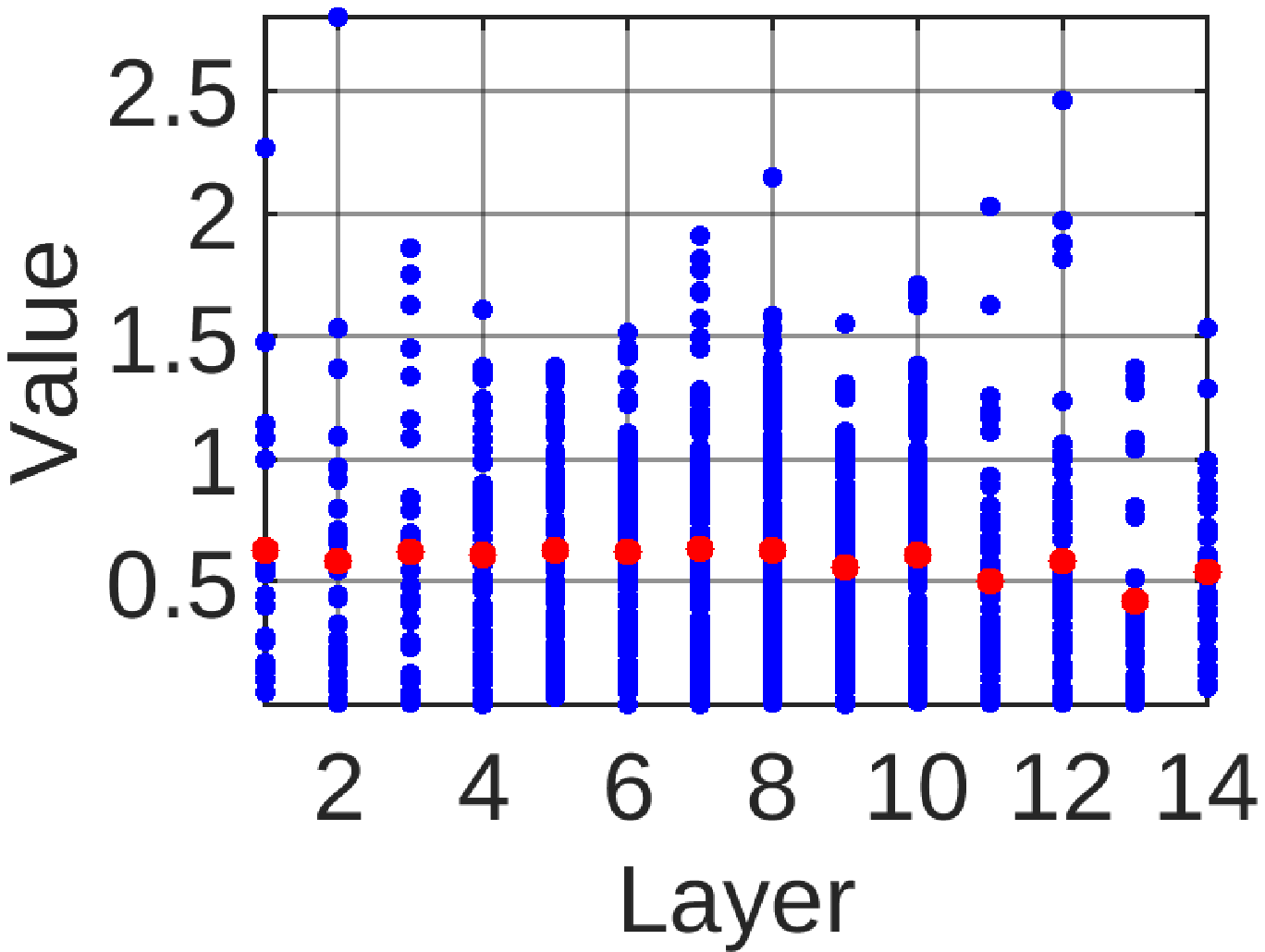}
  \caption{\scriptsize{BraTS'18, RANP-f-xn}}
  \label{fig:brats_ranp_xn_neuron_value}
  \end{subfigure}
\caption{Neuron importance of 15-layer 3D-UNet by MPMG-sum with orthogonal and Glorot initialization.
Blue: neuron values; red: mean values.
By vanilla NP, orthogonal initialization does not result in a balanced neuron importance distribution compared to Glorot initialization whereas by our RANP-f, the values are more balanced and resource aware on FLOPs, enabling pruning at the extreme sparsity.
}
\label{fig:ort_xn_neuron_value}
\end{center}
\vspace{-5mm}
\end{figure*}

\begin{figure*}[!t]
\begin{center}
  \begin{subfigure}[b]{0.23\textwidth}
  \centering
  \includegraphics[width=\textwidth]{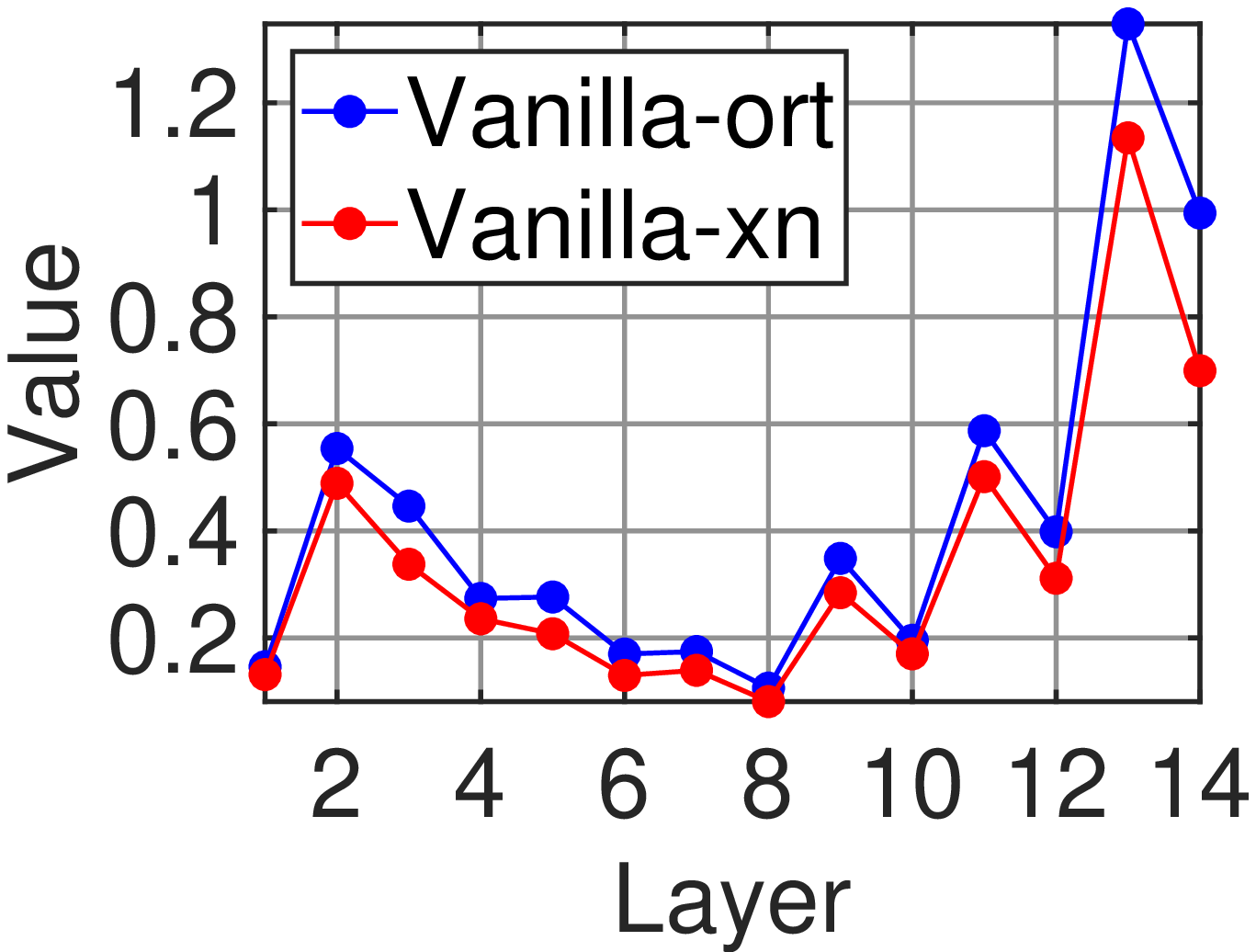}
  \caption{ShapeNet}
  \label{fig:shapenet_vanilla_ort_xn}
  \end{subfigure}
  ~
  \begin{subfigure}[b]{0.23\textwidth}
  \centering
  \includegraphics[width=\textwidth]{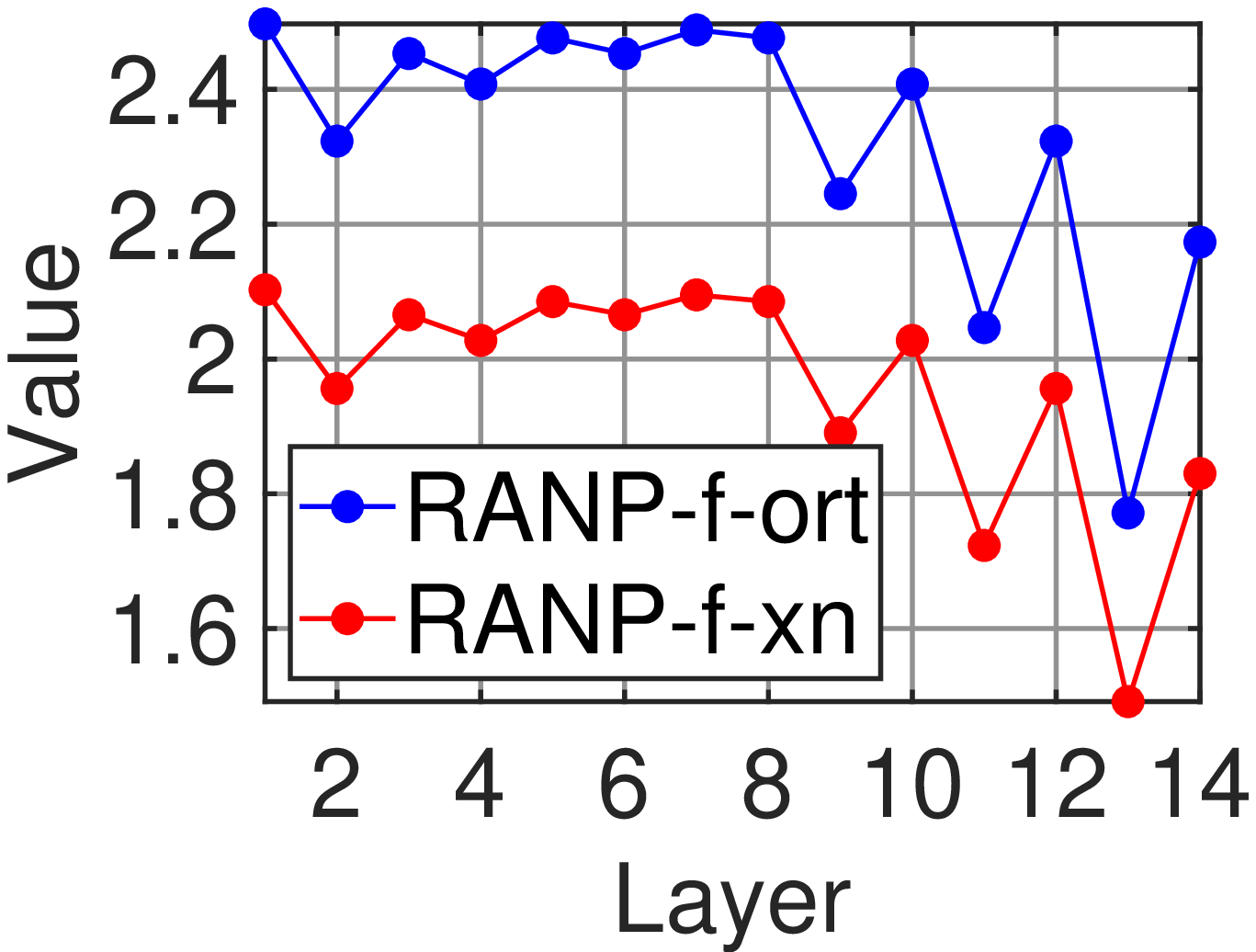}
  \caption{ShapeNet}
  \label{fig:shapenet_ranp_ort_xn}
  \end{subfigure}
  ~
  \begin{subfigure}[b]{0.23\textwidth}
  \centering
  \includegraphics[width=\textwidth]{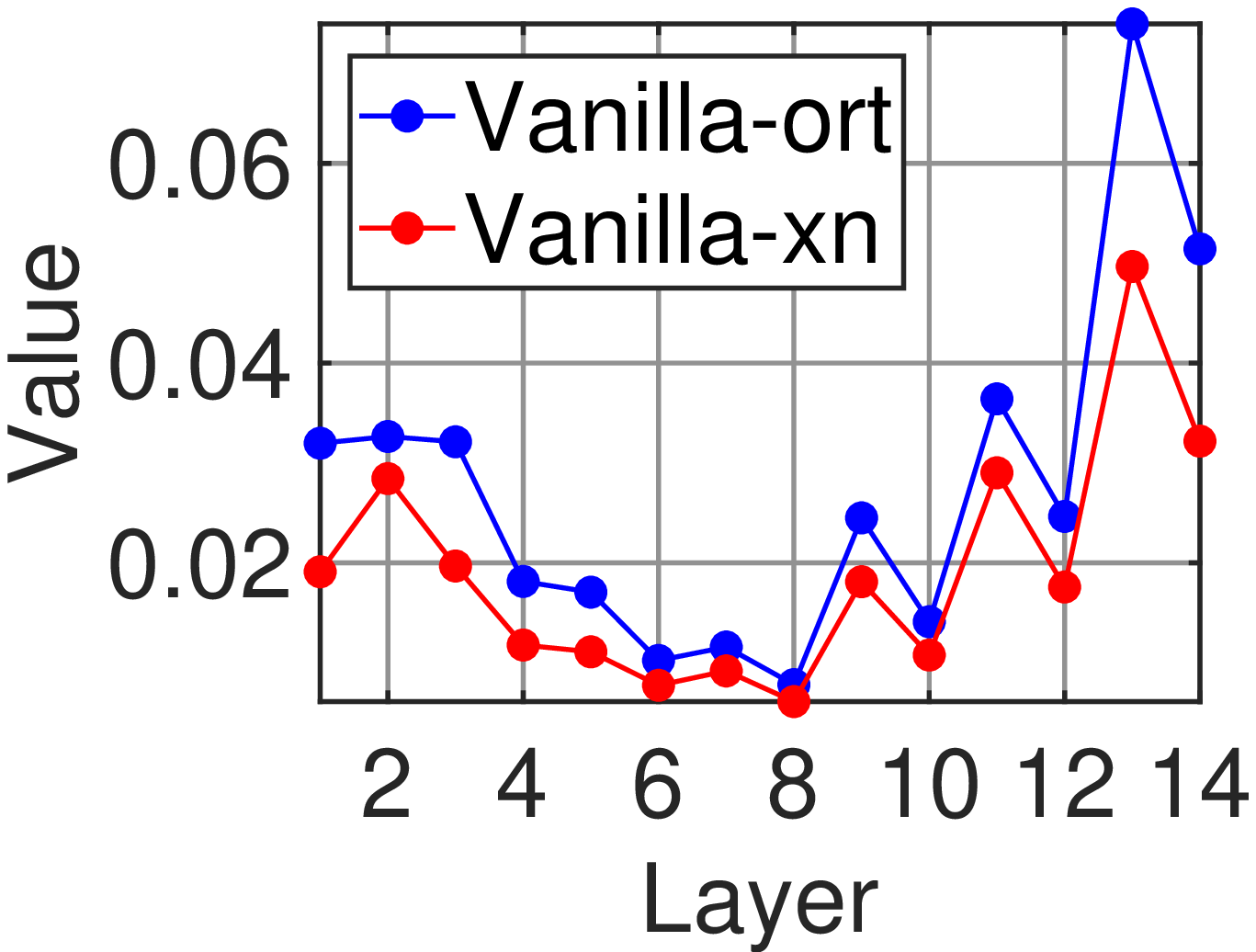}
  \caption{BraTS'18}
  \label{fig:brats_vanilla_ort_xn}
  \end{subfigure}
  ~
  \begin{subfigure}[b]{0.23\textwidth}
  \centering
  \includegraphics[width=\textwidth]{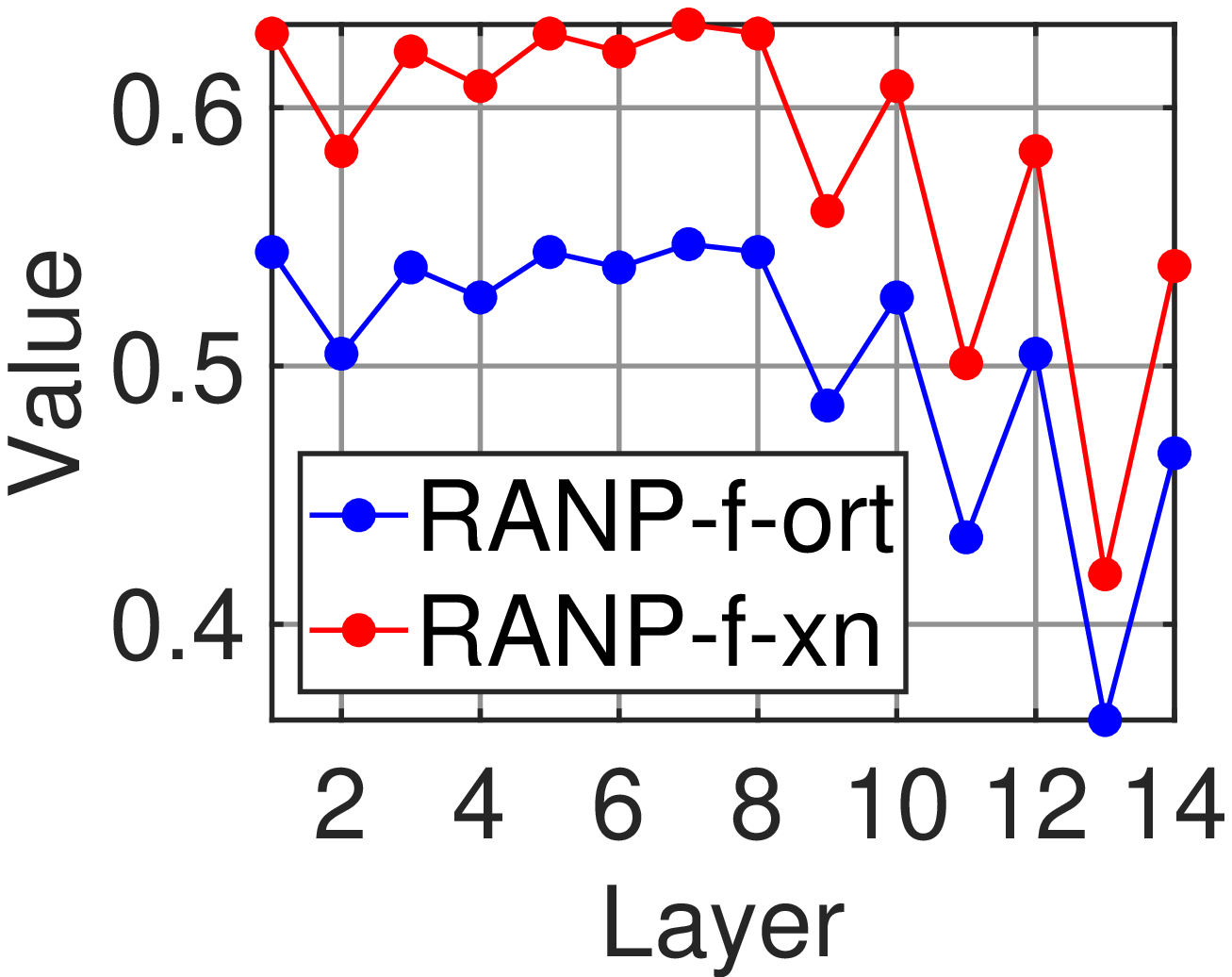}
  \caption{BraTS'18}
  \label{fig:brats_ranp_ort_xn}
  \end{subfigure}
  \\
  \begin{subfigure}[b]{0.4\textwidth}
  \centering
  \includegraphics[width=\textwidth]{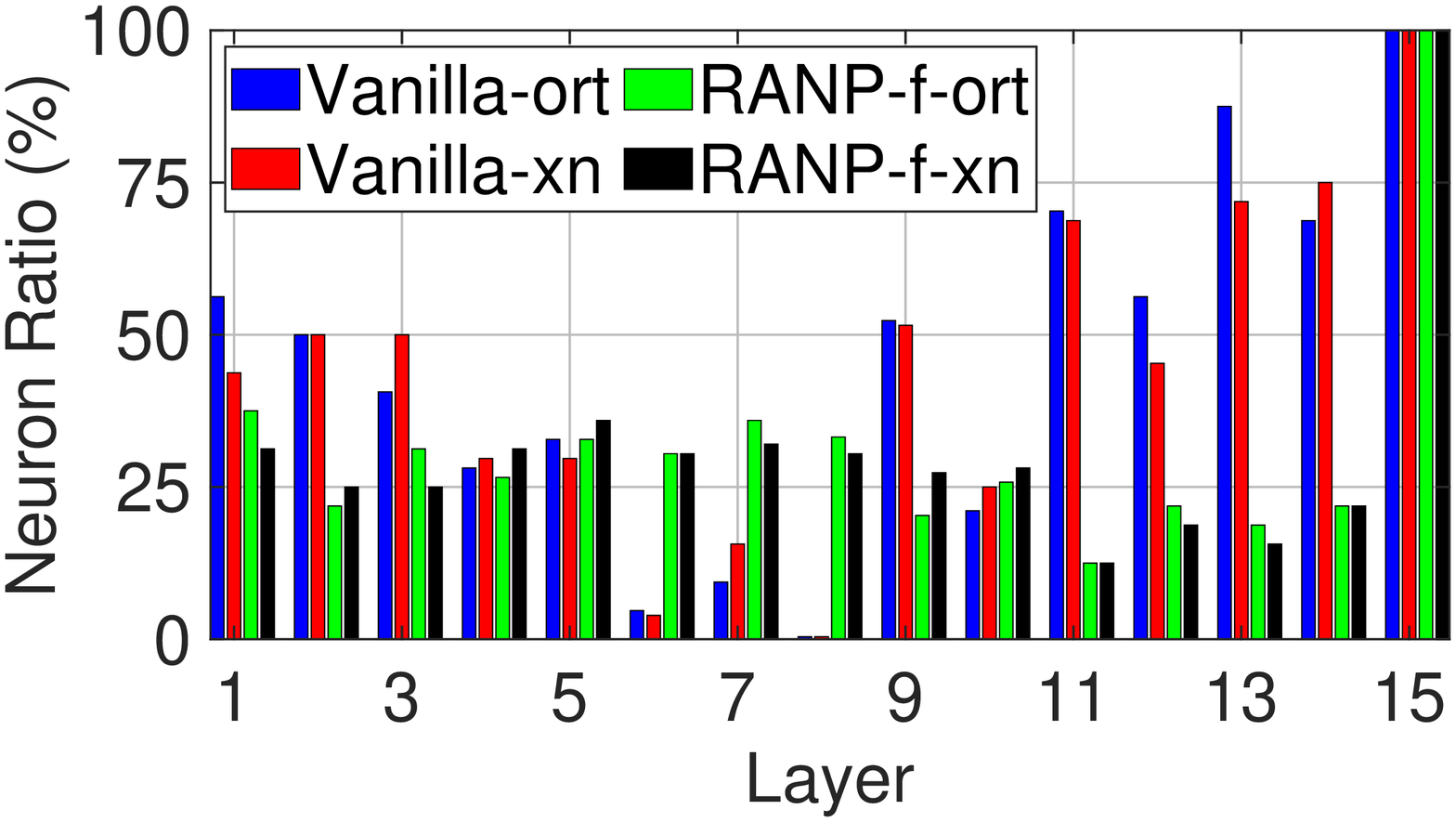}
  \caption{ShapeNet}
  \label{fig:shapenet_ort_xn_ratio}
  \end{subfigure}
  ~
  \hspace{3em}
  \begin{subfigure}[b]{0.4\textwidth}
  \centering
  \includegraphics[width=\textwidth]{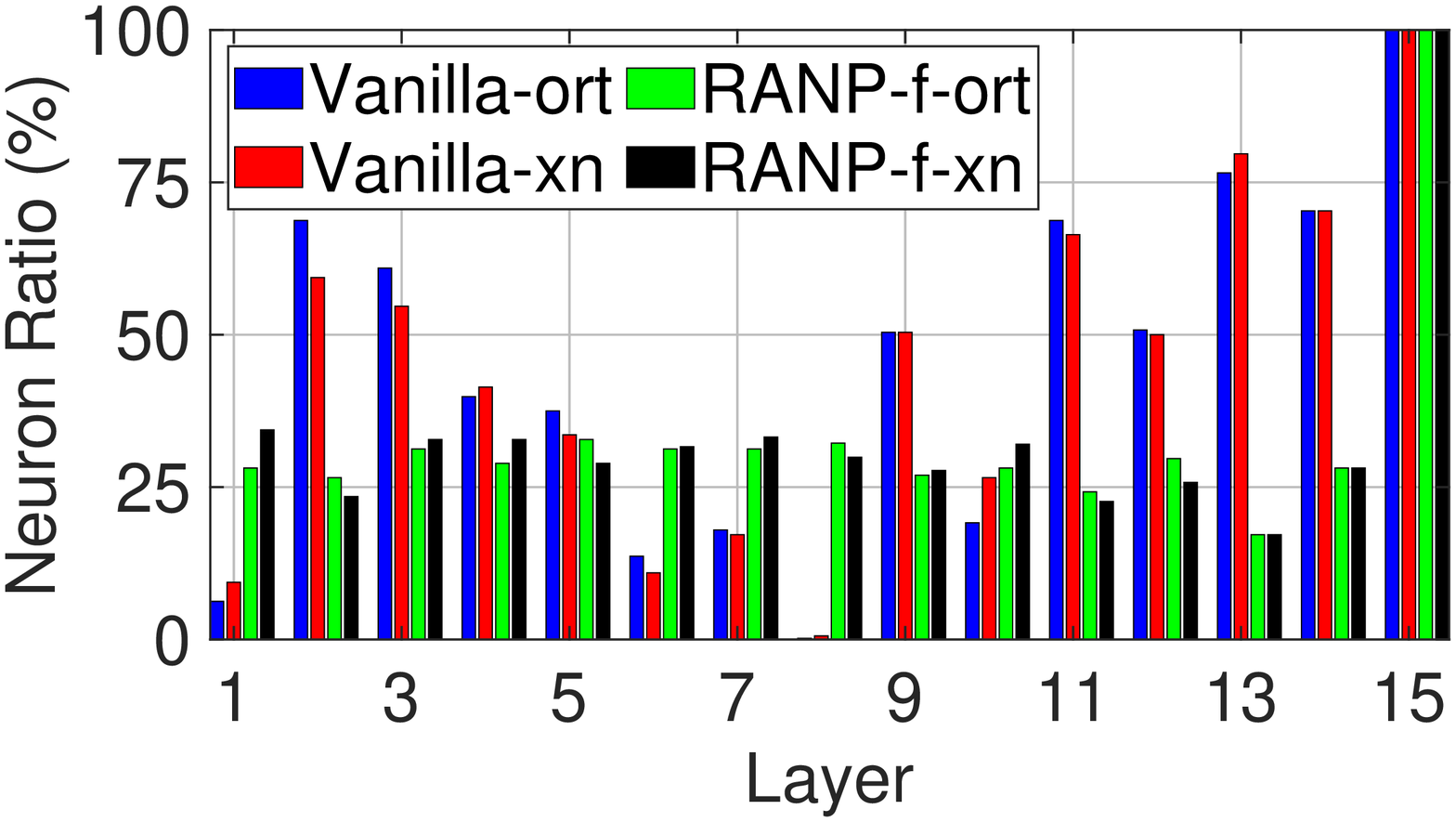}
  \caption{BraTS'18}
  \label{fig:brats_ort_xn_ratio}
  \end{subfigure}
\vspace{-1mm}
\caption{Comparison of neuron distribution with orthogonal and Glorot initialization before and after reweighting. (a)-(d) are neuron importance values. (e)-(f) are neuron retained ratios.
Vanilla versions (both orthogonal and Glorot initializations) prune all the neuron in layer 8, leading to network infeasibility while our RANP-f versions have a balanced distribution of retained neurons.
}
\label{fig:ort_xn}
\end{center}
\vspace{-7mm}
\end{figure*}

\textbf{Resource reductions.} In Table~\myblue{\ref{tb:orthogonal}}, vanilla neuron pruning (\ie, MPMG-sum) with Glorot initialization, \ie, vanilla-xn, achieves smaller FLOPs and memory consumption than those with orthogonal initialization, \ie, vanilla-ort, except FLOPs with 3D-UNet on ShapeNet and I3D on UCF101.
This exception of I3D on UCF101 is possibly caused by the high ratio of 1$^3$ kernel size filters in I3D, \ie, 37 out of 57 convolutional layers, because those 1$^3$ kernel size filters can be regarded as 2D filters on which orthogonal initialization can effectively deal with \cite{signal_propagation}.
While this ratio is also high in MobileNetV2, \ie, 34 out of 52 convolutional layers, it is unnecessary to have the same problem as I3D since it is also affected by the number of neurons in each layer.
Note that since 3D-UNets used are all with 3$^3$ kernel size filters, the orthogonal initialization for 3D-UNet in most cases is inferior to Glorot initialization according to our experiments.
Meanwhile, in Table~\ref{tb:orthogonal}, this gap between vanilla-ort and vanilla-xn is very small on MobileNetV2 and I3D.

Nevertheless, with RANP-f and Glorot initialization, \ie, RANP-f-xn, more FLOPs and memory can be reduced than using orthogonal initialization, \ie, RANP-f-ort.

\textbf{Balance of Neuron Importance Distribution.} More importantly, with reweighting by RANP in Fig.~\myblue{\ref{fig:ort_xn_neuron_value}}, the values of neuron importance are more balanced and stable than those of vanilla neuron importance.
This can largely avoid network infeasibility without pruning the whole layer(s).

Now, we analyse the neuron distribution from the observation of neuron importance values and network structures.
\textit{Fig.~\myblue{\ref{fig:ort_xn}} illustrates a detailed comparison between orthogonal and Glorot initialization by each two subfigures in column of Fig.~\myblue{\ref{fig:ort_xn_neuron_value}}.}
In Figs.~\myblue{\ref{fig:shapenet_vanilla_ort_xn}}-\myblue{\ref{fig:brats_vanilla_ort_xn}}, vanilla neuron importance by Glorot initialization is more stable and compact than that by orthogonal initialization.
After applying the reweighting scheme of RANP-f, the importance tends to be in a similar tendency, shown in Figs.~\myblue{\ref{fig:shapenet_ranp_ort_xn}}-\myblue{\ref{fig:brats_ranp_ort_xn}}.
Consequently, in Figs.~\myblue{\ref{fig:shapenet_ort_xn_ratio}}-\myblue{\ref{fig:brats_ort_xn_ratio}},
neuron ratios are more balanced after the reweighting than without reweighting, especially the 8th layer.
Thus, we choose Glorot initialization as network initialization.
Note that we adopt the same neuron sparsity for these two initialization experiments in Table~\myblue{\ref{tb:orthogonal}} and Fig.~\myblue{\ref{fig:ort_xn}}.

\subsection{Visualization of Balanced Neuron Distribution by RANP}

\begin{figure*}[!ht]
\begin{center}
  \begin{subfigure}[b]{0.23\textwidth}
  \centering
  \includegraphics[width=\textwidth]{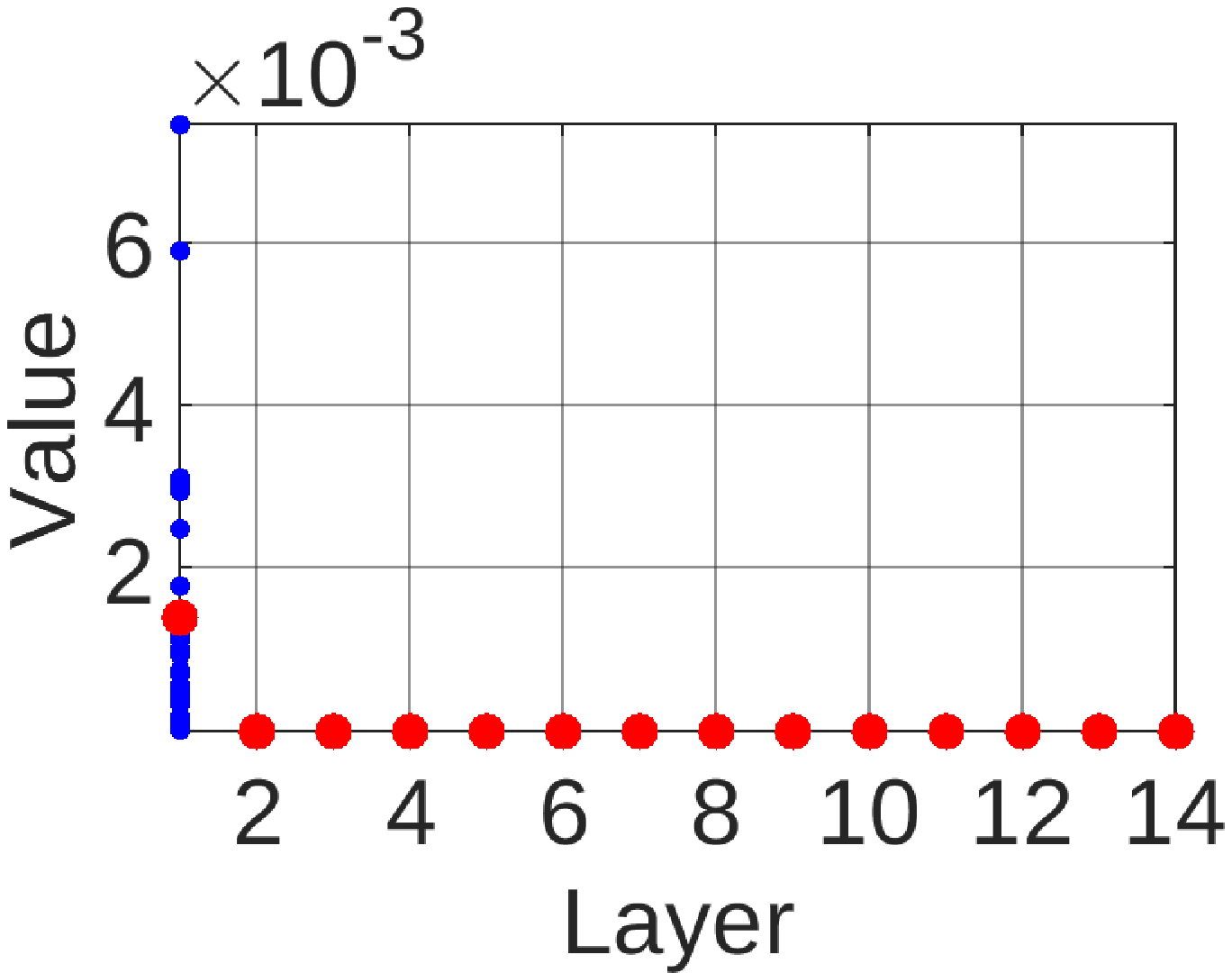}
  \caption{ShapeNet, MNMG-sum}
  \end{subfigure}
  \begin{subfigure}[b]{0.23\textwidth}
  \centering
  \includegraphics[width=\textwidth]{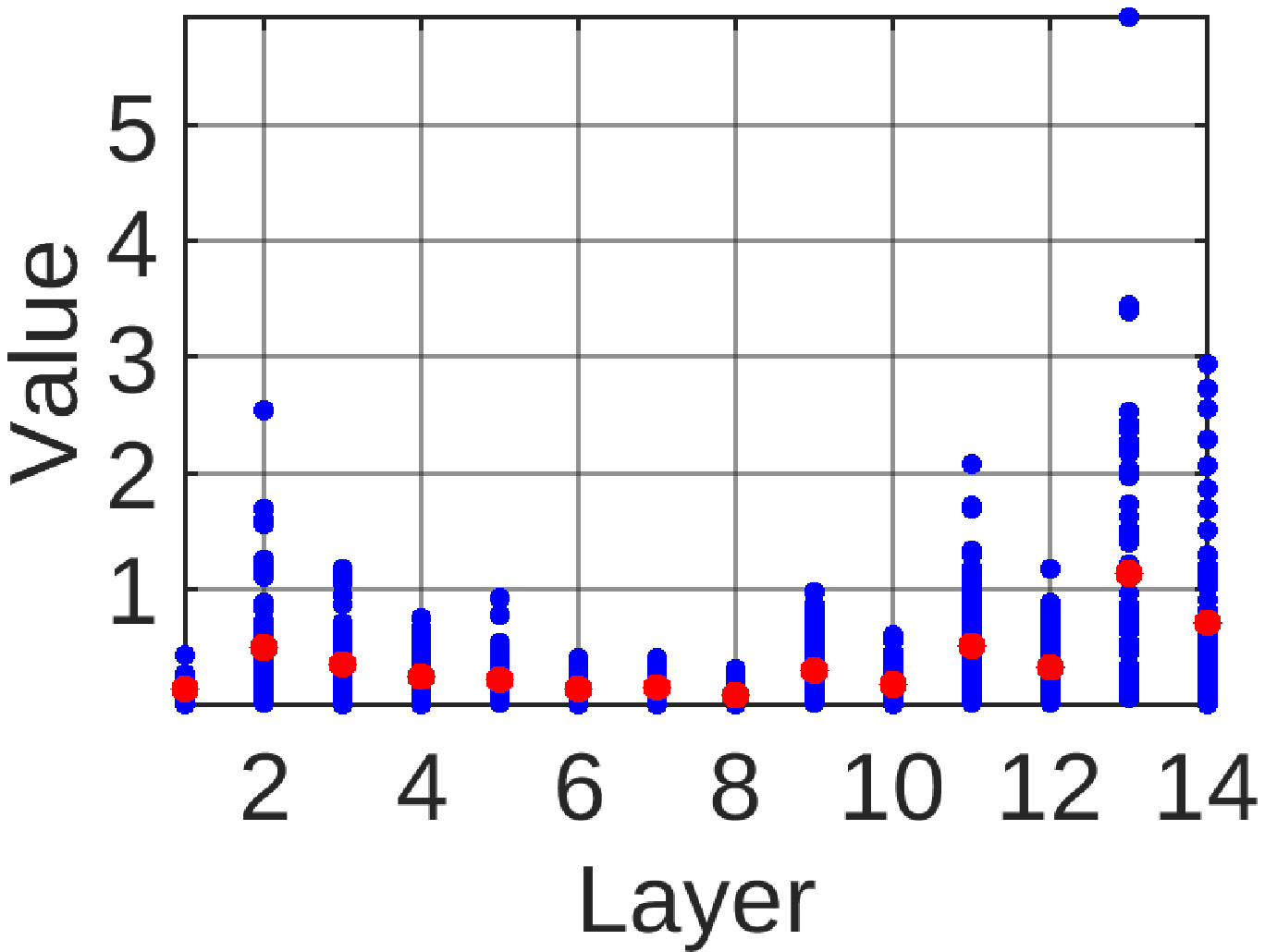}
  \caption{ShapeNet, MPMG-sum}
  \end{subfigure}
  \begin{subfigure}[b]{0.23\textwidth}
  \centering
  \includegraphics[width=\textwidth]{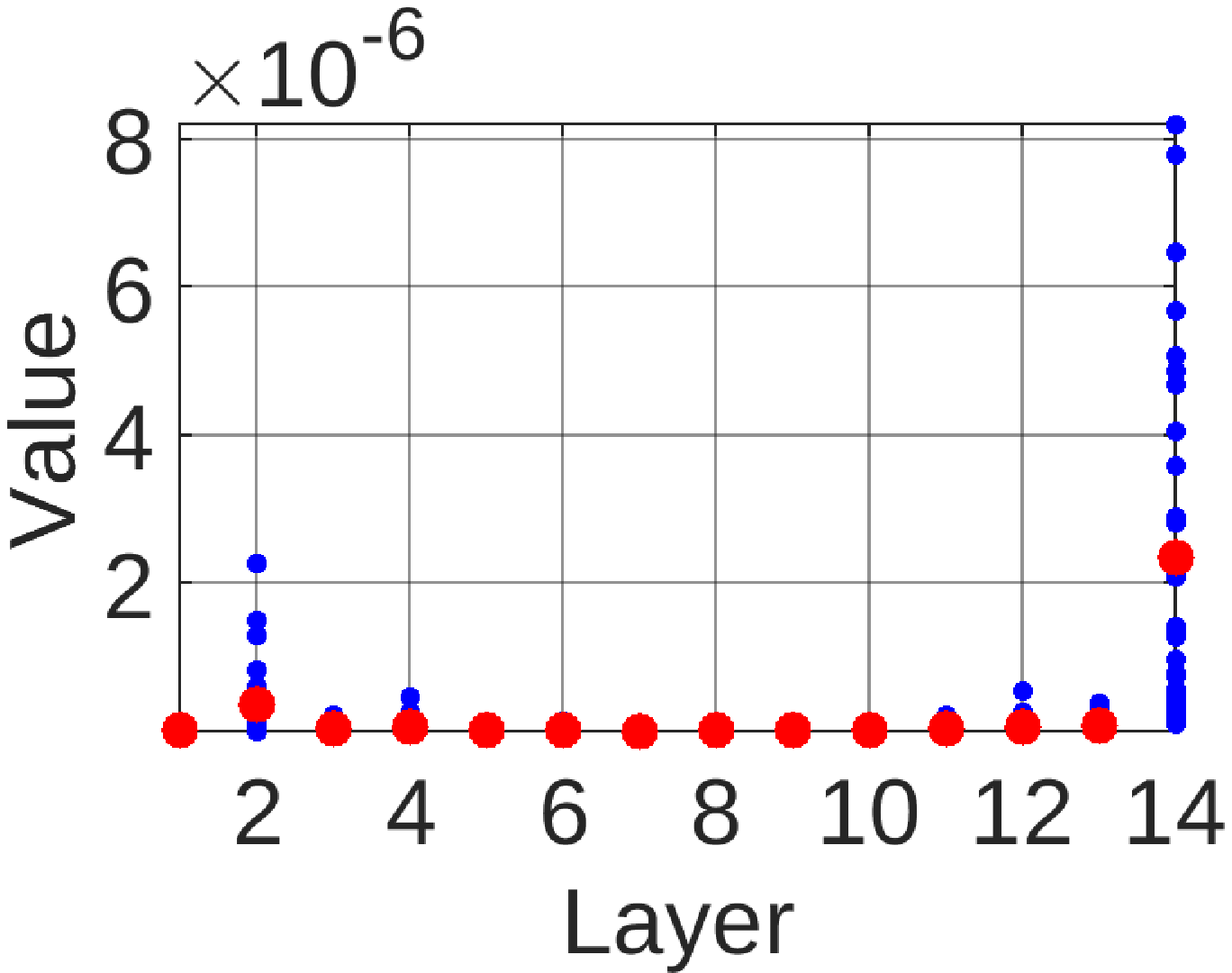}
  \caption{BraTS'18, MNMG-sum}
  \end{subfigure}
  \begin{subfigure}[b]{0.23\textwidth}
  \centering
  \includegraphics[width=\textwidth]{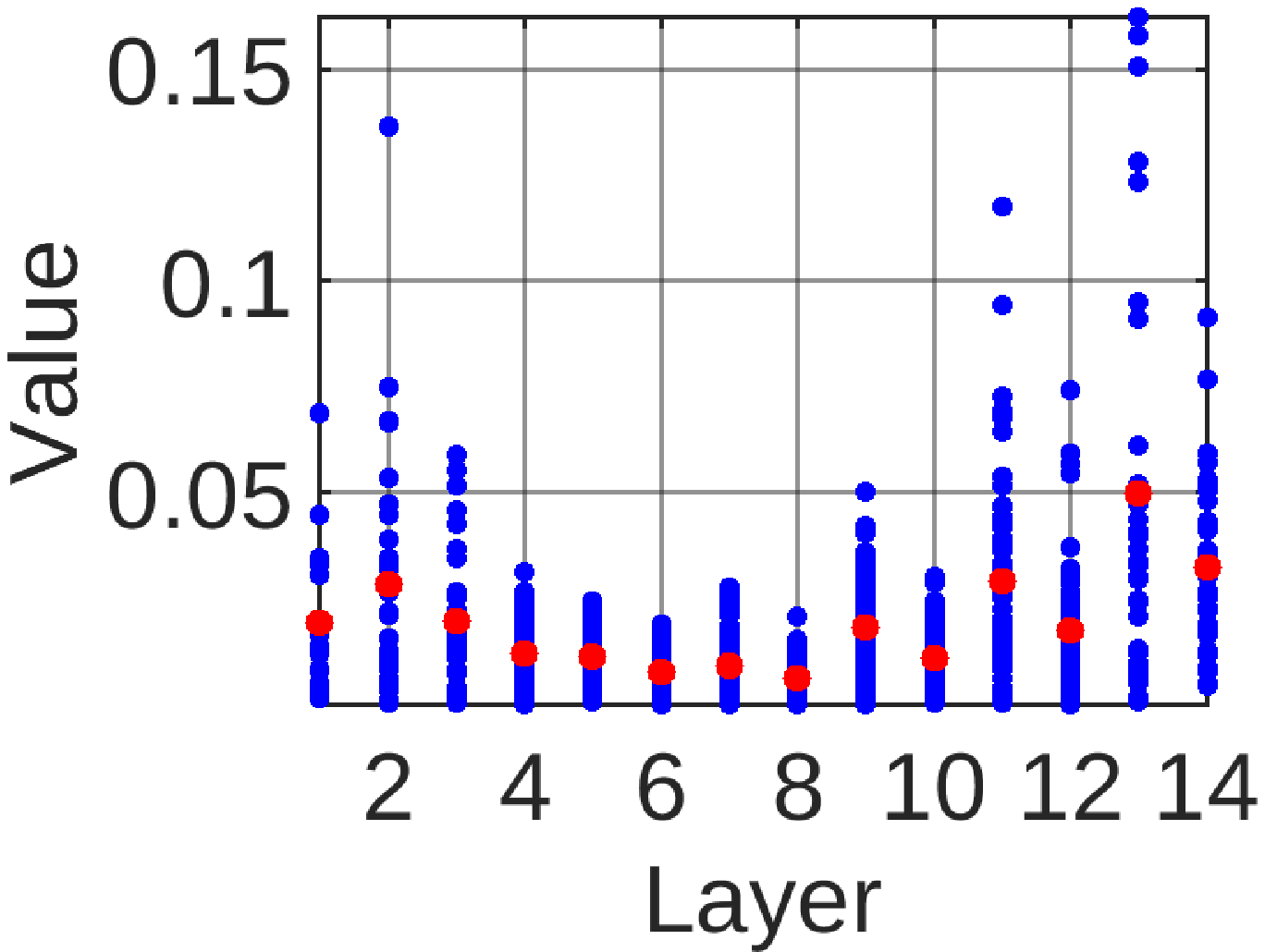}
  \caption{BraTS'18, MPMG-sum}
  \end{subfigure}
\vspace{-1mm}
\caption{MNMG-sum and MPMG-sum on ShapeNet and BraTS'18 with max neuron sparsity in Table~\ref{tb:neuron_importance_full}.
Blue: neuron values; red: mean values.
Clearly, neuron importance distribution by MPMG-sum is more balanced than by MNMG-sum.}
\label{fig:neuron_distribution_mnmg}
\end{center}
\vspace{-5mm}
\end{figure*}

\begin{figure*}[!ht]
\begin{center}
  \begin{subfigure}[b]{0.23\textwidth}
  \centering
  \includegraphics[width=\textwidth]{distribution/old/shapenet_vanilla.eps}
  \caption{\scriptsize{ShapeNet, Vanilla NP Eq.~\myblue{6}}}
  \end{subfigure}
  \begin{subfigure}[b]{0.23\textwidth}
  \centering
  \includegraphics[width=\textwidth]{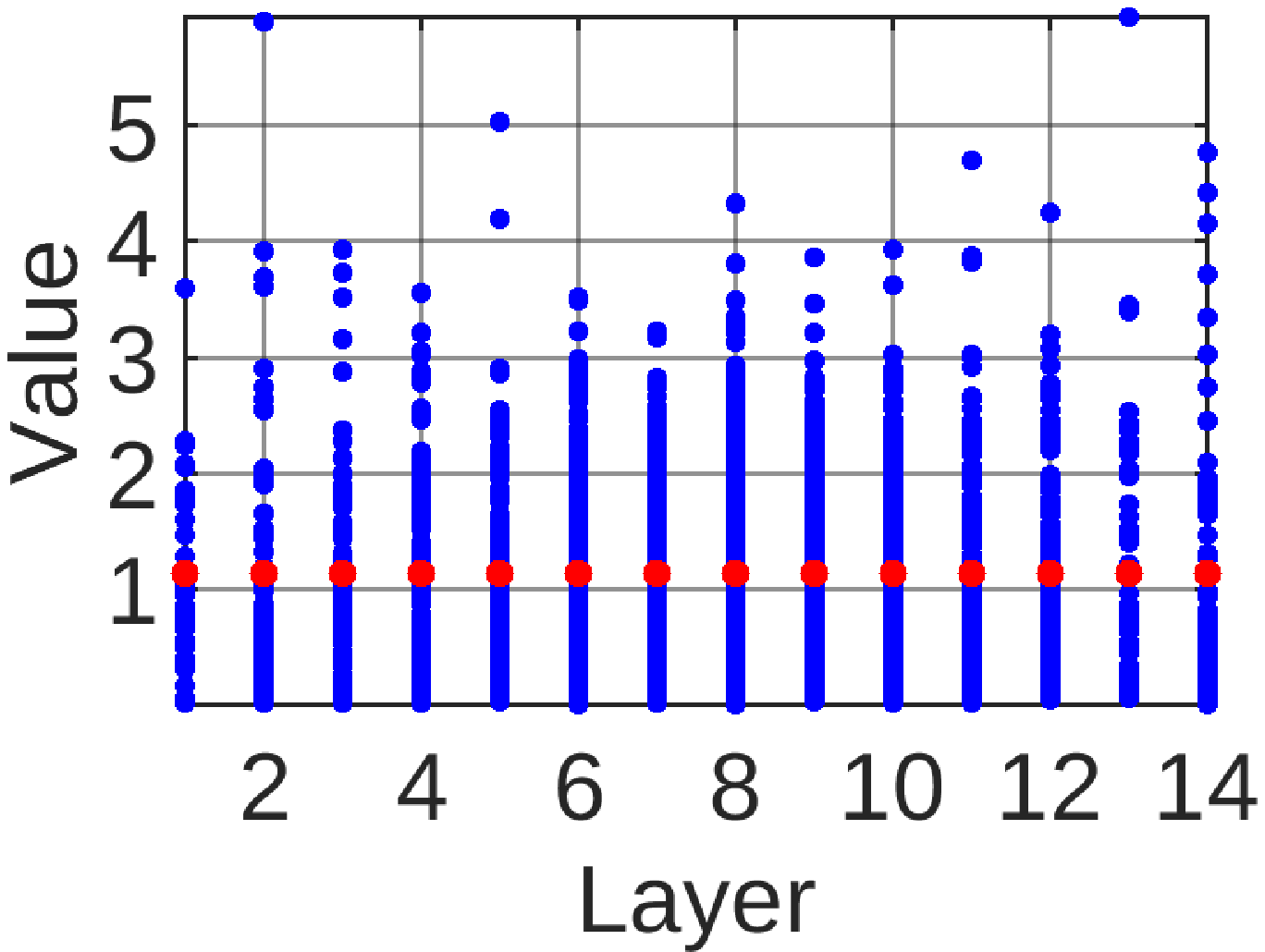}
  \caption{\scriptsize{ShapeNet, Weighted NP Eq.~\myblue{9}}}
  \end{subfigure}
  \begin{subfigure}[b]{0.23\textwidth}
  \centering
  \includegraphics[width=\textwidth]{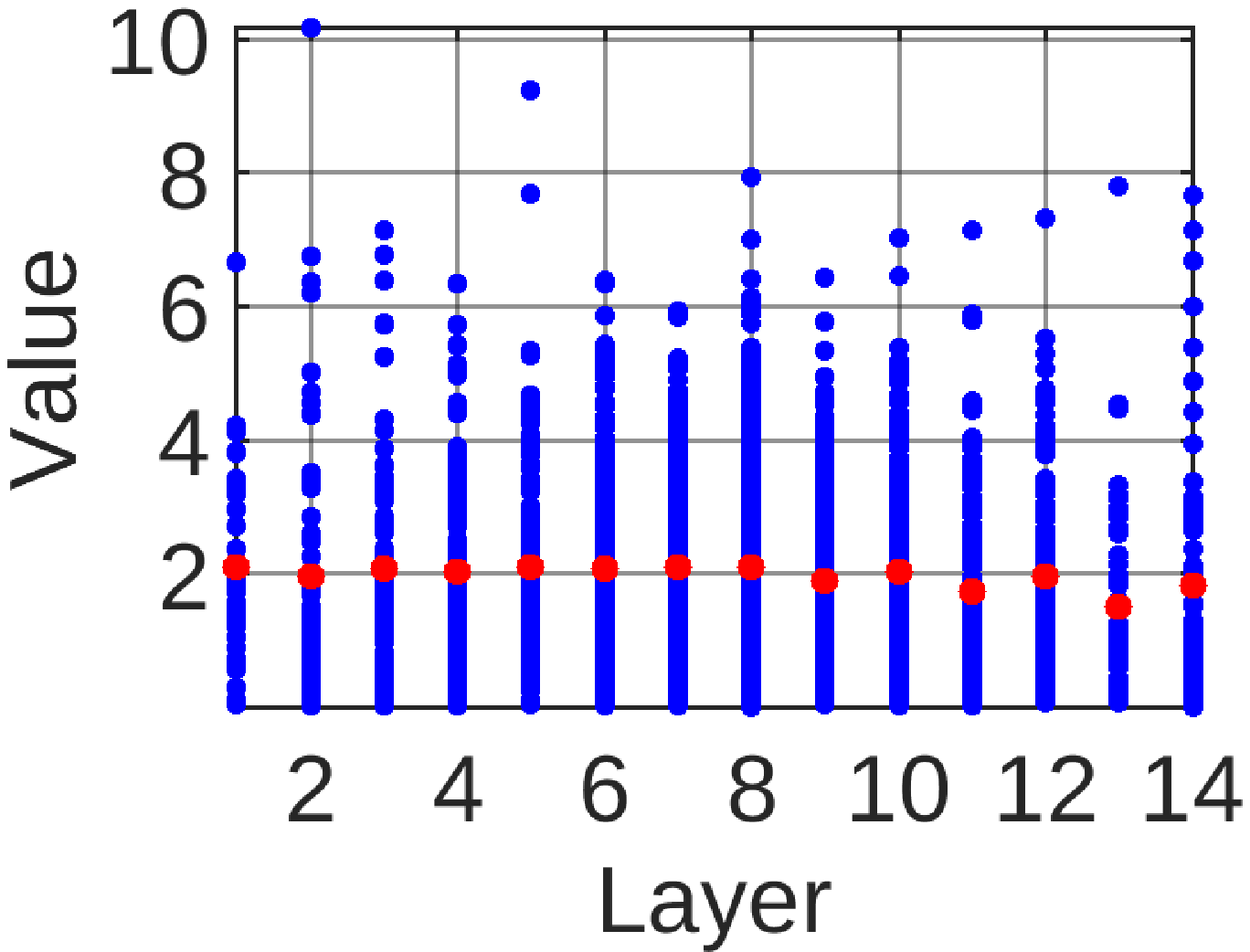}
  \caption{\scriptsize{ShapeNet, RANP-f Eq.~\myblue{10}}}
  \end{subfigure}
  \begin{subfigure}[b]{0.23\textwidth}
  \centering
  \includegraphics[width=\textwidth]{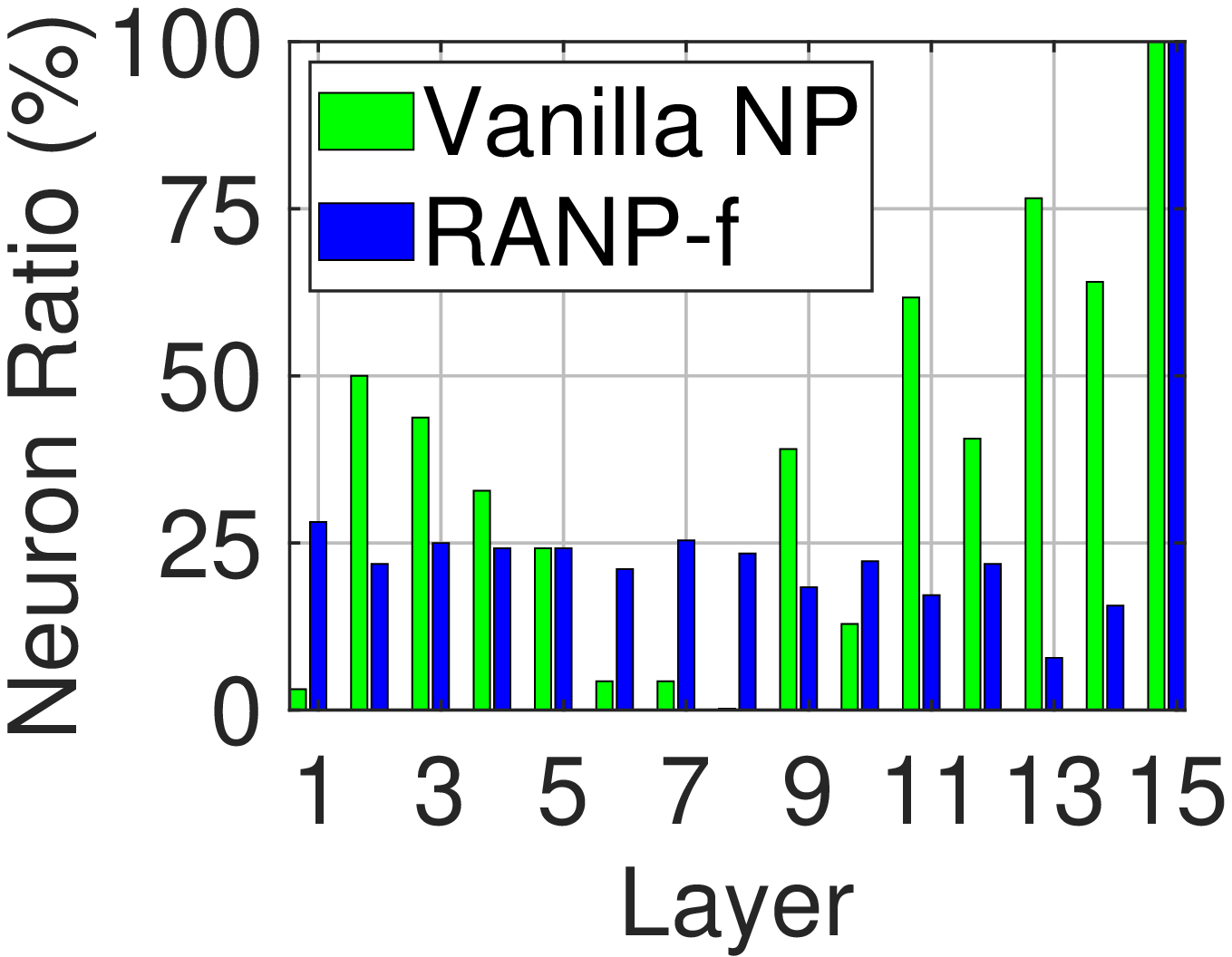}
  \caption{\scriptsize{ShapeNet, Retain ratio}}
  \end{subfigure}
 \\
  \begin{subfigure}[b]{0.23\textwidth}
  \centering
  \includegraphics[width=\textwidth]{distribution/brats_vanilla.eps}
  \caption{\scriptsize{BraTS'18, Vanilla NP Eq.~\myblue{6}}}
  \end{subfigure}
  \begin{subfigure}[b]{0.23\textwidth}
  \centering
  \includegraphics[width=\textwidth]{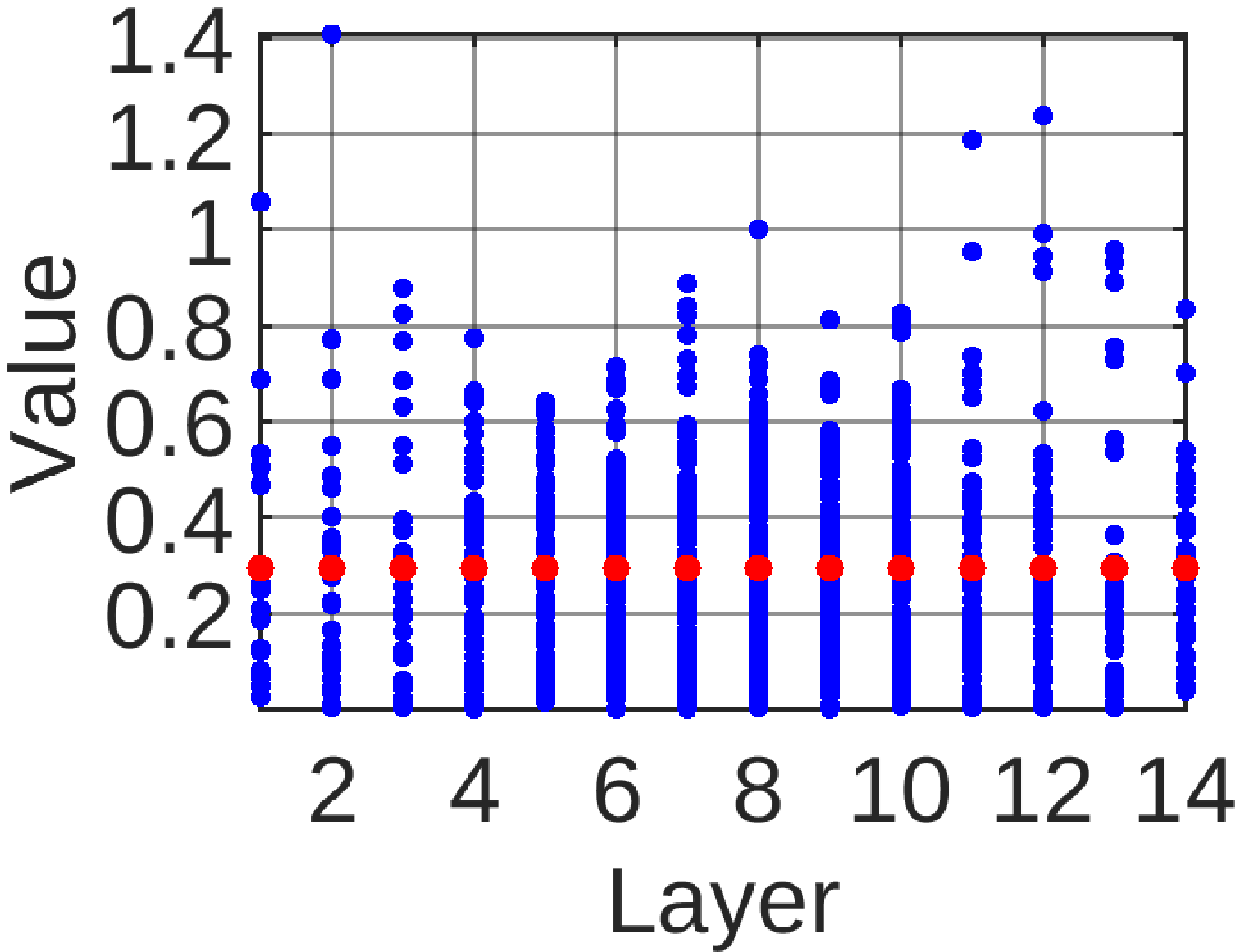}
  \caption{\scriptsize{BraTS'18, Weighted NP Eq.~\myblue{9}}}
  \end{subfigure}
  \begin{subfigure}[b]{0.23\textwidth}
  \centering
  \includegraphics[width=\textwidth]{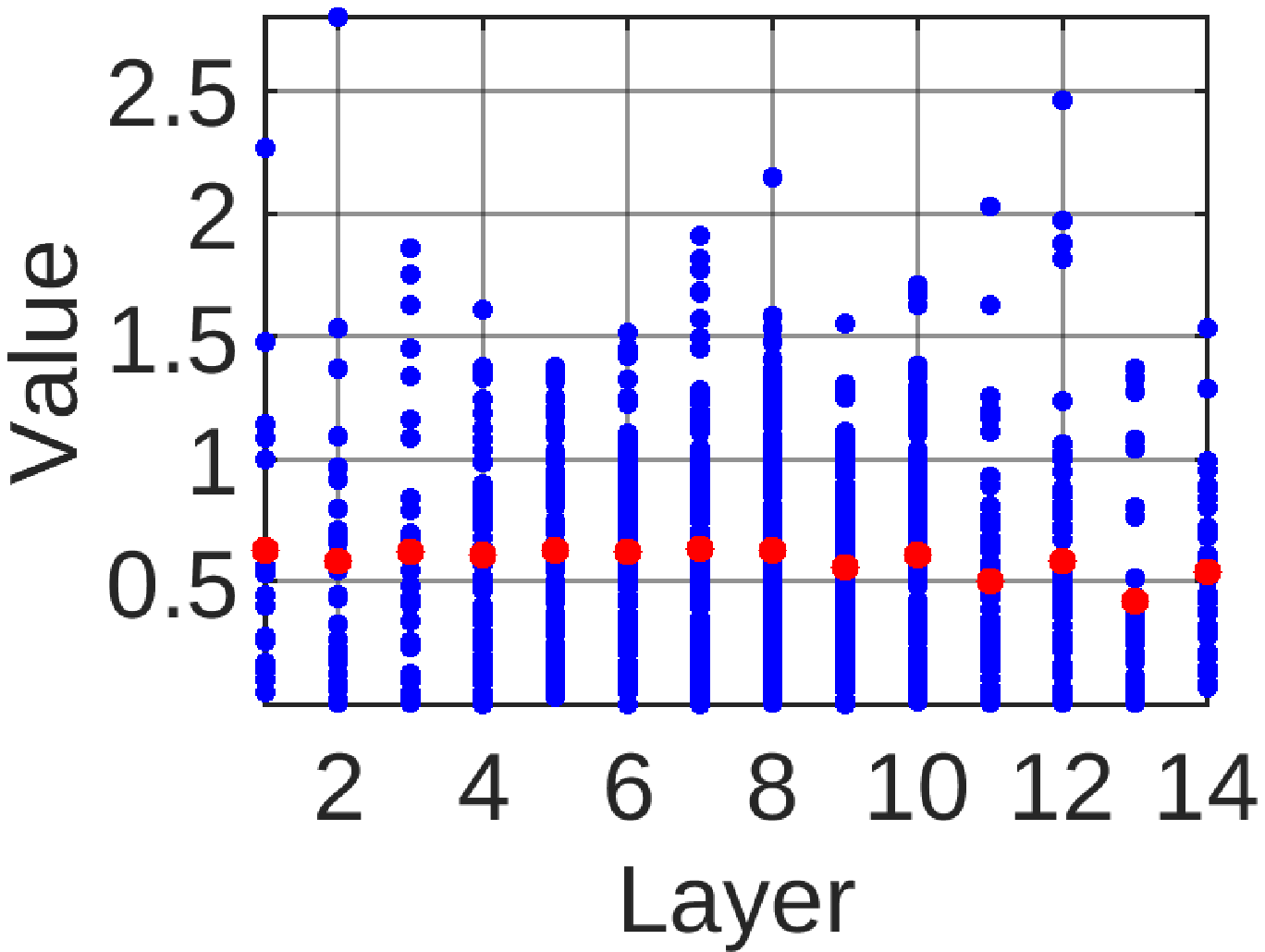}
  \caption{\scriptsize{BraTS'18, RANP-f Eq.~\myblue{10}}}
  \end{subfigure}
  \begin{subfigure}[b]{0.23\textwidth}
  \centering
  \includegraphics[width=\textwidth]{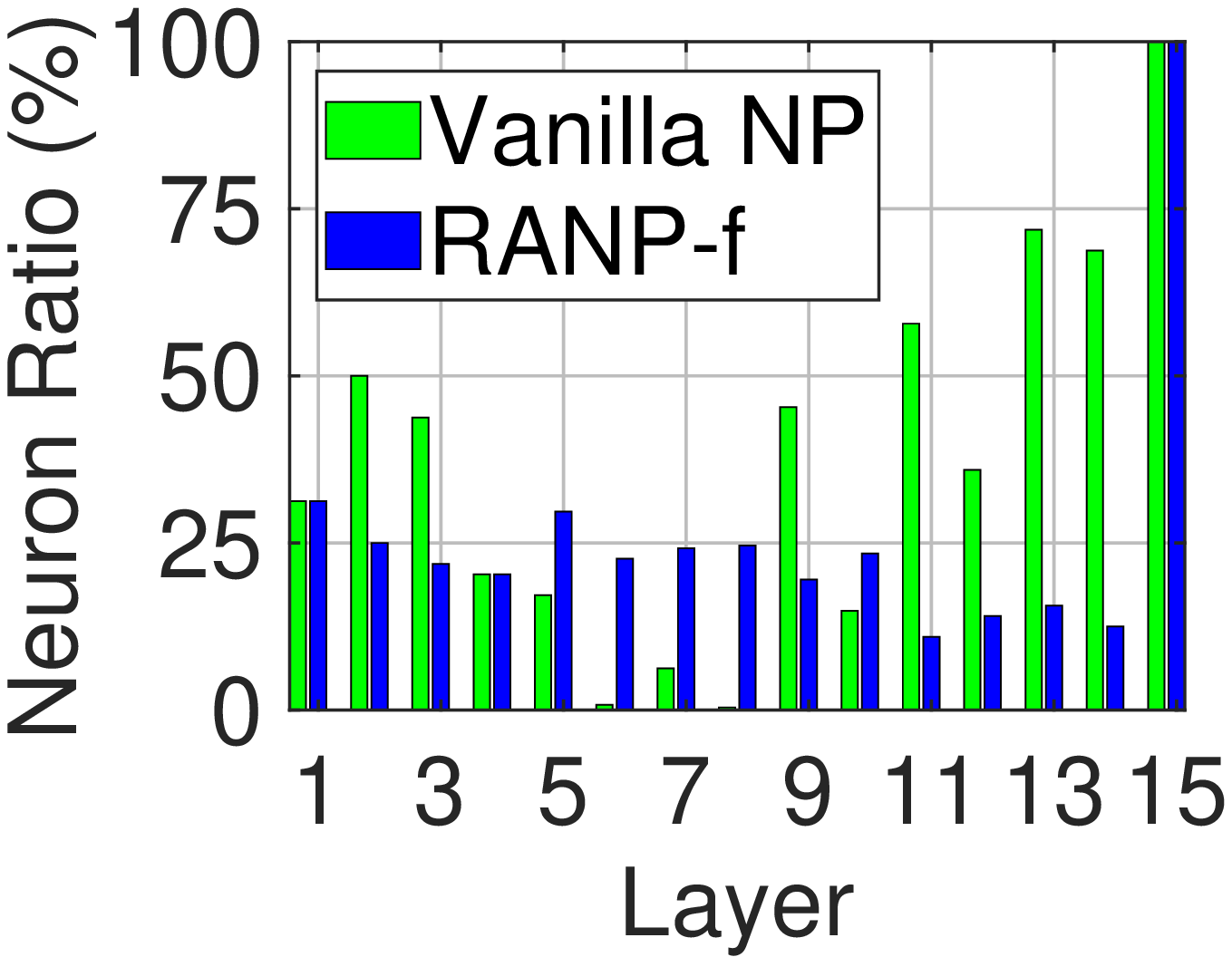}
  \caption{\scriptsize{BraTS'18, Retain ratio}}
  \end{subfigure}
\caption{Balanced neuron importance distribution by MPMG-sum on ShapeNet and BraTS'18.
Neuron sparsity is 78.24\% on ShapeNet and 78.17\% on BraTS'18.
Blue: neuron values; red: mean values.}
\label{fig:neuron_distribution_mpmg}
\end{center}
\vspace{-5mm}
\end{figure*}

\begin{figure*}[!ht]
\begin{center}
  \begin{subfigure}[b]{0.4\textwidth}
  \centering
  \includegraphics[width=\textwidth]{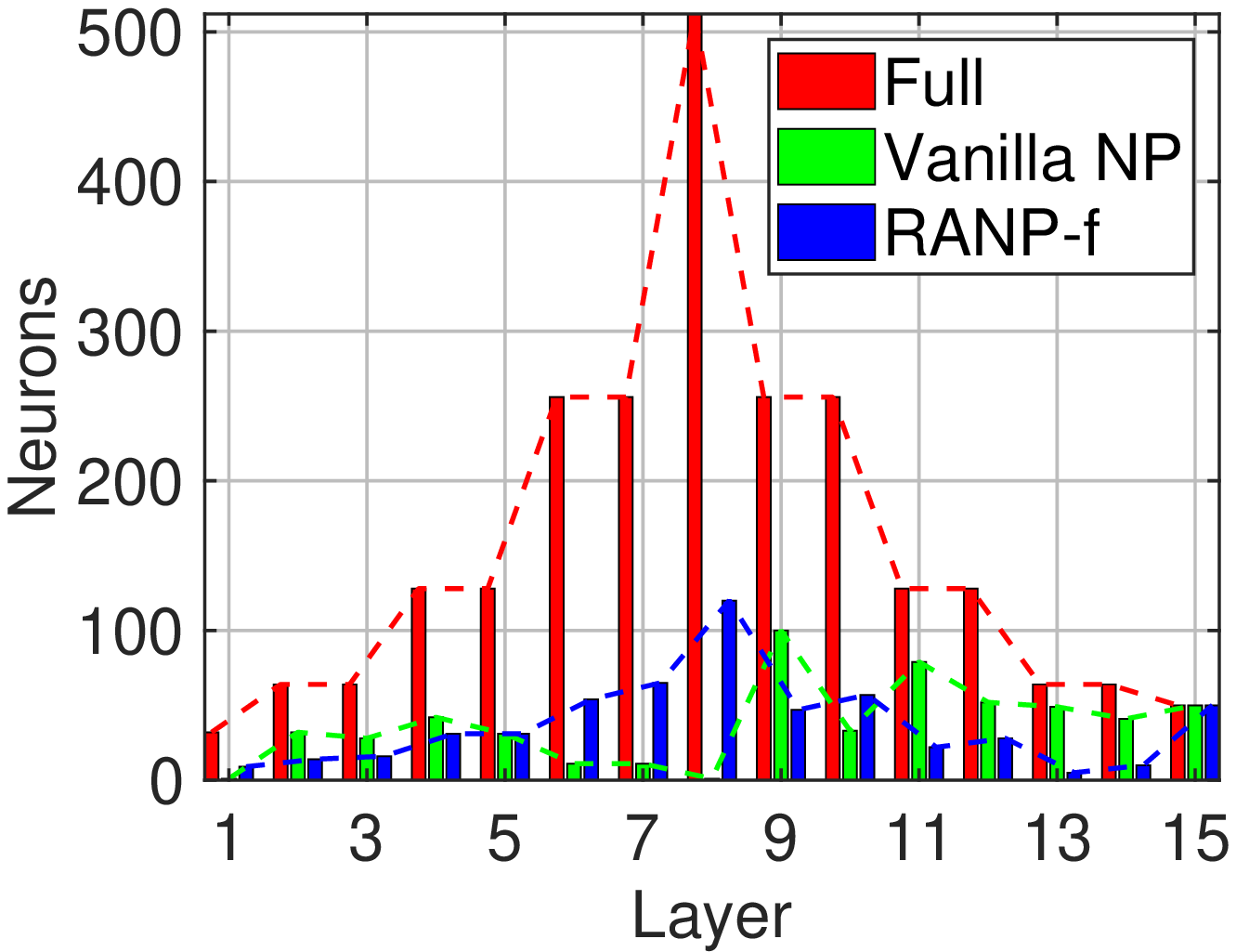}
  \caption{ShapeNet}
  \end{subfigure}
  ~
  \begin{subfigure}[b]{0.4\textwidth}
  \centering
  \includegraphics[width=\textwidth]{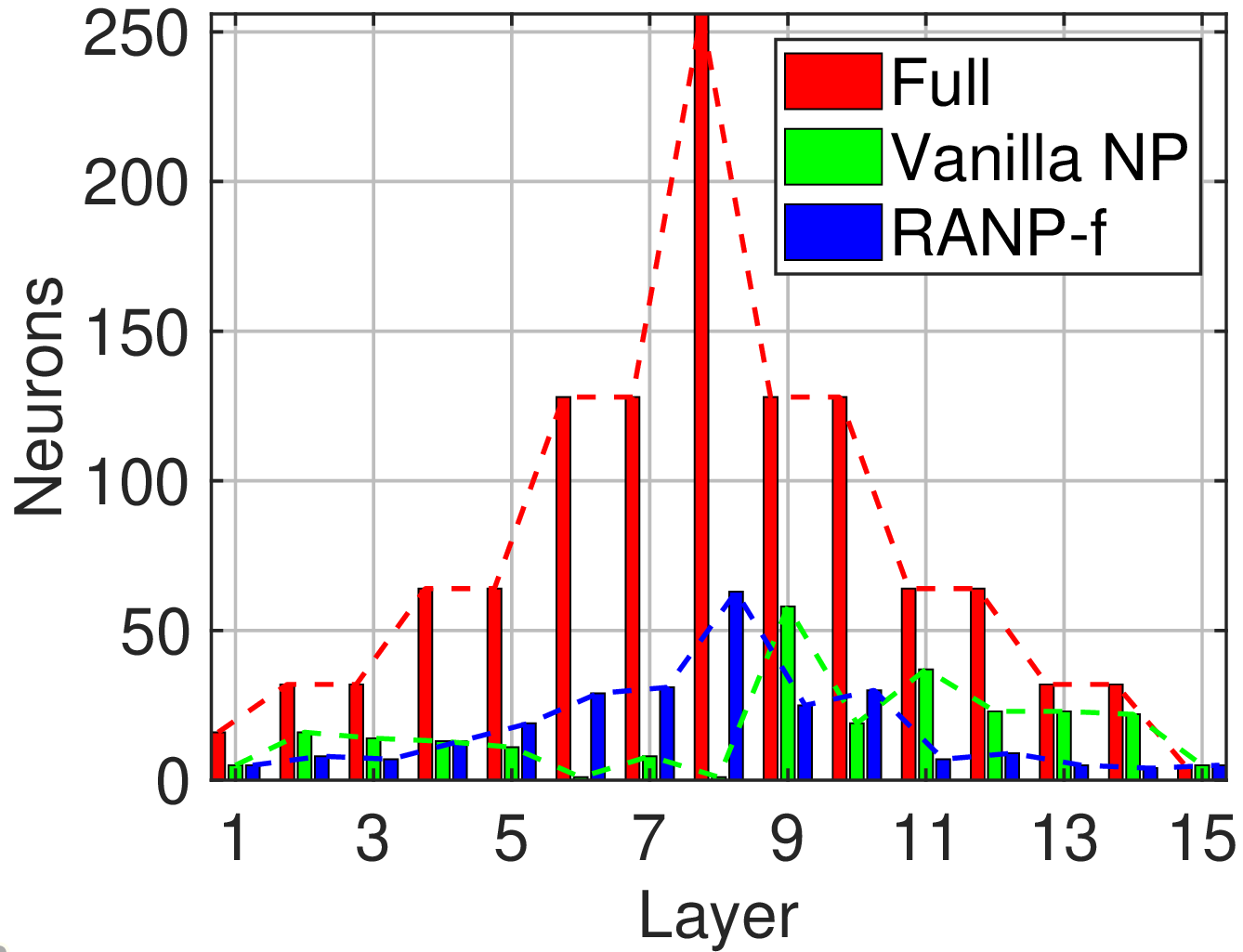}
  \caption{BraTS'18}
  \end{subfigure}
\vspace{-1mm}
\caption{Layer-wise neuron distribution of 3D-UNets.}
\label{fig:neuron_per_layer}
\end{center}
\vspace{-5mm}
\end{figure*}

In Fig.~\ref{fig:neuron_distribution_mnmg}, neuron importance by MPMG-sum is more balanced than by MNMG-sum, which avoids pruned networks by MPMG-sum to be infeasible, that is at least 1 neuron will be retained in each layer.

In addition to the distribution of retained neuron ratios in Fig.~\myblue{2} in the main paper for ShapeNet, which is also shown in the first row of Fig.~\ref{fig:neuron_distribution_mpmg}, the last row of Fig.~\myblue{\ref{fig:neuron_distribution_mpmg}} is for BraTS'18.
Moreover, Fig.~\myblue{\ref{fig:neuron_per_layer}} illustrates the distribution of neurons retained in each layer by vanilla neuron pruning (\ie, vanilla NP) and RANP-f compare to the full network.

Clearly, upon pruning, neurons in each layer are largely reduced except the last layer where all neurons are retained for the number of segmentation classes.
In Fig.~\myblue{\ref{fig:neuron_per_layer}}, vanilla NP has very few neurons in, \eg, the 8th layer, resulting in low accuracy or network infeasibility.
By contrast, the neuron distribution by RANP-f is more balanced to improve the pruning capability.
\else
    \thispagestyle{empty}
\begin{abstract}
Although 3D Convolutional Neural Networks (CNNs) are essential for most learning based applications involving dense 3D data, their applicability is limited due to excessive memory and computational requirements.
Compressing such networks by pruning therefore becomes highly desirable.
However, pruning 3D CNNs is largely unexplored possibly because of the complex nature of typical pruning algorithms that embeds pruning into an iterative optimization paradigm.
In this work, we introduce a Resource Aware Neuron Pruning (RANP) algorithm that prunes 3D CNNs at initialization to high sparsity levels.
Specifically, the core idea is to obtain an importance score for each neuron based on their sensitivity to the loss function.
This neuron importance is then reweighted according to the neuron resource consumption related to FLOPs or memory.
We demonstrate the effectiveness of our pruning method on 3D semantic segmentation with widely used 3D-UNets on ShapeNet and BraTS'18 as well as on video classification with MobileNetV2 and I3D on UCF101 dataset.
In these experiments, our RANP leads to roughly {\bf 50\%-95\% reduction in FLOPs and 35\%-80\% reduction in memory} with negligible loss in accuracy compared to the unpruned networks.
This significantly reduces the computational resources required to train 3D CNNs.
The pruned network obtained by our algorithm can also be easily scaled up and transferred to another dataset for training.
\end{abstract}

\vspace{-5mm}



\section{Introduction} \label{sec:introduction}

\begin{figure}[!t]
\begin{center}
\centering
\includegraphics[width=0.48\textwidth]{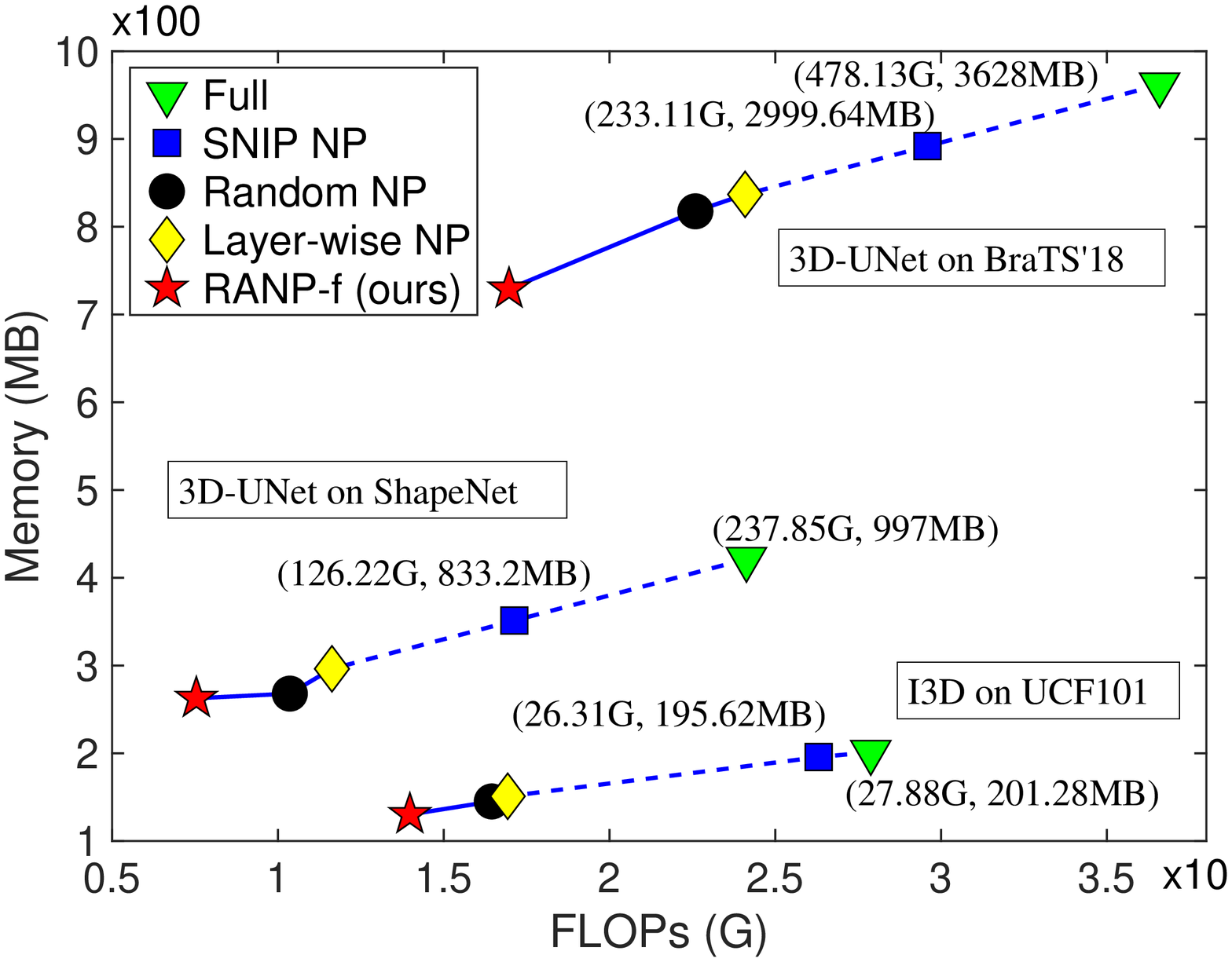}
\caption{\textbf{Bottom-left is the best}. Comparison of neuron pruning methods.
``Full" and ``SNIP NP" are not drawn by scale but with their FLOPs (G) and memory (MB) values next to the markers.
Our RANP-f performs best with large resource reductions while maintaining the accuracy.
More details are in Table~\myblue{\ref{tb:full_table}}.
}
\label{fig:fig_all}
\end{center}
\vspace{-10mm}
\end{figure}

3D image analysis is important in various real-world applications including scene understanding~\cite{video_seg,3dcnn_ref1}, object recognition~\cite{human_seg,3dcnn_ref2}, medical image analysis~\cite{3dunet,3dcnn_ref3,3dcnn_ref4}, and video action recognition~\cite{video_class1,video_class2}.
Typically, sparse 3D data can be represented using point clouds~\cite{shapenet} whereas volumetric representation is required for dense 3D data which arises in domains such as medical imaging~\cite{brats1} and video segmentation and classification~\cite{video_seg,video_class1,video_class2}.
While efficient neural network architectures can be designed for sparse point cloud data~\cite{ssc,pointnet}, conventional dense 3D Convolutional Neural Network (CNN) is required for volumetric data.
Such 3D CNNs are computationally expensive with excessive memory requirements for large-scale 3D tasks.
Therefore, it is highly desirable to reduce the memory and FLOPs required to train 3D CNNs while maintaining the accuracy.
This will not only enable large-scale applications but also 3D CNN training on resource-limited devices.

Network pruning is a prominent approach to compress a neural network by reducing the number of parameters or the number of neurons in each layer~\cite{weight_pruning_1,weight_pruning_2,prune_2,prune_3}.
However, most of the network pruning methods aim at 2D CNNs while pruning 3D CNNs is largely unexplored.
This is mainly because pruning is typically targeted at reducing the test-time resource requirements while computational requirements of training time are as large as (if not more than) the unpruned network.
Such pruning schemes are not suitable for 3D CNNs with dense volumetric data where training-time resource requirement is prohibitively large.

In this work, we introduce a Resource\footnote{We concretely define ``resource'' as FLoating Point Operations per second (FLOPs) and memory required by one forward pass.} Aware Neuron Pruning (RANP) 
that {\em prunes 3D CNNs at initialization}.
Our method is inspired by, but superior to, SNIP~\cite{snip} which prunes redundant parameters of a network at initialization and only tests with small scale 2D CNNs for image classification.
With the same characteristics of effectively pruning at initialization without requiring large computational resources, RANP yields better-pruned networks compared to SNIP by removing neurons that largely contribute to the high resource requirement.  
In our experiments on video classification and more challenging 3D semantic segmentation, with minimal accuracy loss, RANP yields 50\%-95\% reduction in FLOPs and 35\%-80\% reduction in memory while only 5\%-51\% reduction in FLOPs and 1\%-17\% reduction in memory are achieved by SNIP NP.

The main idea of RANP is to prune based on a {\em neuron importance} criterion analogous to the connection sensitivity in SNIP.
Note that, pruning based on such a simple criterion as SNIP has the risk of pruning the whole layer(s) at extreme sparsity levels especially on large networks~\cite{signal_propagation}.
Even though an orthogonal initialization that ensures layer-wise dynamical isometry is sufficient to mitigate this issue for parameter pruning on 2D CNNs~\cite{signal_propagation}, it is unclear if this could be directly applied to neuron pruning on 3D CNNs.
To tackle this and improve pruning, we introduce a {\em resource aware reweighting scheme} that first balances the mean value of neuron importance in each layer and then reweights the neuron importance based on the resource consumption of each neuron. 
As evidenced by our experiments, such a reweighting scheme is crucial to obtain large reductions in memory and FLOPs while maintaining high accuracy.

We firstly evaluate our RANP on 3D semantic segmetation on a sparse point-cloud dataset, ShapeNet~\cite{shapenet}, and a dense medical image dataset, BraTS'18~\cite{brats1,brats2}, with widely used 3D-UNets~\cite{3dunet}.
We also evaluate RANP on video classification using UCF101 with MobileNetV2~\cite{mobilenetv2} and I3D~\cite{i3d}.
Our RANP-f significantly outperforms 
other neuron pruning methods in \textit{resource efficiency} by yielding large reductions in computational resources ({\bf 50\%-95\% FLOPs reduction and 35\%-80\% memory reduction}) with comparable accuracy to the unpruned network (Fig.~\myblue{\ref{fig:fig_all}}).

Furthermore, we perform extensive experiments to demonstrate \textbf{1) scalability} of RANP by pruning with a small input spatial size and training with a large one, \textbf{2) transferability} by pruning using ShapeNet and training on BraTS'18 and vice versa, \textbf{3) lightweight} training on a single GPU, and \textbf{4) fast} training with increased batch size.

\section{Related Work} \label{sec:related_work}

Previous works of network pruning mainly focus on 2D CNNs by parameter pruning \cite{snip,prune_3,weight_pruning_1,weight_pruning_2,eccv_w_pruning_1} and neuron pruning \cite{group_wise,prune_2,filter_1,nisp,eccv_f_pruning_1,eccv_f_pruning_2}.
While a majority of the pruning methods use the traditional prune-retrain scheme with a combined loss function of pruning criteria \cite{prune_3,prune_2,nisp}, some pruning at initialization methods is able to reduce computational complexity in training \cite{snip,single_shot_1,single_shot_2,single_shot_3,ticket}.
While very few are for 3D CNNs \cite{resource_efficient,3d_pruning_unclear,frequency_domain}, none of them prune networks at initialization, and thus, none of them effectively reduce the training-time computational and memory requirements of 3D CNNs.

\textbf{2D CNN pruning.}
\textit{Parameter pruning} merely sparsifies filters for a high learning capability with small models.
Han~\etl~\cite{prune_3} adopted an iterative method of removing parameters with values below a threshold.
Lee~\etl~\cite{snip} recently proposed a single-shot method with connection sensitivity by magnitudes of parameter mask gradients to retain top-$\kappa$ parameters.
These filter-sparse methods, however, do not directly yield large speedup and memory reductions.

By contrast, \textit{neuron pruning}, also known as filter pruning or channel pruning, can effectively reduce computational resources.
For instance, Li \etl \cite{filter_1} used $l_1$ normalization to remove unimportant filters with connecting features.
He \etl \cite{prune_2} adopted a LASSO regression to prune network layers with reconstruction in the least square manner.
Yu \etl \cite{group_wise} proposed a group-wise 2D-filter pruning from each 3D-filter by a learning-based method and a knowledge distillation.
Structure learning based MorphNet \cite{morphnet} and SSL \cite{ssl} aim at pruning activations with structure constraints or regularization.
These approaches only reduce the test-time resource requirement while we focus on reducing those of large 3D CNNs at training time.

\textbf{3D CNN pruning.}
To improve the efficiency on 3D CNNs, some works like SSC \cite{ssc} and OctNet \cite{octree} use efficient data structures to reduce the memory requirement for sparse point-cloud data.
However, these approaches are not useful for dense data, \eg, MRI images and videos, and the resource requirement remains prohibitively large.

Hence, it is desirable to develop an efficient pruning for 3D CNNs that can handle dense 3D data which is common in real applications.
Only very few works are relevant to 3D CNN pruning.
Molchanov \etl \cite{resource_efficient} proposed a greedy criteria-based method to reduce resources via backpropagation with a small 3D CNN for hand gesture recognition.
Zhang \etl \cite{3d_pruning_unclear} used a regularization-based pruning method by assigning regularization to weight groups with 4$\times$ speedup in theory.
Recently, Chen \etl \cite{frequency_domain} converted 3D filters into frequency domain to eliminate redundancy in an iterative optimization for convergence.
Being a parameter pruning method, this does not lead to large FLOPs and memory reductions, \eg, merely a $2\times$ speedup compared to our $28\times$ (ref. Sec.~\myblue{\ref{subsec:pruning_capability}}).
In summary, these methods embed pruning in the iterative network optimization and require extensive resources, which is inefficient for 3D CNNs.

\textbf{Pruning at Initialization.}
While few works adopted pruning at initialization, some achieved impressive success.
SNIP \cite{snip} is the first single-shot pruning method that presented a high possibility of pruning networks at initialization with minimal accuracy loss in training, followed by many recent works on single-shot pruning \cite{single_shot_1,single_shot_2,single_shot_3,ticket}.
But none are for 3D CNNs pruning.

In addition to being a parameter pruning approach, the benefits of SNIP was demonstrated only on small-scale datasets \cite{snip}, such as MNIST and CIFAR-10.
Therefore, it is unclear that whether these benefits could be transposed to 3D CNNs applied to large-scale datasets.
Our experiments indicate that, while SNIP itself is not capable of yielding large resource reduction on 3D CNNs, our RANP can greatly reduce the computational resources without causing network infeasibility.
Furthermore, we show that RANP enjoys strong transferability among datasets and enables fast and lightweight training of large 3D volumetric data segmentation on a single GPU.

\section{Preliminaries}\label{sec:preliminaries}
We first briefly describe the main idea of SNIP~\cite{snip} which removes redundant parameters prior to training. 
Given a dataset $\mathcal{D}=\{(\mathbf{x}_i, \mathbf{y}_i)\}_{i=1}^S$ with input $\mathbf{x}_i$ and ground truth $\mathbf{y}_i$ and the sparsity level $\kappa$, the optimization problem associated with SNIP can be written as

\vspace{-5mm}
\begin{equation}
\begin{aligned}
\min_{\mathbf{c}, \mathbf{w}} L(\mathbf{c} \odot \mathbf{w}; \mathcal{D}) &= \min_{\mathbf{c}, \mathbf{w}} \frac{1}{S} \sum^{S}_{i=1} \ell\left(\mathbf{c} \odot \mathbf{w}, (\mathbf{x}_i, \mathbf{y}_i)\right)\ , \\
\text{s.t.} \quad \mathbf{w} \in \mathbb{R}^m&, \enskip \mathbf{c} \in \{0,1\}^m, \enskip \| \mathbf{c} \|_{0} \leq \kappa\ ,
\end{aligned}
\end{equation}
where $\mathbf{w}$ is denoted a $m$-dimensional vector of parameters, $\mathbf{c}$ is the corresponding binary mask on the parameters, $\ell(\cdot)$ is the standard loss function (\eg, cross-entropy loss), and $\|\cdot\|_0$ denotes $l_0$ norm.
The mask $c_{j}\in \{0,1\}$ for parameter $w_{j}$ denotes that the parameter is retained in the compressed model if $c_{j}=1$ and otherwise it is removed. 
In order to optimize the above problem, they first relax the binary constraint on the masks such that $\mathbf{c}\in [0,1]^m$.
Then an importance function for parameter $w_j$ is calculated by the normalized magnitude of the loss gradient over mask $c_j$ as
\begin{equation}
\label{eq:gradient_mask}
s_j = \frac{\left|g_j\left(\mathbf{w}; \mathcal{D}\right)\right|}{\sum _{k=1}^m \left|g_k\left(\mathbf{w}; \mathcal{D}\right)\right|}\ ,\mbox{where $g_j\left(\mathbf{w}; \mathcal{D}\right) = \left.\frac{\partial L\left(\mathbf{c} \odot \mathbf{w}; \mathcal{D}\right)}{\partial c_j}\right|_{\mathbf{c}=\mathbf{1}}$}\ .
\end{equation}
Then, only top-$\kappa$ parameters are retained based on the parameter importance, called connection sensitivity in \cite{snip}, $\mathbf{s}$ defined above. Upon pruning, the retained parameters are trained in the standard way.
It is interesting to note that, even though having the mask $\mathbf{c}$ is easier to explain the intuition, SNIP can be implemented without these additional variables by noting that $g_j\left(\mathbf{w}; \mathcal{D}\right) = \left(\partial L\left(\mathbf{w}; \mathcal{D}\right)/ \partial w_j\right)w_j$~\cite{signal_propagation}.
This method has shown remarkable results in achieving $ > 95\%$ sparsity on 2D image classification tasks with minimal loss of accuracy.
Such a parameter pruning method is important, however, it cannot lead to sufficient computation and memory reductions to train a deep 3D CNN on current off-the-shelf graphics hardware.  
In particular, the sparse weight matrices cannot efficiently reduce memory or FLOPs, and they require specialized sparse matrix implementations for speedup.
In contrast, neuron pruning directly translates into practical gains of reducing both memory and FLOPs.
This is crucial in 3D CNNs due to their substantially higher resource requirement compared to 2D CNNs.

\vspace{-1mm}
\section{Resource Aware NP at Initialization} \label{sec:method}
\vspace{-2mm}
\begin{figure}[t]
\centering
\includegraphics[width=0.45\textwidth]{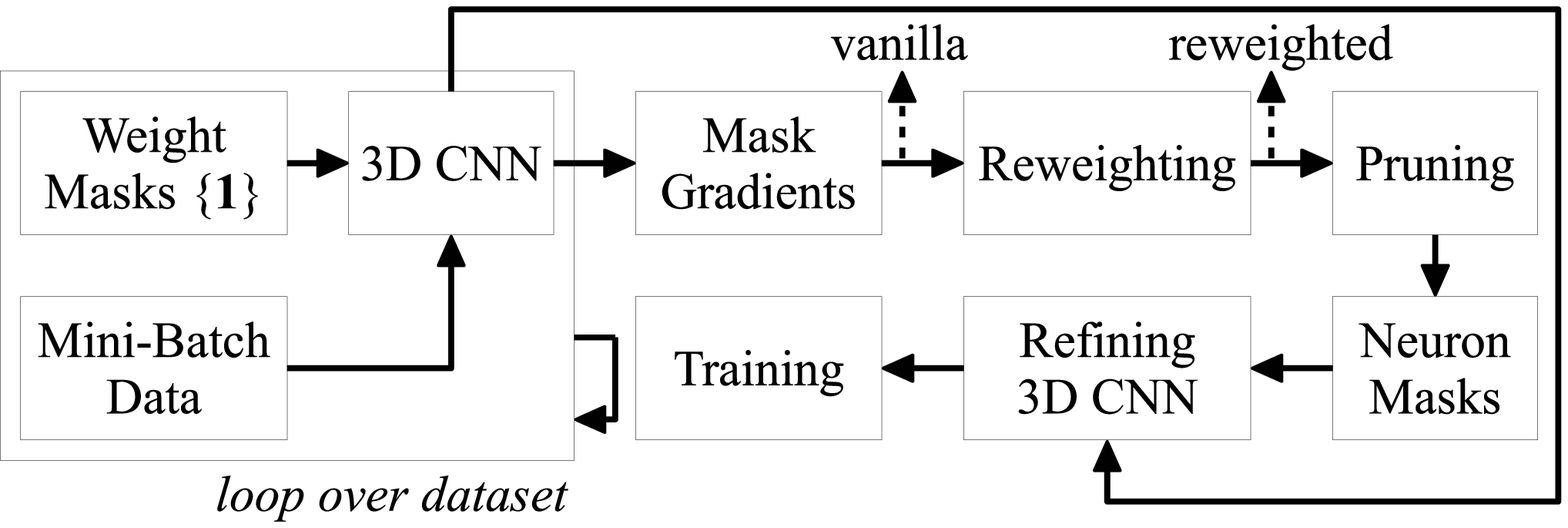}
\vspace{-3mm}
\caption{Flowchart of RANP algorithm. The refining generates a new yet slim network for resource-efficient training.}
\label{fig:flowchart}
\vspace{-5mm}
\end{figure}

To explain the proposed RANP, we first extend the SNIP idea to neuron pruning at initialization. 
Then we discuss a resource aware reweighting strategy to further reduce the computational requirements of the pruned network.
The flowchart of our RANP algorithm is shown in Fig.~\myblue{\ref{fig:flowchart}}.

Before introducing our neuron importance, we first consider a fully-connected feed-forward neural network for the simplicity of notations.
Consider weight matrices ${\mathbf{W}^l\in\mathbb{R}^{N_l \times N_{l-1}}}$, biases $\mathbf{b}^l\in\mathbb{R}^{N_{l}}$, pre-activations $\mathbf{h}^l\in\mathbb{R}^{N_l}$, and post-activations $\mathbf{x}^l\in\mathbb{R}^{N_l}$, for layer ${l\in \mathcal{K}=\{1,\ldots,K\}}$.
Now the feed-forward dynamics is

\vspace{-4mm}
\begin{equation}\label{eq:feed-forward}
\mathbf{x}^l = \phi\left(\mathbf{h}^l\right)\ ,\quad \mbox{where $\mathbf{h}^l = \mathbf{W}^l \mathbf{x}^{l-1} + \mathbf{b}^l$}\ ,
\vspace{-1mm}
\end{equation}
where the activation function $\phi:\mathbb{R} \to \mathbb{R}$ has elementwise nonlinearity and the network input is denoted by $\mathbf{x}^0$.
Now we introduce binary masks on neurons (\ie, post-activations).
The feed-forward dynamics is then modified to include this masking operation as
\vspace{-0.5em}
\begin{equation}\label{eq:feed-forward-mask}
\mathbf{x}^l = \mathbf{c}^l \odot \phi\left(\mathbf{h}^l\right)\ ,\quad \mbox{where $\mathbf{c}^l\in\{0,1\}^{N_{l}}$}\ , \enskip \forall l \in \mathcal{K}\ ,
\vspace{-0.2em}
\end{equation}
where neuron mask $c^l_u = 1$ indicates neuron $x_u^l$ is retained and otherwise pruned. 
Here, neuron pruning can be written as the following optimization problem
\vspace{-1mm}
\begin{equation}
\begin{aligned}
\min_{\mathbf{w}}L(\mathbf{c}, \mathbf{w};\mathcal{D}) &= \min_{\mathbf{c}, \mathbf{w}} \frac{1}{S} \sum^{S}_{i=1} \ell\left(\mathbf{c}, \mathbf{w}; (\mathbf{x}_i, \mathbf{y}_i)\right)\ , \\
\text{s.t.} \quad \mathbf{w} \in \mathbb{R}^m &, \enskip \mathbf{c} \in \{0,1\}^n, \enskip \| \mathbf{c} \|_{0} \leq \kappa\ ,
\end{aligned}
\end{equation}
where $n$ is the total number of neurons and $\ell(\mathbf{c}, \cdot;\cdot)$ denotes a standard loss function of the feed-forward mapping with neuron masks $\mathbf{c}$ defined in Eq.~\myblue{\ref{eq:feed-forward-mask}}.
This can be easily extended to convolutional and skip-concatenation operations.

As removing neurons could largely reduce memory and FLOPs compared to merely sparsifying parameters, the core of our approach is benefited by removing redundant neurons from the model.
We use an influence function concept developed for parameters to establish neuron importance through the loss function, to locate redundant neurons.

\subsection{Neuron Importance} \label{sec:vanilla_neuron_importance}
Note that, neuron importance can be derived from the SNIP-based parameter importance discussed in Sec.~\myblue{\ref{sec:preliminaries}}. 
Another approach is to directly define neuron importance as the normalized magnitude of the neuron mask gradients analogous to parameter importance.

\textbf{Neuron Importance with Parameter Mask Gradients.}
\label{sec:vanilla_importance_parameter_mask}
The first approach to calculate neuron importance depends on the magnitude of parameter mask gradients, denoted as Magnitude of Parameter Mask Gradients (MPMG). 
Thus, the importance of neuron $x^l_u$ is
\vspace{-2mm}
\begin{equation}
\label{eq:neuron_importance_parameter_mask}
s^l_u = f \left(|g^l_{u1}|,\ldots, |g^l_{uN_{l-1}}|\right)\ ,
\end{equation}
where $g^l_{uv} = \partial L\left(\mathbf{c} \odot \mathbf{w}; \mathcal{D}\right) / \partial c^l_{uv}$ with $c^l_{uv}$ as the mask of parameter $w^l_{uv}$.
Refer to Eq.~\myblue{\ref{eq:gradient_mask}}.
Here, $f(\cdot):\mathbb{R}^{N_{l-1}}\to\mathbb{R}$ is a function mapping a set of values to a scalar.
We choose $f(\cdot)=\text{sum}(\cdot)$ with alternatives, \ie, mean and max functions, in Appendix~\myref{D}.
Now, we set neuron masks as 1 for neurons with top-$\kappa$ largest neuron importance.

\textbf{Neuron Importance with Neuron Mask Gradients.} \label{sec:vanilla_importance_neuron_mask}
Another approach is to directly compute mask gradients on neurons and treat their magnitudes as neuron importance, denoted as Magnitude of Neuron Mask Gradients (MNMG).
The neuron importance of $x^l_u$ is calculated by

\begin{equation}
\label{eq:gradient_neuron}
s^l_u = \left|\left.\frac{\partial L\left(\mathbf{c}, \mathbf{w}; \mathcal{D}\right)}{\partial c^l_u}\right|_{\mathbf{c} = \mathbf{1}}\right|\ .
\end{equation}
Noting that a non-linear activation function $\phi(\cdot)$ in CNN including but not limited to ReLU can satisfy $\phi(c h) = c \phi(h), \forall c \geq 0$.
Given such a homogeneous function, the calculation of neuron importance with neuron masks can be derived from parameter mask gradients in the form of

\vspace{-5mm}
\begin{equation}
    \left.\frac{\partial L\left(\mathbf{c}, \mathbf{w}; \mathcal{D}\right)}{\partial c^l_u}\right|_{\mathbf{c} = \mathbf{1}}
    = \sum_{v=1}^{N_{l-1}} \left.\frac{\partial L\left(\mathbf{c} \odot \mathbf{w}; \mathcal{D}\right)}{\partial c^l_{uv}}\right|_{\mathbf{c} = \mathbf{1}}\ .
\end{equation}

Details of the influence of such an activation function on neuron importance are provided in Appendix~\myref{B}.
These two approaches for neuron importance are in a similar form that while MPMG is by the sum of magnitudes, MNMG is by the magnitude of the sum of parameter mask gradients.
It can be implemented directly from parameter gradients.

The neuron importance based on MPMG or MNMG approach can be used to remove redundant neurons. 
However, they could lead to an imbalance of sparsity levels of each layer in 3D network architectures. 
As shown in Table~\myblue{\ref{tb:full_table}}, the computational resources required by vanilla neuron pruning are much higher than those by other sparsity enforcing methods, \eg, random neuron pruning and layer-wise neuron pruning. 
We hypothesize that this is caused by the layer-wise imbalance of neuron importance which unilaterally emphasizes on some specific layer(s) and may lead to network infeasibility by pruning the whole layer(s).
This behavior is also observed in \cite{signal_propagation}, and orthogonal initialization is thus recommended to solve the problem for 2D CNN pruning, which however cannot result in balanced neuron importance in our case, see results in Appendix~\myref{D}.

In order to resolve this issue, we propose resource aware neuron pruning (RANP) with reweighted neuron importance, and the details are provided below.

\subsection{Resource Aware Reweighting} \label{sec:neuron_importance_reweighted}

\begin{figure}[t]
\begin{center}
  \begin{subfigure}[b]{0.2\textwidth}
  \centering
  \includegraphics[width=\textwidth]{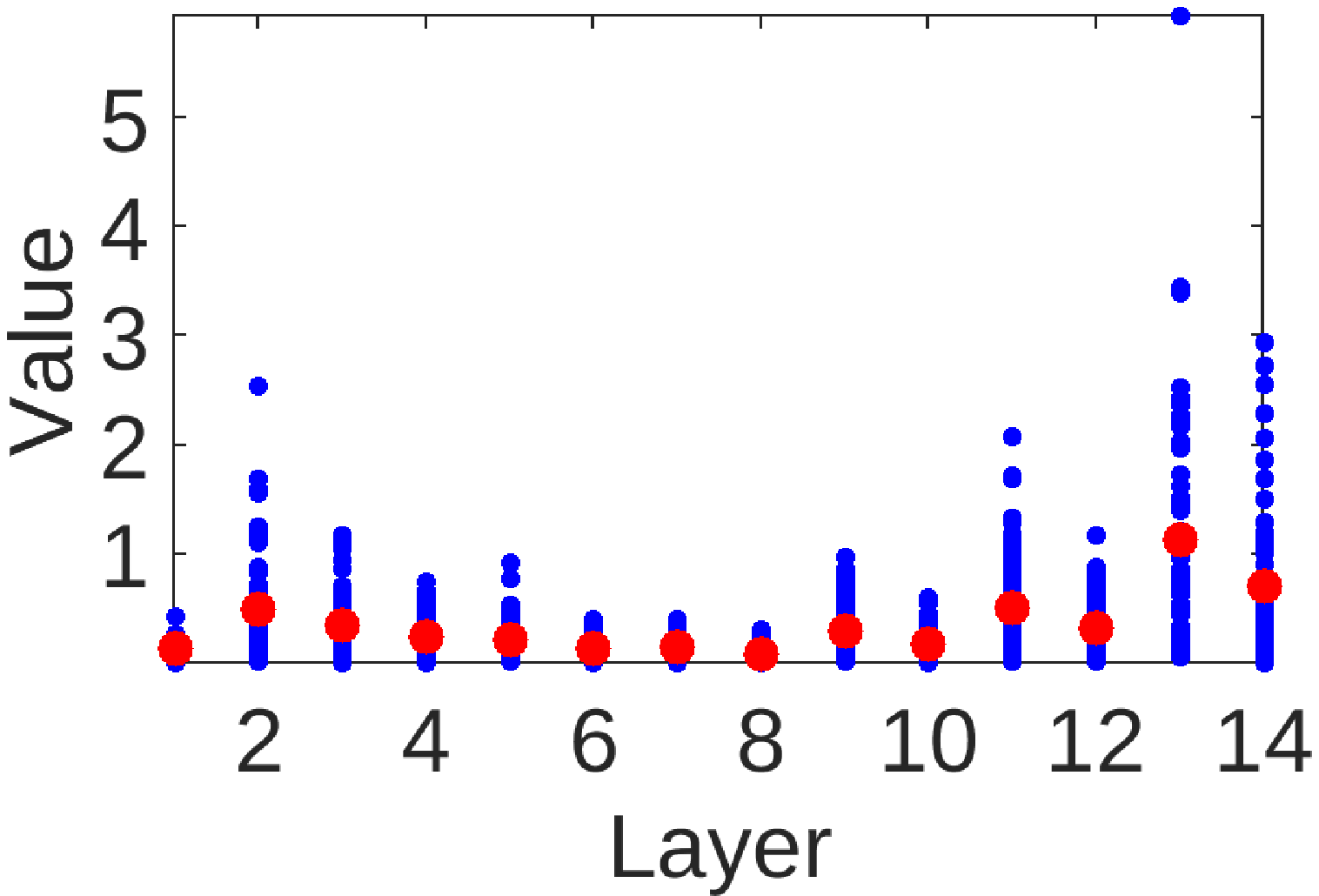}
  \caption{Vanilla NP Eq.~\myblue{\ref{eq:neuron_importance_parameter_mask}}}
  \label{fig:shapenet_org}
  \end{subfigure}
  \hspace{3mm}
  \begin{subfigure}[b]{0.2\textwidth}
  \centering
  \includegraphics[width=\textwidth]{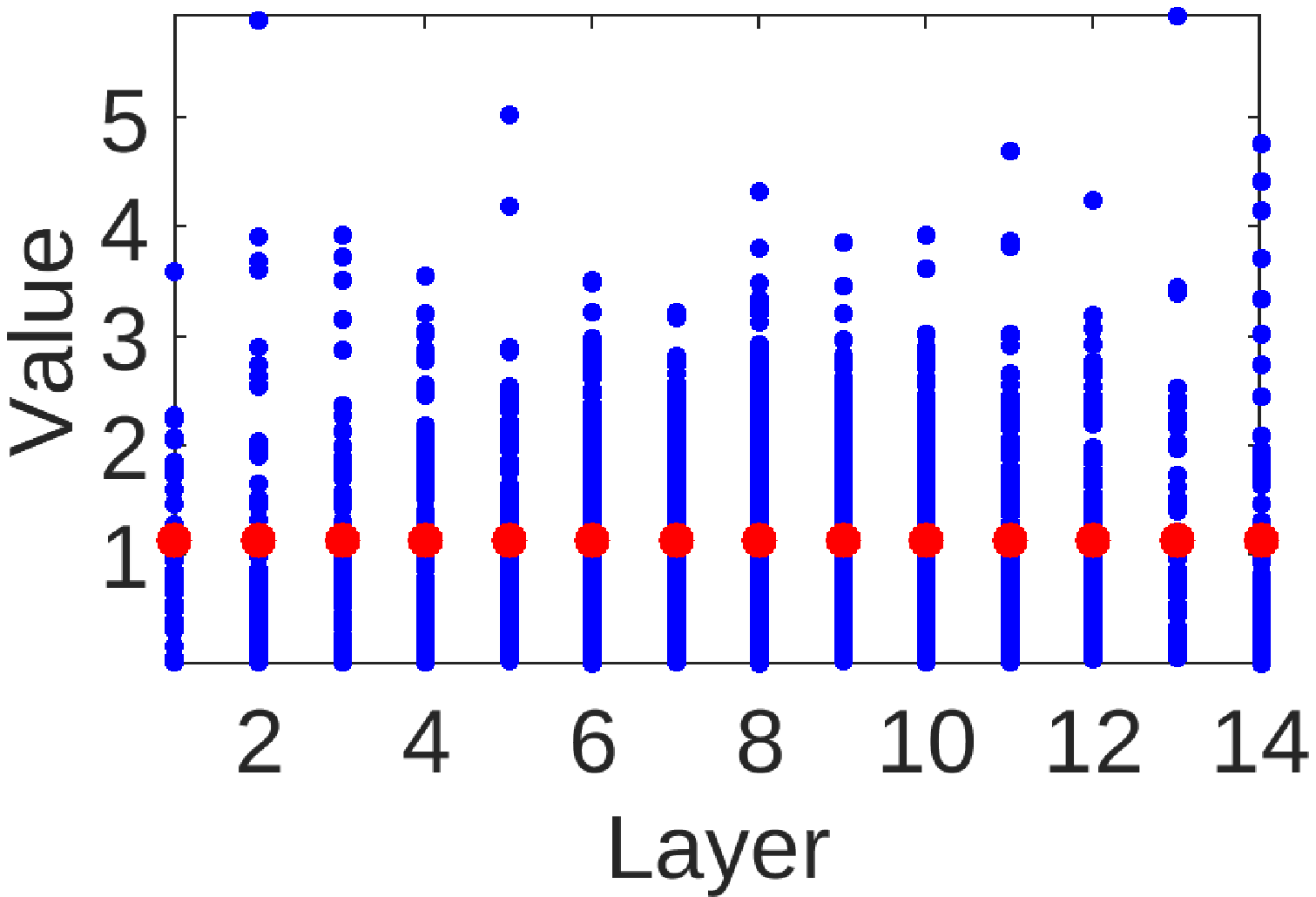}
  \caption{Weighted NP Eq.~\myblue{\ref{eq:neuron_importance_weighted}}}
  \label{fig:shapenet_grads}
  \end{subfigure}
  \\
  \begin{subfigure}[b]{0.2\textwidth}
  \centering
  \includegraphics[width=\textwidth]{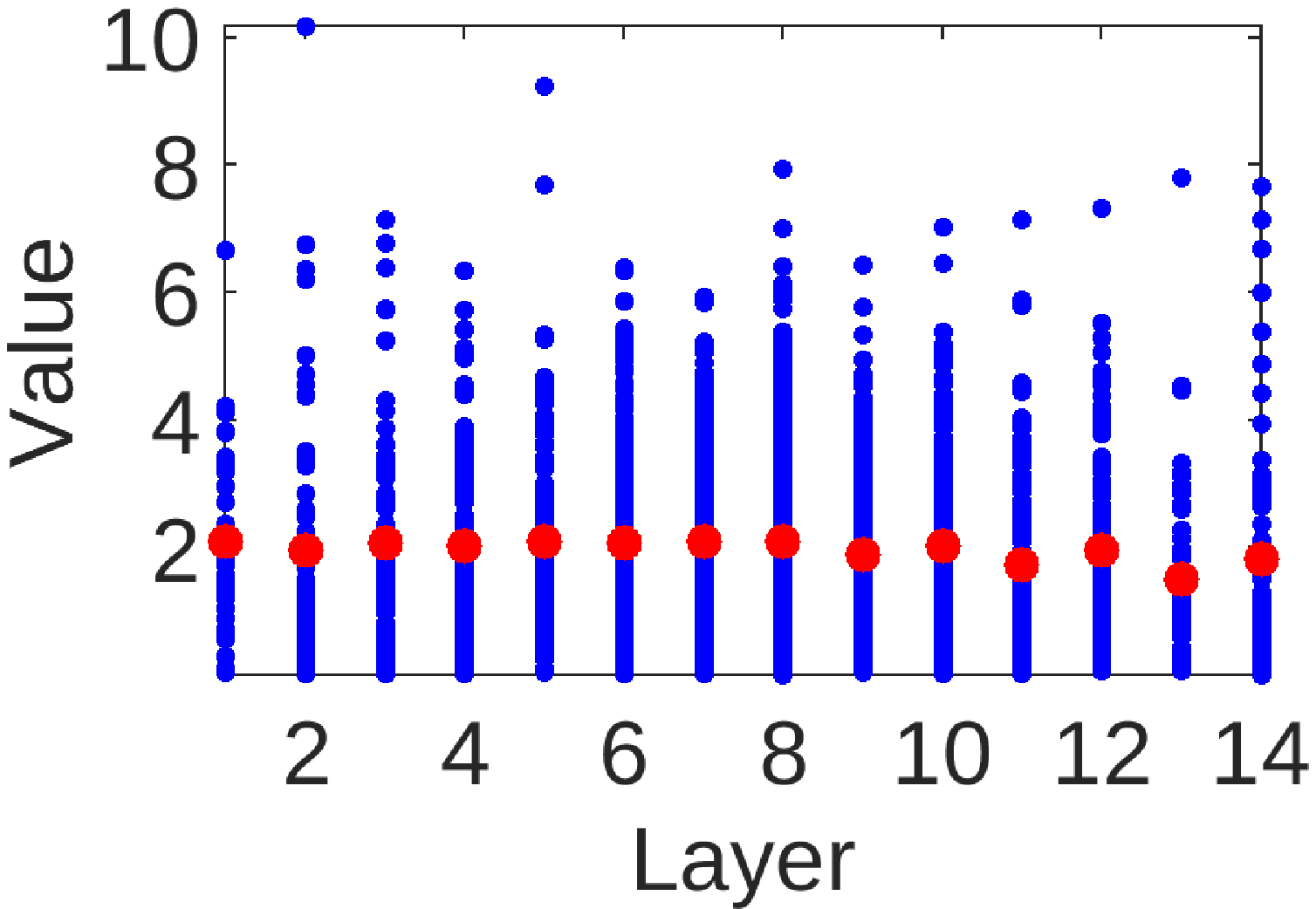}
  \caption{RANP-f Eq.~\myblue{\ref{eq:neuron_importance_reweighted}}}
  \label{fig:shapenet_grad_FLOPs}
  \end{subfigure}
  \hspace{3mm}
  \begin{subfigure}[b]{0.2\textwidth}
  \centering
  \includegraphics[width=\textwidth]{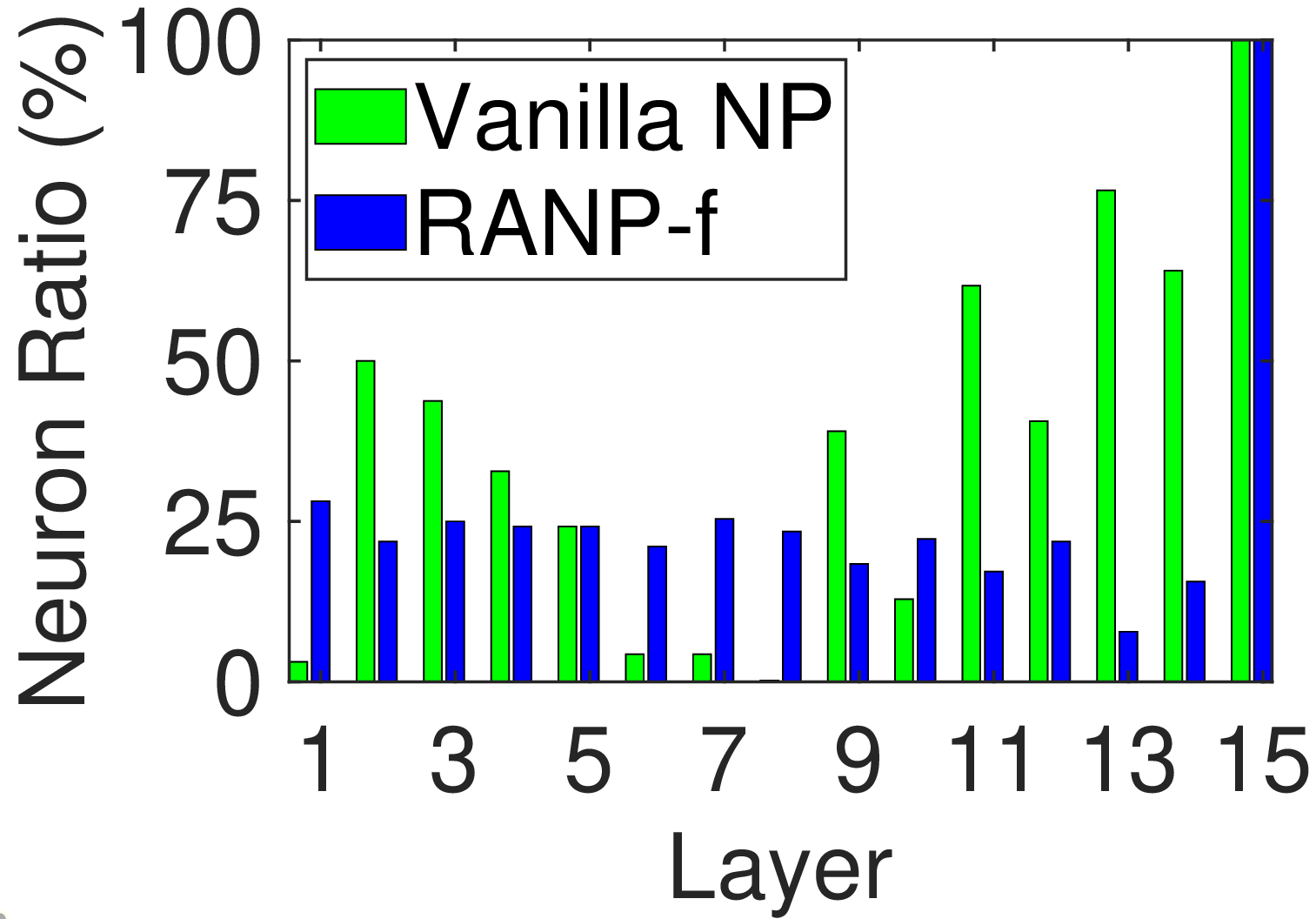}
  \caption{Retain ratio}
  \label{fig:shapenet_neuron_ratio_per_layer}
  \end{subfigure}
\vspace{-0.6em}
\caption{ShapeNet: neuron importance of 3D-UNet becomes balanced and resource-aware from (a) to (c) at neuron sparsity 78.24\%.
Blue: neuron importance; red: mean values.
More illustrations are in Appendix~\myref{D}.}
\label{fig:neuron_distribution_shapenet}
\end{center}
\vspace{-9mm}
\end{figure}

To tackle the imbalanced neuron importance issue above, we first weight the neuron importance across layers.
Weighting neuron importance of $x^l_u$ can be expressed as
\vspace{-0.5em}
\begin{equation}
\label{eq:neuron_importance_weighted}
\begin{aligned}
\tilde{s}^l_u = \frac{\max_{k=1}^K \bar{s}^k}{\bar{s}^l} s^l_u\ , \enskip \mbox{where $\bar{s}^k = \frac{1}{N_k} \sum_{u=1}^{N_k} s^k_u, \enskip \forall\,k\in\mathcal{K}$}\ .
\end{aligned}
\end{equation}
Here, $\bar{s}^l$ is the mean neuron importance of layer $l$ and $\tilde{s}^l_u$ is the updated neuron importance. 
This helps to achieve the same mean neuron importance in each layer, which largely avoids underestimating neuron importance of specific layer(s) to prevent from pruning the whole layer(s).

To further reduce the memory and FLOPs with minimal accuracy loss, we then reweight the neuron importance $\tilde{s}^l_u$ by available resource, \ie, memory or FLOPs.
This reweighting counts on the addition of weighted neuron importance and the effect of the computational resource, denoted as RANP-[m$\vert$f], where ``m" is for memory and ``f" is for FLOPs .
We represent the importance of this available resource in layer $l$ as $\tau_l$, refer to Appendix~\myref{C} for details.

The reweighted neuron importance of neuron $x^l_u$ by following weighted addition variant RANP-[m$\vert$f] is
\vspace{-0.7em}
\begin{equation}
\label{eq:neuron_importance_reweighted}
\begin{aligned}
\hat{s}^l_u
= \left(1 + \lambda\, \text{softmax}(-\tau_l) \right) \tilde{s}^l_u
= \left( 1 + \lambda \frac{e^{-\tau_l}}{\sum_{k=1}^K e^{-\tau_k}} \right) \tilde{s}^l_u\ ,
\end{aligned}
\end{equation}
where coefficient $\lambda>0$ helps to control the effect of resource on neuron importance.
This effect represented by softmax constrains the values into a controllable range [0,1], making it easy to determine $\lambda$ and function a high resource influence with a small resource occupation.

We demonstrate the effect of this reweighting strategy over vanilla pruning in Fig.~\myblue{\ref{fig:neuron_distribution_shapenet}}.
In more detail, vanilla neuron importance tends to have high values in the last few layers, making it highly possible to remove all neurons of such as the 7th and 8th layers.
Weighting the importance in Fig.~\myblue{\ref{fig:shapenet_grads}} makes the distribution of importance balanced with the same mean value in each layer.
Furthermore, since some neurons have different numbers of input channels, each layer requires different FLOPs and memory.
Considering the effect of computational resources on training, we embed them into neuron importance as weights.

In Fig.~\myblue{\ref{fig:shapenet_grad_FLOPs}}, the last few layers require larger computational resources than the others, and thus their neurons share lower weights, see the tendency of mean values.
Vividly, neuron ratio in Fig.~\myblue{\ref{fig:shapenet_neuron_ratio_per_layer}} indicates a more balanced distribution by RANP-f than vanilla NP.
For instance, very few neurons are retained in the 8th layer by vanilla NP, resulting in low accuracy and low maximum neuron sparsity.
With reweighting by RANP-f, however, more neurons can be retained in this layer.
Moreover, in Table~\myblue{\ref{tb:full_table}}, while weighted NP achieves high accuracy, its computational resource reductions are small.
In contrast, RANP-f largely decreases the computational resources with a small accuracy loss.

Then, with reweighted neuron importance by Eq.~\myblue{\ref{eq:neuron_importance_reweighted}} and $\ddot{s}_{\kappa}$ as the $\kappa$th reweighted neuron importance in a descending order, the binary mask of neuron $x^l_u$ can be obtained by

\vspace{-3mm}
\begin{equation}
\label{eq:binary_mask}
c^l_{u}=1[\hat{s}^l_u - \ddot{s}_{\kappa} \geq 0]\ .
\end{equation}

As mentioned in Sec.~\myblue{\ref{sec:related_work}},
our RANP is more effective in reducing memory and FLOPs than SNIP-based pruning which merely sparsifies parameters but needs high memory required by dense operations in training.
RANP can easily remove neurons and all involved input channels at once, leading to huge reductions of input and output channels of the filter.
Pseudocode is provided in Appendix~\myref{A}.

\section{Experiments}
We evaluated RANP on 3D-UNets for 3D semantic segmentation and MobileNetV2 and I3D for video classification.
Experiments are on Nvidia Tesla P100-SXM2-16GB GPUs in PyTorch.
More results are in Appendix~\myref{D}.
Our code is available at \textit{\url{https://github.com/zwxu064/RANP.git}}.

\subsection{Experimental Setup} \label{sec:setup}

\textbf{3D Datasets.} For 3D semantic segmentation, we adopted the large-scale 3D sparse point-cloud dataset, ShapeNet \cite{shapenet}, and dense biomedical MRI sequences, BraTS'18 \cite{brats1,brats2}.
\textit{ShapeNet} consists of
50 object part classes, 14007 training samples, and 2874 testing samples.
We split it into 6955 training samples and 7052 validation samples as \cite{ssc} to assign each point/voxel with a part class.

\textit{BraTS'18} includes 210 High Grade Glioma (HGG) and 75 Low Grade Glioma (LGG) cases.
Each case has 4 MRI sequences, \ie,
T1, T1\_CE, T2, and FLAIR.
The task is to detect and segment brain scan images into 3 categories: Enhancing Tumor (ET), Tumor Core (TC), and Whole Tumor (WT).
The spatial size is 240$\times$240$\times$155 in each dimension.
We adopted the splitting strategy of cross-validation in \cite{cross_validation} with 228 cases for training and 57 cases for validation.

For video classification, we used video dataset, \textit{UCF101} \cite{ucf101} with 101 action categories and 13320 videos.
2D spatial dimension from images and temporal dimension from frames are cast as dense 3D inputs.
Among the 3 official train/test splits, we used split-1 which has 9537 videos for training and 3783 videos for validation.

\textbf{3D CNNs.} For 3D semantic segmentation on ShapeNet (sparse data) and BraTS'18 (dense data), we used the standard 15-layer 3D-UNet \cite{3dunet}
including 4 encoders, each consists of two ``3D convolution + 3D batch normalization + ReLU", a ``3D max pooling", four decoders, and a confidence module by softmax.
It has 14 convolution layers with 3$^3$ kernels and 1 layer with 1$^3$ kernel.

For video classification, we used the popular MobileNetV2 \cite{mobilenetv2,3d_resource_efficient} and I3D (with inception as backbone) \cite{i3d} on UCF101.
MobileNetV2 has a linear layer and 52 convolution layers while 18 of them are 3$^3$ kernels and the rest are 1$^3$.
I3D has a linear layer and 57 convolution layers, 19 of which are 3$^3$ kernels, 1 is 7$^3$, and the rest are 1$^3$.

\textbf{Hyper-parameters in learning.} For \textit{ShapeNet}, we set learning rate as 0.1 with an exponential decay rate $\gamma=0.04$ by 100 epochs; batch size is 12 on 2 GPUs; spatial size for pruning and training is $64^3$ while the spatial size for training is $128^3$ in Sec.~\myblue{\ref{sec:fast_convergence}}; optimizer is SGD-Nesterov \cite{sgd} with weight decay 0.0001 and momentum 0.9.

For \textit{BraTS'18}, learning rate is 0.001, decayed by 0.1 at 150th epoch with 200 epochs; optimizer is Adam\cite{adam} with weight decay 0.0001 and AMSGrad\cite{amsgrad}; batch size is 2 on 2 GPUs; spatial size for pruning is $96^3$ and $128^3$ for training.

For \textit{UCF101}, we adopted similar setup from \cite{3d_resource_efficient} with learning rate 0.1, decayed by 0.1 at \{40, 55, 60, 70\}th epoch;
optimizer by SGD with weight decay 0.001;
batch size 8 on one GPU.
Spatial size for pruning and training is $112^2$ for MobileNetV2 and $224^2$ for I3D; 16 frames are used for the temporal size.
Note that in \cite{3d_resource_efficient} networks for UCF101 had higher performance since they were pretrained on Kinetics600, while we directly trained on UCF101.
A feasible train-from-scratch reference could be \cite{ucf101}.

For Eq.~\myblue{\ref{eq:neuron_importance_reweighted}}, we empirically set the coefficient $\lambda$ as 11 for ShapeNet, 15 for BraTS'18, and 80 for UCF101.
Glorot initialization \cite{xavier} was used for weight initialization.
Note that we used orthogonal initialization \cite{orthogonal} to handle imbalanced layer-wise neuron importance distribution \cite{signal_propagation} but obtained lower maximum neuron sparsity.

In addition, loss function and metrics are in Appendix~\myref{A}.

\subsection{Maximum Neuron Sparsity by Vanilla NP}

\begin{table}[t]
\centering
\caption{Vanilla NP by max neuron sparsity.
``Metric" is mIoU for ShapeNet, ET for BraTS'18, top-1 for UCF101.
``Param" and ``Mem" are in MB.
MPMG-sum is vanilla NP for large resource reductions and small metric loss.
}
\vspace{-2mm}
\label{tb:neuron_importance}
\setlength{\tabcolsep}{2pt}
\resizebox{0.48\textwidth}{!}{\begin{tabular}{llrrrrc}
  \hline
   Dataset (Model) & Manner & Sparsity
   & Param & GFLOPs & Mem & Metric \\
  \hline
  \multirow{3}{*}{\makecell[l]{ShapeNet\\(3D-UNet)}}
  & Full\cite{3dunet} & 0 & 62.26 & 237.85 & 997.00
  & \accerror{83.79}{0.21} \\
  &\makecell{MNMG-sum} & 66.93 & 4.29 & 100.34 & 783.14
  & \textbf{\accerror{83.65}{0.02}} \\
  &\makecell{MPMG-sum} & 78.24 & 2.54
  & \textbf{55.69} & \textbf{557.32}
  & \accerror{83.26}{0.14} \\
  \hline
  \multirow{3}{*}{\makecell[l]{BraTS'18\\(3D-UNet)}}
  & Full\cite{3dunet} & 0 & 15.57 & 478.13 & 3628.00
  & \accerror{72.96}{0.60} \\
  & MNMG-sum & 81.32 & 0.35 & \textbf{73.50}
  &\textbf{1933.20} & \accerror{64.48}{1.10} \\
  &MPMG-sum & 78.17 & 0.55 & 104.50 & 1936.44
  & \textbf{\accerror{71.94}{1.68}} \\
  \hline
  \multirow{3}{*}{\makecell[l]{UCF101\\(MobileNetV2)}}
  & Full\cite{mobilenetv2} & 0 & 9.47 & 0.58 & 157.47
  & \accerror{47.08}{0.72} \\
  &\makecell{MNMG-sum} & 39.89 & 4.66
  & \textbf{0.43} & \textbf{120.01}
  & \accerror{1.03}{0.00}\footnotemark \\
  &\makecell{MPMG-sum} & 33.15 & 6.35 & 0.55 & 155.17
  & \textbf{\accerror{46.32}{0.79}} \\
  \hline
  \multirow{3}{*}{\makecell[l]{UCF101\\(I3D)}}
  & Full\cite{i3d} & 0 & 47.27 & 27.88 & 201.28
  & \accerror{51.58}{1.86} \\
  & MNMG-sum & 32.87 & 20.00 & \textbf{16.03}
  &\textbf{125.17} & \accerror{49.02}{3.33} \\
  & MPMG-sum & 25.32 & 29.93 & 25.76 & 192.42
  & \textbf{\accerror{51.57}{1.46}} \\
  \hline
\end{tabular}}
\vspace{-6mm}
\end{table}
\footnotetext[3]{Since 2 layers of the pruned MobileNetV2 by MNMG-sum have only 1 neuron due to the imbalanced layer-wise neuron importance distribution.}

We selected MPMG-sum and MNMG-sum for vanilla neuron importance for comparison.
All neurons of the last convolutional layer are retained for the given classes.

In Table~\myblue{\ref{tb:neuron_importance}}, MPMG-sum for ShapeNet achieves the largest neuron sparsity 78.24\% by reducing 76.59\% FLOPs, 95.92\% parameters, and 44.10\% memory with 0.53\% accuracy loss.
Meanwhile, for BraTS'18, MNMG-sum achieves the largest neuron sparsity 81.32\% but has up to 8.48\% accuracy loss.
MPMG-sum, however, has the largest neuron sparsity 78.17\% but smaller accuracy loss with decreased 78.14\% FLOPs, 96.46\% parameters, and 46.63\% memory.

Hence, we selected MPMG-sum as vanilla NP considering the trade-off between the maximum neuron sparsity and the accuracy loss.
This is applied to all methods related to weighted neuron pruning and RANP in our experiments.
Results of mean and max are in Appendix~\myref{D}.

\subsection{Evaluation of RANP on Pruning Capability} \label{subsec:pruning_capability}
\vspace{-1mm}

\begin{table*}[!t]
\vspace{-3mm}
\centering
\caption{Evaluation of neuron pruning capability.
All models are \textbf{trained from scratch} for 100 epochs on ShapeNet and UCF101, 200 on BraTS'18.
Metrics are calculated by the last 5 epochs.
``sparsity" is max parameter sparsity for SNIP NP and max neuron sparsity for others.
Among the neuron pruning methods, we marked bold \textbf{the best} and underlined \sunderline{the second best}.
``$\downarrow$" denotes reduction in \%.
Overall, our RANP-f performs best with large reductions of \textbf{main resource consumption} (GFLOPs and memory) with negligible accuracy loss.
}
\label{tb:full_table}
\vspace{-2mm}
\centering
\resizebox{\textwidth}{!}{
\begin{tabular}{l|lrlrrrrcccccc}
  \hline
  \multicolumn{1}{c}{Dataset} &
  \multicolumn{1}{c}{Model} &
  & \multicolumn{1}{c}{Manner}
  & \multicolumn{1}{c}{Sparsity(\%)}
  & \multicolumn{1}{c}{Param(MB)}
  & \multicolumn{1}{c}{GFLOPs}
  & \multicolumn{1}{c}{Memory(MB)}
  & \multicolumn{6}{c}{Metrics(\%)} \\
  \hline
  &&&&&&&& \multicolumn{6}{c}{mIoU} \\
  \multirow{8}{*}{\makecell{ShapeNet\\\cite{shapenet}}} & \multirow{8}{*}{\makecell{3D-UNet}}
  && Full\cite{3dunet}
  & 0 & 62.26\hspace{26pt}
  & 237.85\hspace{26pt} & 997.00\hspace{26pt}
  & \multicolumn{6}{c}{\accerror{83.79}{0.21}} \\
  & &
  & SNIP\cite{snip} NP
  & 98.98 & 5.31 (91.5$\downarrow$)
  & 126.22 (46.9$\downarrow$)
  & 833.20 (16.4$\downarrow$)
  & \multicolumn{6}{c}{\textbf{\accerror{83.70}{0.20}}} \\
  & && Random NP & \multirow{6}{*}{78.24} & 3.05 (95.1$\downarrow$)
  & 10.36 (95.6$\downarrow$) & 267.95 (73.1$\downarrow$)
  & \multicolumn{6}{c}{\accerror{82.90}{0.19}} \\
  & && Layer-wise NP && 2.99 (95.2$\downarrow$)
  & 11.63 (95.1$\downarrow$) & 296.22 (70.3$\downarrow$)
  & \multicolumn{6}{c}{\accerror{83.25}{0.14}} \\
  & & \multicolumn{1}{c|}{\multirow{4}{*}{\rotatebox{90}{ours}}}
  & Vanilla NP &
  & 2.54 (95.9$\downarrow$)
  & 55.69 (76.6$\downarrow$) & 557.32 (44.1$\downarrow$)
  & \multicolumn{6}{c}{\sunderline{\accerror{83.26}{0.14}}} \\
  & &\multicolumn{1}{c|}{} & Weighted NP
  & & 2.97 (95.2$\downarrow$)
  & 12.06 (94.9$\downarrow$)
  & 301.56 (69.8$\downarrow$)
  & \multicolumn{6}{c}{\accerror{83.12}{0.09}} \\
  & & \multicolumn{1}{c|}{} & RANP-m
  & & 3.39 (94.6$\downarrow$)
  & \textbf{6.68 (97.2$\downarrow$)}
  & \textbf{214.95 (78.4$\downarrow$)} & \multicolumn{6}{c}{\accerror{82.35}{0.24}} \\
  & & \multicolumn{1}{c|}{} & RANP-f &
  & 2.94 (95.3$\downarrow$)
  & \sunderline{7.54 (96.8$\downarrow$)}
  & \sunderline{262.66 (73.7$\downarrow$)} & \multicolumn{6}{c}{\accerror{83.07}{0.22}} \\
  \hline
  &&&&&&&& \multicolumn{2}{c}{ET} & \multicolumn{2}{c}{TC} & \multicolumn{2}{c}{WT} \\
  \multirow{8}{*}{\makecell{BraTS'18\\\cite{brats1,brats2}}} & \multirow{8}{*}{\makecell{3D-UNet}}
  && Full\cite{3dunet} & 0
  & 15.57\hspace{26pt}
  & 478.13\hspace{26pt}
  & 3628.00\hspace{26pt} & \multicolumn{2}{c}{\accerror{72.96}{0.60}}
  & \multicolumn{2}{c}{\accerror{73.51}{1.54}}
  & \multicolumn{2}{c}{\accerror{86.79}{0.35}} \\
  && & SNIP\cite{snip} NP & 98.88
  & 1.09 (93.0$\downarrow$)
  & 233.11 (51.2$\downarrow$)
  & 2999.64 (17.3$\downarrow$)
  & \multicolumn{2}{c}{\textbf{\accerror{73.33}{1.89}}}
  & \multicolumn{2}{c}{\accerror{71.98}{2.15}}
  & \multicolumn{2}{c}{\textbf{\accerror{86.44}{0.39}}} \\
  && & Random NP & \multirow{6}{*}{78.17}
  & 0.75 (95.2$\downarrow$)
  & 22.59 (95.3$\downarrow$)
  & 817.59 (77.5$\downarrow$)
  & \multicolumn{2}{c}{\accerror{67.27}{0.99}}
  & \multicolumn{2}{c}{\accerror{71.62}{1.20}}
  & \multicolumn{2}{c}{\accerror{74.16}{1.33}} \\
  && & Layer-wise NP &
  & 0.75 (95.2$\downarrow$)
  & 24.09 (95.0$\downarrow$)
  & 836.88 (77.0$\downarrow$)
  & \multicolumn{2}{c}{\accerror{69.74}{1.33}}
  & \multicolumn{2}{c}{\accerror{71.49}{1.62}}
  & \multicolumn{2}{c}{\sunderline{\accerror{86.38}{0.39}}} \\
  & & 
  \multicolumn{1}{c|}{\multirow{4}{*}{\rotatebox{90}{ours}}}
  & Vanilla NP &
  & 0.55 (96.5$\downarrow$)
  & 104.50 (78.1$\downarrow$)
  & 1936.44 (46.6$\downarrow$)
  & \multicolumn{2}{c}{\sunderline{\accerror{71.94}{1.68}}}
  & \multicolumn{2}{c}{\accerror{69.39}{2.29}}
  & \multicolumn{2}{c}{\accerror{84.68}{0.78}} \\
  && \multicolumn{1}{c|}{} & Weighted NP &
  & 0.79 (95.0$\downarrow$)
  & 22.40 (95.3$\downarrow$)
  & 860.64 (76.3$\downarrow$)
  & \multicolumn{2}{c}{\accerror{71.50}{0.63}}
  & \multicolumn{2}{c}{\textbf{\accerror{75.05}{1.19}}}
  & \multicolumn{2}{c}{\accerror{84.05}{0.65}} \\
  && \multicolumn{1}{c|}{} & RANP-m &
  & 0.87 (94.4$\downarrow$)
  & \textbf{13.47 (97.2$\downarrow$)}
  & \textbf{506.97 (86.0$\downarrow$)}
  & \multicolumn{2}{c}{\accerror{66.70}{2.94}}
  & \multicolumn{2}{c}{\accerror{62.99}{2.38}}
  & \multicolumn{2}{c}{\accerror{82.90}{0.41}} \\
  && \multicolumn{1}{c|}{} & RANP-f &
  & 0.76 (95.1$\downarrow$)
  & \sunderline{16.97 (96.5$\downarrow$)}
  & \sunderline{729.11 (80.0$\downarrow$)}
  & \multicolumn{2}{c}{\accerror{70.73}{0.66}}
  & \multicolumn{2}{c}{\sunderline{\accerror{74.50}{1.05}}}
  & \multicolumn{2}{c}{\accerror{85.45}{1.06}} \\
  \hline
  &&&&&&&& \multicolumn{3}{c}{\hspace{10mm} Top-1} & \multicolumn{3}{c}{\hspace{-12mm} Top-5} \\
  \multirow{16}{*}{\makecell{UCF101\\\cite{ucf101}}}
  & \multirow{8}{*}{MobileNetV2}
  && Full\cite{mobilenetv2} & 0
  & 9.47\hspace{26.5pt}
  & 0.58\hspace{26pt}
  & 157.47\hspace{26pt}
  & \multicolumn{3}{c}{\hspace{10mm}
  \accerror{47.08}{0.72}}
  & \multicolumn{3}{c}{\hspace{-12mm}
  \accerror{76.68}{0.50}} \\
  && & SNIP\cite{snip} NP
  & 86.26
  & 3.67 (61.3$\downarrow$)
  & 0.54 (\hspace{4pt}6.9$\downarrow$)
  & 155.35 (\hspace{1pt} 1.3$\downarrow$)
  & \multicolumn{3}{c}{\hspace{10mm}
  \accerror{45.78}{0.04}}
  & \multicolumn{3}{c}{\hspace{-12mm}
  \accerror{75.08}{0.17}} \\
  && & Random NP & \multirow{6}{*}{33.15}
  & 4.58 (51.6$\downarrow$)
  & 0.34 (41.4$\downarrow$)
  & 106.68 (32.3$\downarrow$)
  & \multicolumn{3}{c}{\hspace{10mm}
  \accerror{44.74}{0.36}}
  & \multicolumn{3}{c}{\hspace{-12mm}
  \accerror{74.69}{0.58}} \\
  && & Layer-wise NP &
  & 4.56 (51.8$\downarrow$)
  & 0.33 (43.1$\downarrow$)
  & 106.92 (32.1$\downarrow$)
  & \multicolumn{3}{c}{\hspace{10mm}
  \accerror{44.90}{0.36}}
  & \multicolumn{3}{c}{\hspace{-12mm}
  \accerror{75.54}{0.34}} \\
  & & \multicolumn{1}{c|}{\multirow{4}{*}{\rotatebox{90}{ours}}}
  & Vanilla NP &
  & 6.35 (32.9$\downarrow$)
  & 0.55 (\hspace{4pt}5.2$\downarrow$)
  & 155.17 (\hspace{4pt}1.5$\downarrow$)
  & \multicolumn{3}{c}{\hspace{10mm}
  \textbf{\accerror{46.32}{0.79}}}
  & \multicolumn{3}{c}{\hspace{-12mm}
  \accerror{75.42}{0.60}} \\
  && \multicolumn{1}{c|}{}
  & Weighted NP &
  & 4.82 (49.1$\downarrow$)
  & 0.30 (48.3$\downarrow$)
  & 100.33 (36.3$\downarrow$)
  & \multicolumn{3}{c}{\hspace{10mm}
  \sunderline{\accerror{46.19}{0.51}}}
  & \multicolumn{3}{c}{\hspace{-12mm}
  \sunderline{\accerror{75.72}{0.30}}} \\
  && \multicolumn{1}{c|}{}
  & RANP-m &
  & 4.87 (48.6$\downarrow$)
  & \sunderline{0.27 (53.4$\downarrow$)}
  & \textbf{84.51 (46.3$\downarrow$)}
  & \multicolumn{3}{c}{\hspace{10mm}
  \accerror{45.11}{0.41}}
  & \multicolumn{3}{c}{\hspace{-12mm}
  \accerror{75.53}{0.37}} \\
  && \multicolumn{1}{c|}{}
  & RANP-f &
  & 4.83 (49.0$\downarrow$)
  & \textbf{0.26 (55.2$\downarrow$)}
  & \sunderline{88.01 (44.1$\downarrow$)}
  & \multicolumn{3}{c}{\hspace{10mm}
  \accerror{45.87}{0.41}}
  & \multicolumn{3}{c}{\hspace{-12mm}
  \textbf{\accerror{75.75}{0.30}}} \\
  \cline{2-14}
  & \multirow{8}{*}{\makecell{I3D}}
  && Full\cite{i3d} & 0
  & 47.27\hspace{26pt}
  & 27.88\hspace{26pt}
  & 201.28\hspace{26pt}
  & \multicolumn{3}{c}{\hspace{10mm}
  \accerror{51.58}{1.86}}
  & \multicolumn{3}{c}{\hspace{-12mm}
  \accerror{77.35}{0.63}} \\
  && & SNIP\cite{snip} NP & 81.09
  & 30.06 (36.4$\downarrow$)
  & 26.31 (\hspace{4pt}5.6$\downarrow$)
  & 195.62 (\hspace{4pt}2.8$\downarrow$)
  & \multicolumn{3}{c}{\hspace{10mm}
  \accerror{52.38}{3.55}}
  & \multicolumn{3}{c}{\hspace{-12mm}
  \accerror{78.32}{3.24}} \\
  && & Random NP
  & \multirow{6}{*}{25.32}
  & 26.36 (44.2$\downarrow$)
  & 16.45 (41.0$\downarrow$)
  & 145.07 (27.9$\downarrow$)
  & \multicolumn{3}{c}{\hspace{10mm}
  \accerror{52.42}{2.52}}
  & \multicolumn{3}{c}{\hspace{-12mm}
  \sunderline{\accerror{79.05}{2.06}}} \\
  && & Layer-wise NP &
  & 26.67 (43.6$\downarrow$)
  & 16.93 (39.3$\downarrow$)
  & 150.95 (25.0$\downarrow$)
  & \multicolumn{3}{c}{\hspace{10mm}
  \accerror{52.77}{1.99}}
  & \multicolumn{3}{c}{\hspace{-12mm}
  \accerror{78.41}{1.07}} \\
  & & \multicolumn{1}{c|}{\multirow{4}{*}{\rotatebox{90}{ours}}}
  & Vanilla NP &
  & 29.93 (36.7$\downarrow$)
  & 25.76 (\hspace{4pt}7.6$\downarrow$)
  & 192.42 (\hspace{4pt}4.4$\downarrow$)
  & \multicolumn{3}{c}{\hspace{10mm}
  \accerror{51.57}{1.46}}
  & \multicolumn{3}{c}{\hspace{-12mm}
  \accerror{78.07}{1.34}} \\
  && \multicolumn{1}{c|}{} & Weighted NP &
  & 26.57 (43.8$\downarrow$)
  & 15.56 (44.2$\downarrow$)
  & 142.57 (29.2$\downarrow$)
  & \multicolumn{3}{c}{\hspace{10mm}
  \sunderline{\accerror{54.09}{0.82}}} & \multicolumn{3}{c}{\hspace{-12mm}
  \accerror{79.26}{0.61}} \\
  && \multicolumn{1}{c|}{}
  & RANP-m &
  & 26.75 (43.4$\downarrow$)
  & \sunderline{14.08 (49.5$\downarrow$)}
  & \sunderline{130.44 (35.2$\downarrow$)}
  & \multicolumn{3}{c}{\hspace{10mm}
  \accerror{52.11}{3.05}}
  & \multicolumn{3}{c}{\hspace{-12mm}
  \accerror{77.54}{2.64}} \\
  && \multicolumn{1}{c|}{}
  & RANP-f &
  & 26.69 (43.5$\downarrow$)
  & \textbf{13.98 (49.9$\downarrow$)}
  & \textbf{130.22 (35.3$\downarrow$)}
  & \multicolumn{3}{c}{\hspace{10mm}
  \textbf{\accerror{54.27}{2.88}}}
  & \multicolumn{3}{c}{\hspace{-12mm}
  \textbf{\accerror{79.27}{2.13}}} \\
  \hline
\end{tabular}}
\vspace{-1mm}
\end{table*}

Random NP retains $\kappa$ neurons with neuron indices randomly shuffled.
Layer-wise NP retains neurons using the same retain rate as $\kappa$ in each layer.
For SNIP-based parameter pruning, the parameter masks are post-processed by removing redundant parameters and then making sparse filters dense, which is denoted as SNIP NP.
For a fair comparison with SNIP NP, we used the maximum parameter sparsity 98.98\% for ShapeNet
, 98.88\% for BraTS'18,
86.26\% for MobileNetV2,
and 81.09\% for I3D.

\begin{table*}[!ht]
\centering
\caption{
In addition to Table~\myblue{\ref{tb:full_table}},
with similar GFLOPs or memory on 3D-UNets, our RANP-f achieves the highest accuracy.
}
\vspace{-2mm}
\label{tb:similar_resource}
\resizebox{\textwidth}{!}{
\begin{tabular}{lrrrrc|rrrrccc}
  \hline
  \multicolumn{1}{c}{\multirow{2}{*}{Manner}} & \multicolumn{5}{c}{ShapeNet} & \multicolumn{7}{c}{BraTS'18} \\
  & \multicolumn{1}{c}{Sparsity}
  & \multicolumn{1}{c}{Param}
  & \multicolumn{1}{c}{GFLOPs}
  & \multicolumn{1}{c}{Mem}
  & \multicolumn{1}{c}{mIoU}
  & \multicolumn{1}{c}{Sparsity}
  & \multicolumn{1}{c}{Param}
  & \multicolumn{1}{c}{GFLOPs}
  & \multicolumn{1}{c}{Mem}
  & \multicolumn{1}{c}{ET}
  & \multicolumn{1}{c}{TC}
  & \multicolumn{1}{c}{WT} \\
  \hline
  \multicolumn{1}{l|}{Random NP} & 81.01
  & 2.27
  & $\sim$7.54
  & 253.12 & \accerror{82.66}{0.23}
  & 81.08
  & 0.56
  & $\sim$16.97
  & 685.77 & \accerror{61.09}{1.87} & \accerror{68.94}{2.44} & \accerror{78.89}{2.47} \\
  \multicolumn{1}{l|}{Layer-wise NP} & 82.82
  & 1.84
  & $\sim$7.54
  & 255.67 & \accerror{82.82}{0.26}
  & 83.50
  & 0.46
  & $\sim$16.97
  & 700.64 & \sunderline{\accerror{70.50}{0.63}} & \sunderline{\accerror{74.27}{0.95}} & \sunderline{\accerror{83.63}{0.92}}\\
  \multicolumn{1}{l|}{Random NP} & 78.83
  & 2.87
  & 9.57
  & $\sim$262.66 & \sunderline{\accerror{82.86}{0.45}}
  & 80.90
  & 0.57
  & 17.95
  & $\sim$729.11 & \accerror{68.45}{1.11} & \accerror{70.67}{1.21} & \accerror{75.02}{0.79}\\
  \multicolumn{1}{l|}{Layer-wise NP} & 82.81
  & 1.94
  & 8.14
  & $\sim$262.66 & \accerror{82.52}{0.13}
  & 82.45
  & 0.51
  & 17.31
  & $\sim$729.11 & \accerror{70.45}{1.03} & \accerror{69.27}{1.95} & \accerror{82.42}{0.68}\\
  \multicolumn{1}{l|}{RANP-f(ours)} & 78.24
  & 2.94
  & 7.54
  & 262.66 & \textbf{\accerror{83.07}{0.22}}
  & 78.17
  & 0.76
  & 16.97
  & 729.11 & \textbf{\accerror{70.73}{0.66}} & \textbf{\accerror{74.50}{1.05}} & \textbf{\accerror{85.45}{1.06}} \\
  \hline
\end{tabular}}
\vspace{-6mm}
\end{table*}

\textbf{ShapeNet.} Compared with random NP and layer-wise NP in Table~\myblue{\ref{tb:full_table}}, the maximum reduced resources by vanilla NP are much less due to the imbalanced layer-wise distribution of neuron importance.
Weighted neuron importance by Eq.~\myblue{\ref{eq:neuron_importance_weighted}}, however, further reduces 18.3\% FLOPs and 29.6\% memory with 0.14\% accuracy loss.

Reweighting by RANP-f and RANP-m further reduces FLOPs and memory on the basis of weighted NP.
Here, RANP-f can reduce 96.8\% FLOPs, 95.3\% parameters, and 73.7\% memory over the unpruned networks.
Furthermore, with a similar resource in Table~\myblue{\ref{tb:similar_resource}}, RANP achieves $\sim$0.5\% increase in accuracy.
Note that a too-large $\lambda$ can additionally reduce the resources but at the cost of accuracy.

\textbf{BraTS'18.} In Table~\myblue{\ref{tb:full_table}}, RANP-f achieves 96.5\% FLOPs, 95.1\% parameters, and 80\% memory reductions.
It further reduces 18.3\% FLOPs and 33.3\% memory over vanilla NP while increasing -1.21\% ET, 5.11\% TC, and 0.77\% WT.
With a similar resource in Table~\myblue{\ref{tb:similar_resource}}, RANP achieves higher accuracy than random NP and layer-wise NP.

Additionally, Chen \etl \cite{frequency_domain} achieved 2$\times$ speedup on BraTS'18 with 3D-UNet.
In comparison, our RANP-f has roughly 28$\times$ speedup, which is theoretically evidenced by the reduced FLOPs from 478.13G to 16.97G in Table~\ref{tb:full_table}.

\textbf{UCF101.} In Table~\myblue{\ref{tb:full_table}}, for MobileNetV2, RANP-f reduces 55.2\% FLOPs, 49\% parameters, and 44.1\% memory with around 1\% accuracy loss.
Meanwhile, for I3D, it reduces 49.9\% FLOPs, 43.5\% parameters, and 35.3\% memory with around 2\% accuracy increase.
The RANP-based methods can reduce much more resources than other methods.

\vspace{-2mm}
\subsection{Resources and Accuracy with Neuron Sparsity}
\vspace{-1mm}
Here, we further studied the tendencies of resources and accuracy with an increasing neuron sparsity level from 0 to the maximum one with network feasibility.

\begin{figure*}[t]
\begin{center}
\vspace{-3mm}
  \begin{subfigure}[b]{0.2\textwidth}
  \centering
  \includegraphics[width=\textwidth]{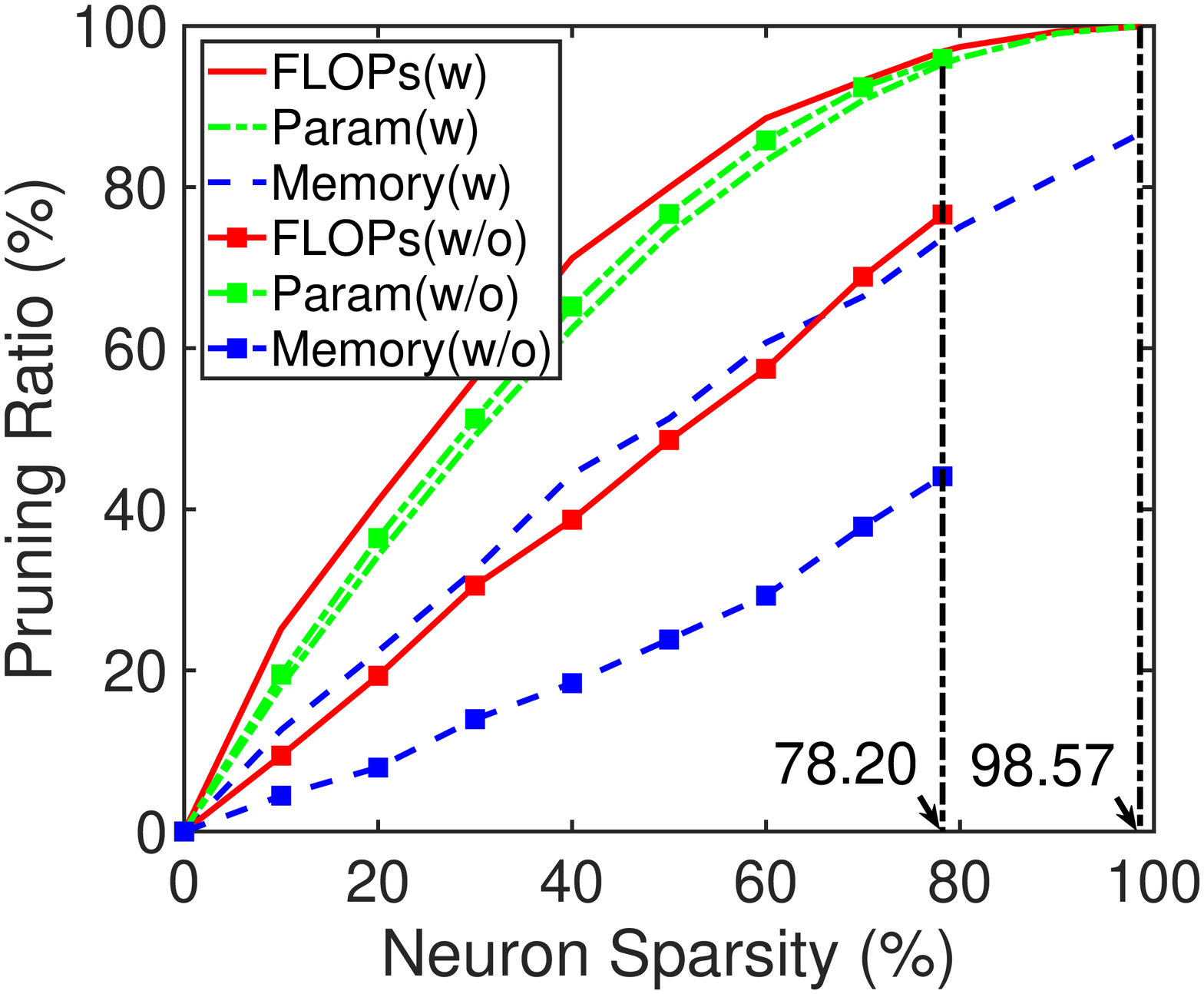}
  \caption{ShapeNet/3D-UNet}
  \label{fig:shapenet_resources}
  \end{subfigure}
  \hspace{3mm}
  \begin{subfigure}[b]{0.2\textwidth}
  \centering
  \includegraphics[width=\textwidth]{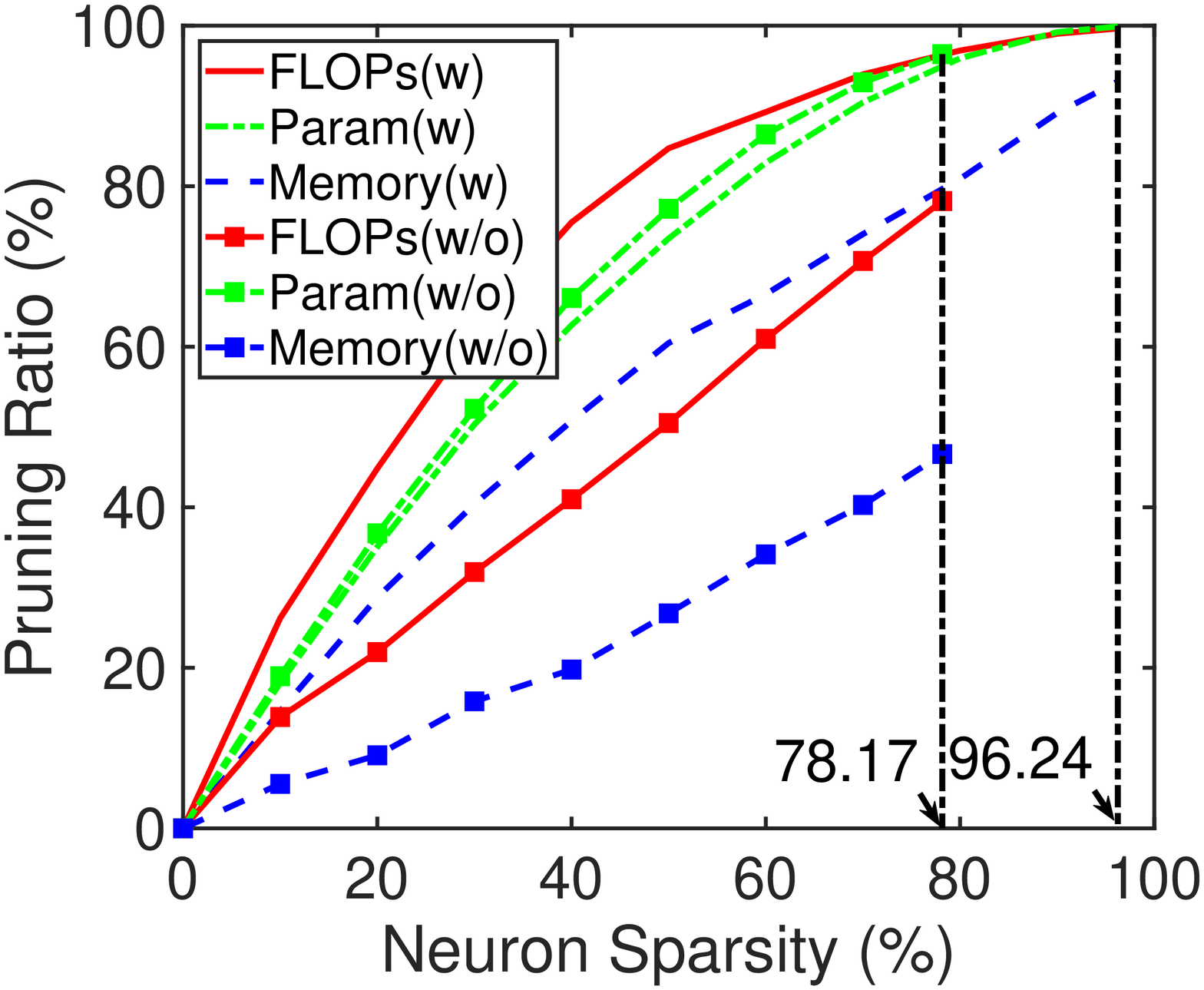}
  \caption{BraTS'18/3D-UNet}
  \label{fig:brats_resources}
  \end{subfigure}
  \hspace{3mm}
  \begin{subfigure}[b]{0.2\textwidth}
  \centering
  \includegraphics[width=\textwidth]{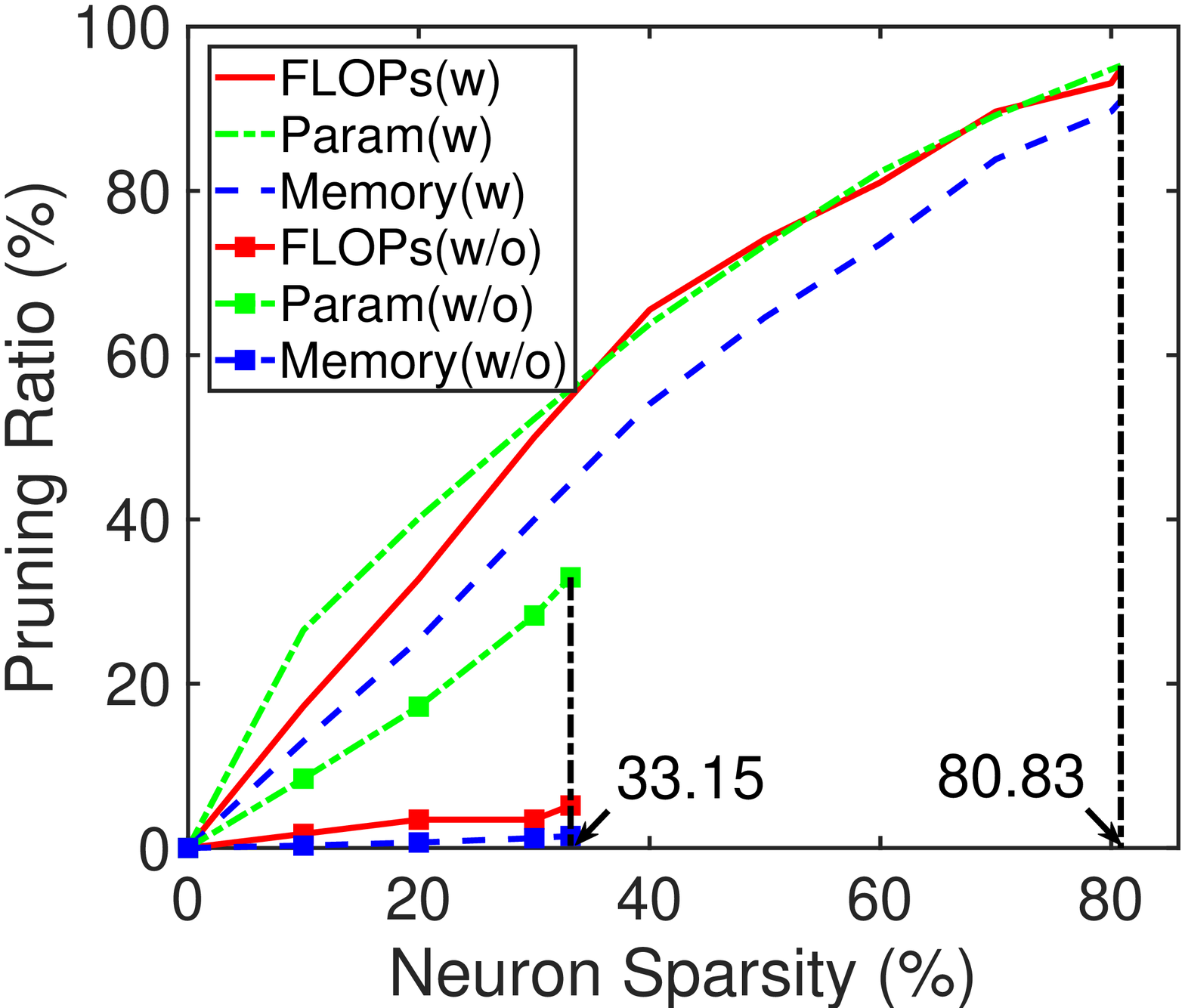}
  \caption{UCF101/MobileNetV2}
  \label{fig:mobilenetv2_resources}
  \end{subfigure}
  \hspace{3mm}
  \begin{subfigure}[b]{0.2\textwidth}
  \centering
  \includegraphics[width=\textwidth]{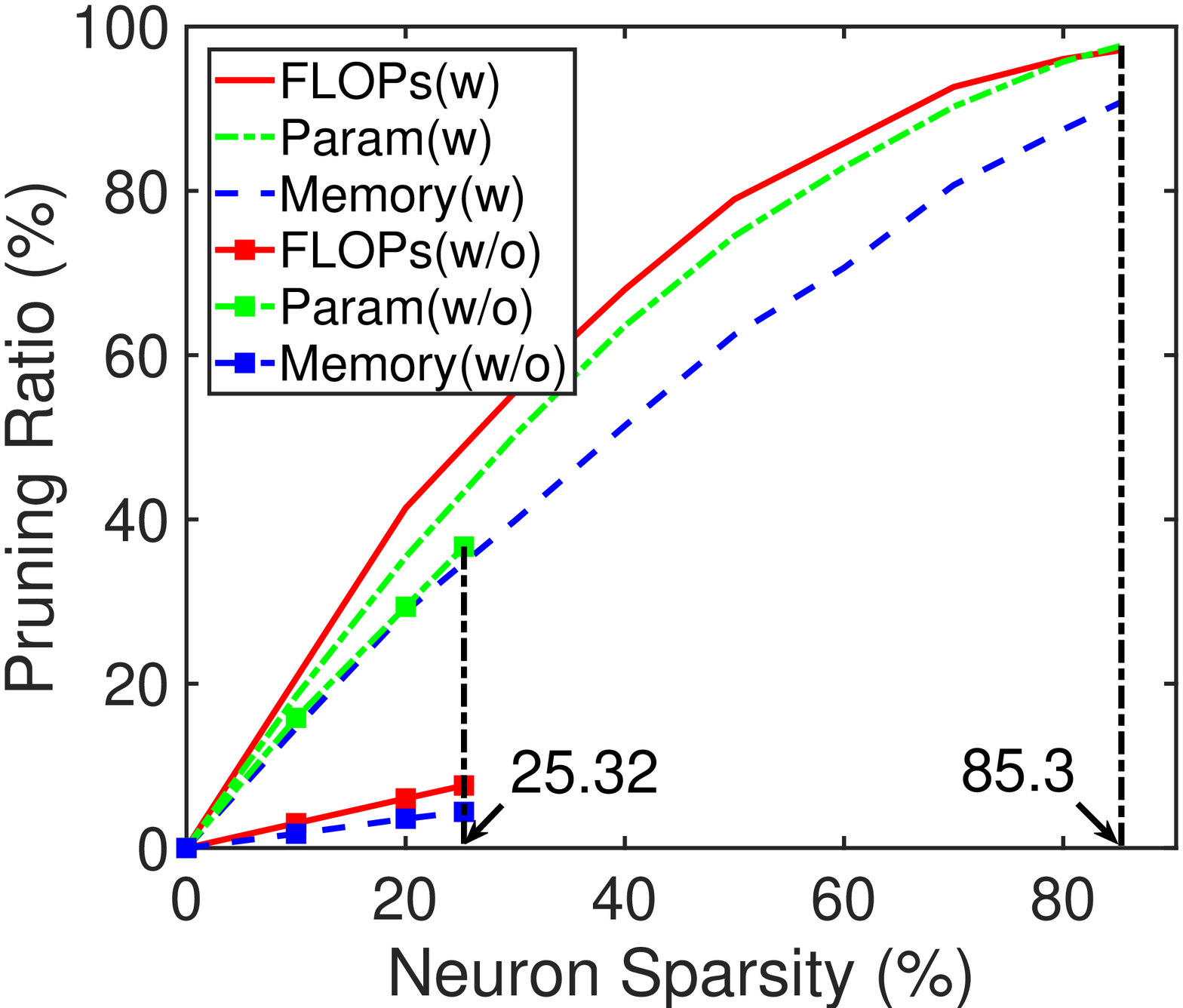}
  \caption{UCF101/I3D}
  \label{fig:i3d_resources}
  \end{subfigure}
  \\
  \begin{subfigure}[b]{0.2\textwidth}
  \centering
  \includegraphics[width=\textwidth]{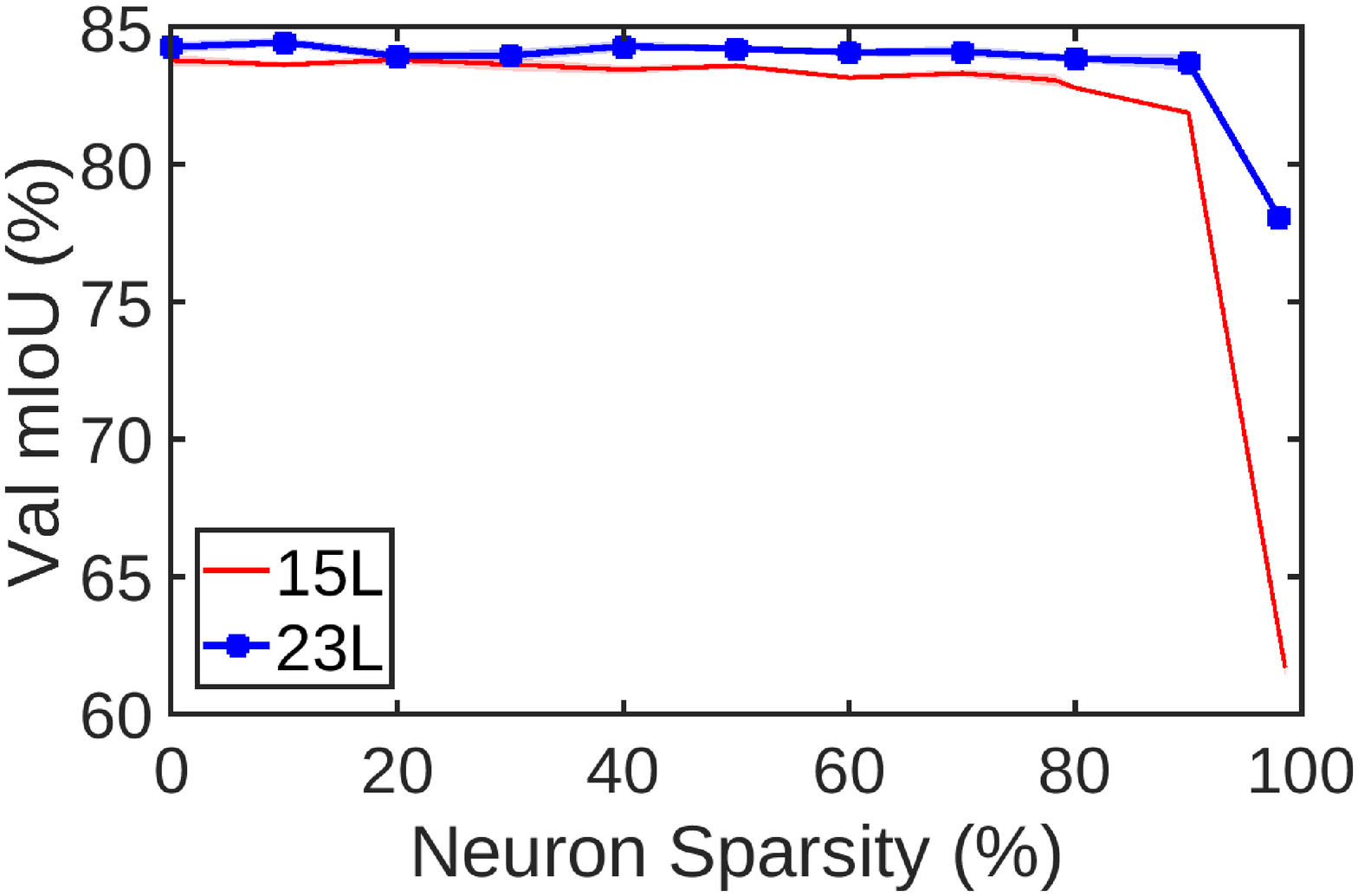}
  \caption{ShapeNet/3D-UNet}
  \label{fig:shapenet_acc}
  \end{subfigure}
  \hspace{3mm}
  \begin{subfigure}[b]{0.2\textwidth}
  \centering
  \includegraphics[width=\textwidth]{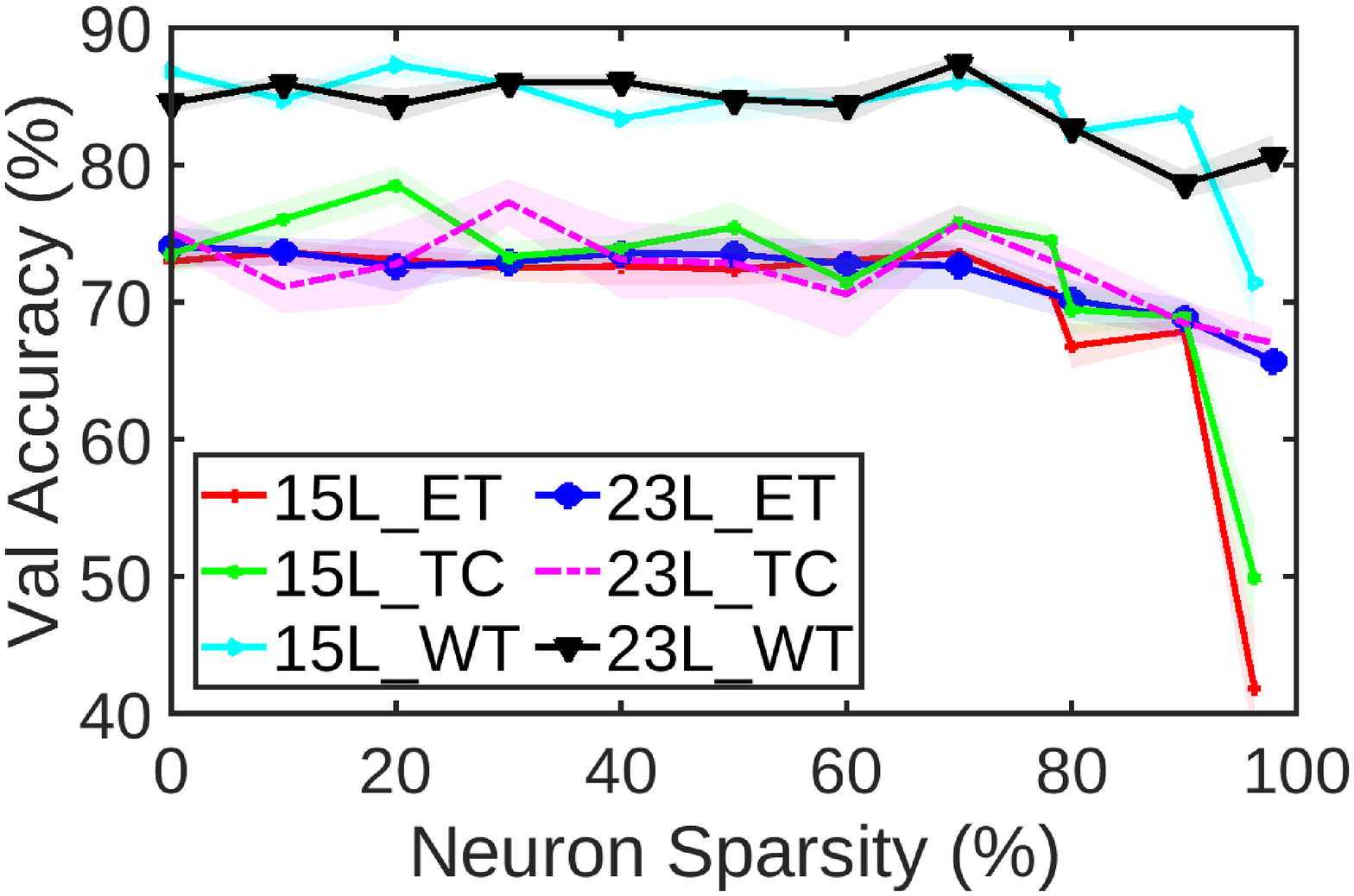}
  \caption{BraTS'18/3D-UNet}
  \label{fig:brats_acc}
  \end{subfigure}
  \hspace{3mm}
  \begin{subfigure}[b]{0.2\textwidth}
  \centering
  \includegraphics[width=\textwidth]{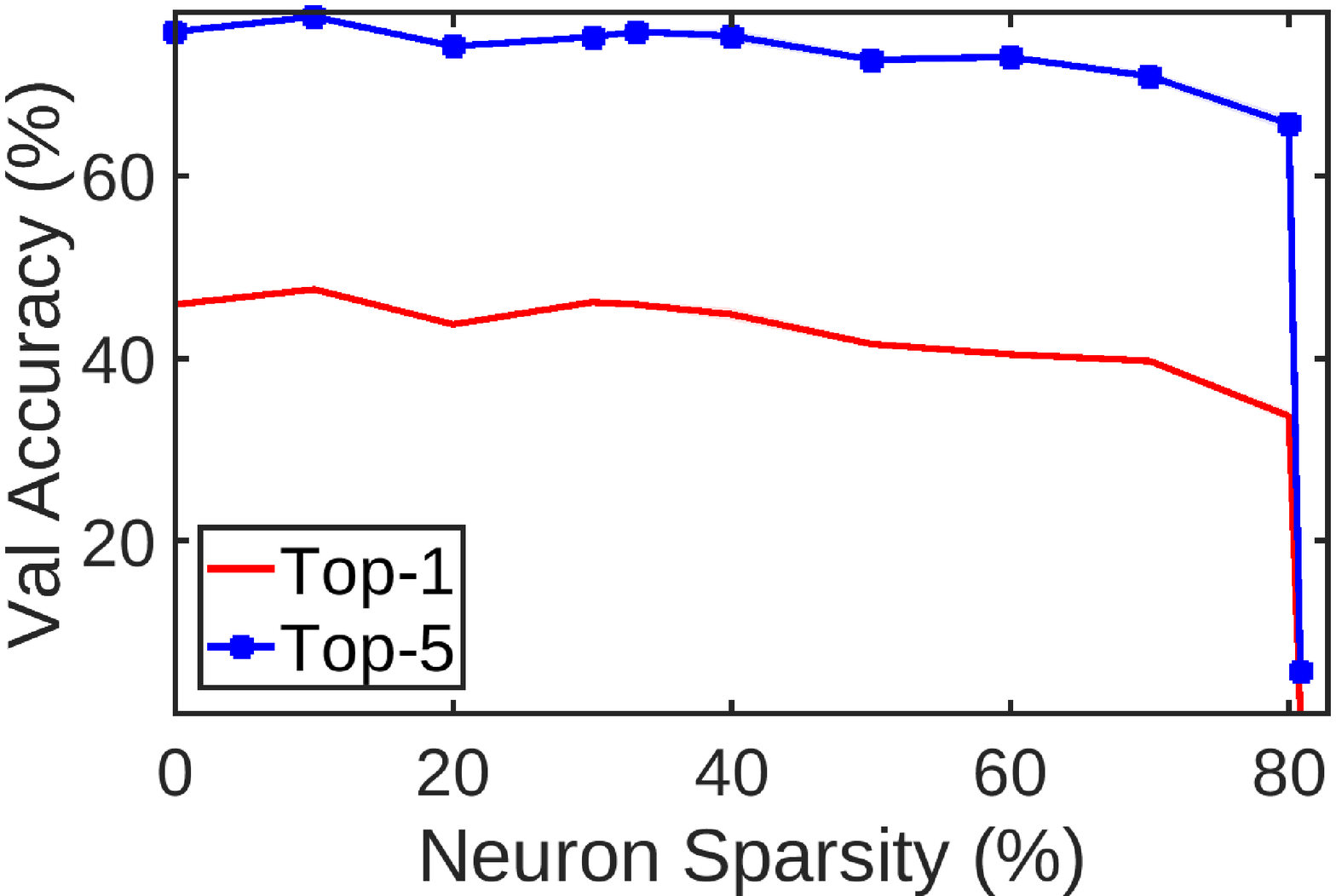}
  \caption{UCF101/MobileNetV2}
  \label{fig:mobilenetv2_acc}
  \end{subfigure}
  \hspace{3mm}
  \begin{subfigure}[b]{0.2\textwidth}
  \centering
  \includegraphics[width=\textwidth]{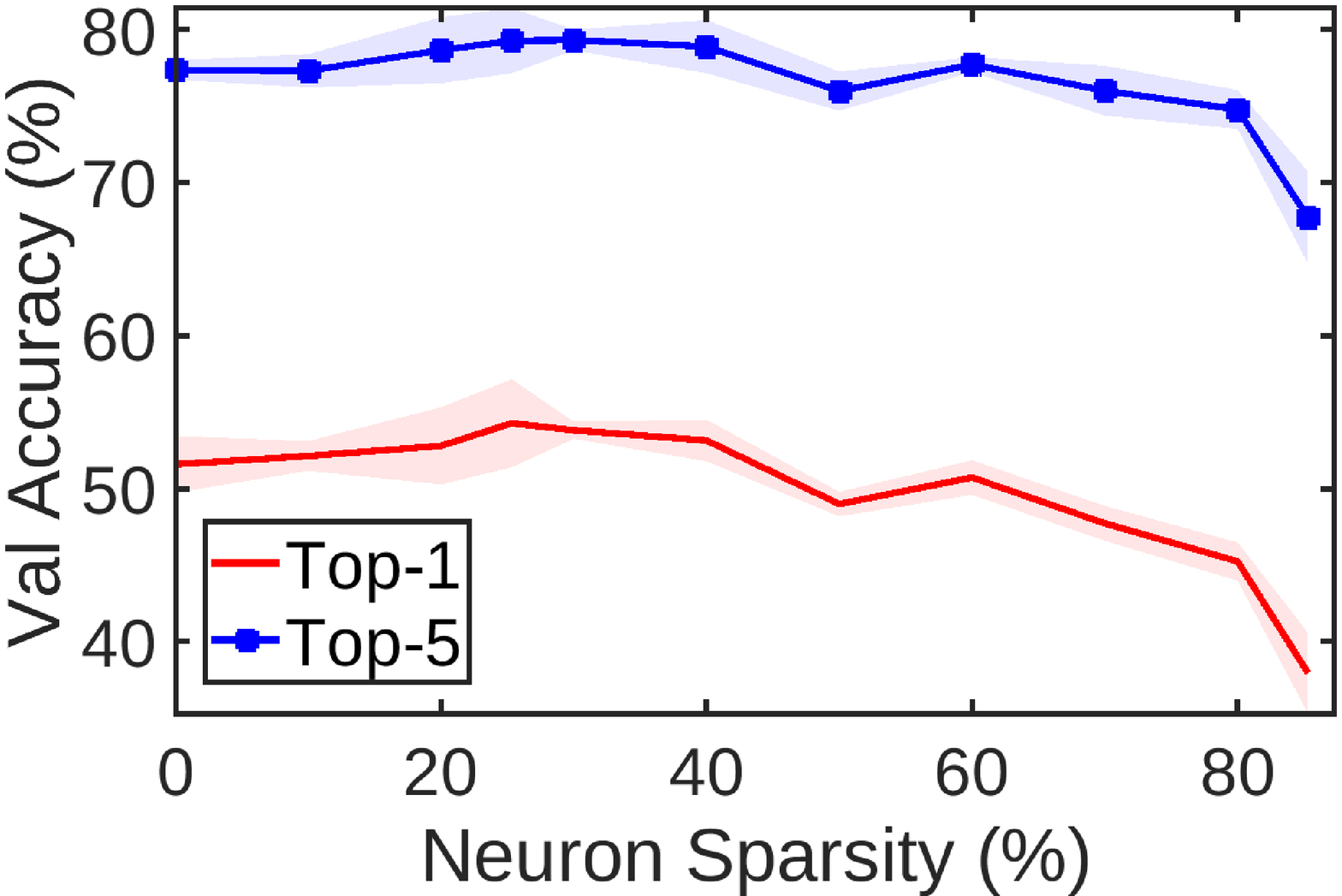}
  \caption{UCF101/I3D}
  \label{fig:i3d_acc}
  \end{subfigure}
\vspace{-2mm}
\caption{
With minimal accuracy loss, more resources are reduced with (w) reweighting by RANP-f than without (w/o) by vanilla NP.
(a)-(d) are resources reductions (w) and (w/o) reweighting;
(e)-(h) are accuracy and sparsity.
Best view in color.}
\end{center}
\vspace{-8mm}
\end{figure*}

\textbf{Resource Reductions.} In Figs.~\myblue{\ref{fig:shapenet_resources}}-\myblue{\ref{fig:i3d_resources}}, RANP, marked with (w), achieves much larger FLOPs and memory reductions than vanilla NP, marked with (w/o),
due to the balanced distribution of neuron importance by reweighting.

Specifically, for \textit{ShapeNet}, RANP prunes up to 98.57\% neurons while only up to 78.24\% by vanilla NP in Fig.~\myblue{\ref{fig:shapenet_resources}}.
For \textit{BraTS'18}, RANP can prune up to 96.24\% neurons while only up to 78.17\% neurons can be pruned by vanilla NP in Fig.~\myblue{\ref{fig:brats_resources}}.
For \textit{UCF101}, RANP can prune up to 80.83\% neurons compared to 33.15\% on MobileNetV2 in Fig.~\myblue{\ref{fig:mobilenetv2_resources}}, and 85.3\% neurons compared to 25.32\% on I3D in Fig.~\myblue{\ref{fig:i3d_resources}}.

\textbf{Accuracy with Pruning Sparsity.} For \textit{ShapeNet} in Fig.~\myblue{\ref{fig:shapenet_acc}}, the 23-layer 3D-UNet achieves a higher mIoU than the 15-layer one.
Extremely, when pruned with the maximum neuron sparsity 97.99\%, it can achieve 78.10\% mIoU.
With the maximum neuron sparsity 98.57\%, however, the 15-layer 3D-UNet achieves only 61.42\%.

For \textit{BraTS'18} in Fig.~\myblue{\ref{fig:brats_acc}}, the 23-layer 3D-UNet does not always outperform the 15-layer one and has a larger fluctuation which could be caused by the limited training samples.
Nevertheless, even in the extreme case, the 23-layer 3D-UNet has small accuracy loss.
Clearly, RANP makes it feasible to use deeper 3D-UNets without the memory issue.

For \textit{UCF101} in Figs.~\myblue{\ref{fig:mobilenetv2_acc}}-\myblue{\ref{fig:i3d_acc}}, RANP-f achieves $<$3\% accuracy loss at 70\% neuron sparsity, indicating its effectiveness of greatly reducing resources with small accuracy loss.

\subsection{Transferability with Interactive Model}

\begin{table}[t]
\centering
\caption{Transfer learning by 23-layer 3D-UNets interactively pruned and trained between ShapeNet and BraTS'18.
Accuracy loss from RANP-f to T-RANP-f is negligible.
``T": transferred.}
\label{tb:transfer}
\vspace{-3mm}
\centering
\resizebox{0.48\textwidth}{!}{\begin{tabular}{lcccc}
  \hline
  \multicolumn{1}{c}{\multirow{2}{*}{Manner}} &
  ShapeNet &
  \multicolumn{3}{c}{BraTS'18} \\
  \cline{3-5}
  & mIoU(\%) & ET(\%) & TC(\%) & WT(\%) \\
  \hline
  Full\cite{3dunet}
  & \textbf{\accerror{84.27}{0.21}} & \textbf{\accerror{74.04}{1.45}} & \textbf{\accerror{75.11}{2.43}} & \sunderline{\accerror{84.49}{0.74}} \\
  RANP-f(ours)
  & \sunderline{\accerror{83.86}{0.15}}
  & \accerror{71.13}{1.43}
  & \accerror{72.40}{1.48}
  & \accerror{83.32}{0.62} \\
  T-RANP-f(ours)
  & \accerror{83.25}{0.17}
  & \sunderline{\accerror{72.74}{0.69}}
  & \sunderline{\accerror{73.25}{1.69}}
  & \textbf{\accerror{85.22}{0.57}} \\
  \hline
\end{tabular}}
\vspace{-7mm}
\end{table}

In this experiment, we trained on ShapeNet with a transferred 3D-UNet by RANP on BraTS'18 with 80\% neuron sparsity.
Interactively, with the same neuron sparsity, a transferred 3D-UNet by RANP on ShapeNet was applied to train on BraTS'18.
Results in Table~\myblue{\ref{tb:transfer}} demonstrate that training with transferred models crossing different datasets can largely maintain high or higher accuracy.

\subsection{Lightweight Training on a Single GPU} \label{sec:single_gpu}

\begin{table}[!t]
\centering
\caption{ShapeNet: a deeper 23-layer 3D-UNet is achievable on a single GPU with 80\% neuron pruning.}
\vspace{-2mm}
\resizebox{0.48\textwidth}{!}{
\begin{tabular}{lrrrrc}
  \hline
  \multicolumn{1}{c}{Manner}
  & \multicolumn{1}{c}{Layer}
  & \multicolumn{1}{c}{Batch}
  & \multicolumn{1}{c}{GPU(s)}
  & \multicolumn{1}{c}{Sparsity(\%)}
  & \multicolumn{1}{c}{mIoU(\%)} \\
  \hline
  Full & 15 & 12 & 2 & 0 & \accerror{83.79}{0.21} \\
  Full & 23 & 12 & 2 & 0 & \sunderline{\accerror{84.27}{0.21}} \\
  RANP-f(ours) & 23 & 12 & \textbf{1} & 80 & \textbf{\accerror{84.34}{0.21}} \\
  \hline
\end{tabular}}
\label{tb:single_GPU}
\vspace{-5mm}
\end{table}

RANP with high neuron sparsity makes it feasible to train with large data size on a single GPU due to the largely reduced resources.
We trained on ShapeNet with the same batch size 12 and spatial size $64^3$ in Sec.~\myblue{\ref{sec:setup}} using a 23-layer 3D-UNet with 80\% neuron sparsity on a single GPU.
With this setup, RANP-f reduces $\sim35\times$ GFLOPs (from 259.59 to 7.39) and $\sim3.9\times$ memory (from 1005.96MB to 255.57MB), making it feasible to train on a single GPU instead of 2 GPUs.
It achieves a higher mIoU, 84.34$\pm$0.21\%, than the 15-layer and 23-layer full 3D-UNets in Table~\myblue{\ref{tb:single_GPU}}.

The accuracy increase is due to the enlarged batch size on each GPU.
With limited memory, however, training a 23-layer full 3D-UNet on a single GPU is infeasible.

\subsection{Fast Training with Increased Batch Size} \label{sec:fast_convergence}
\vspace{-1mm}

Here, we used the largest spatial size $128^3$ of one sample on a single GPU and then extended it to RANP with increased batch size from 1 to 4 to fully fill GPU capacity.
The initial learning rate was reduced from 0.1 to 0.01 due to the batch size decreased from 12 in Table~\myblue{\ref{tb:single_GPU}}.
This is to avoid an immediate increase in training loss right after 1st epoch due to the unsuitably large learning space.

In Fig.~\myblue{\ref{fig:convergence_loss}}, RANP-f enables increased batch size 4 and achieves a faster loss convergence than the full network.
In Fig.~\myblue{\ref{fig:convergence_acc}}, the full network executed 6 epochs while RANP-f reached 26 epochs.
Vividly shown by training time in Figs.~\myblue{\ref{fig:convergence_loss_time}} and \myblue{\ref{fig:convergence_acc_time}}, RANP-f has much lower loss and higher accuracy than the full one.
This greatly indicates the practical advantage of RANP on fastening training convergence.

\begin{figure}[!t]
\begin{center}
  \begin{subfigure}[b]{0.113\textwidth}
  \centering
  \includegraphics[width=\textwidth]{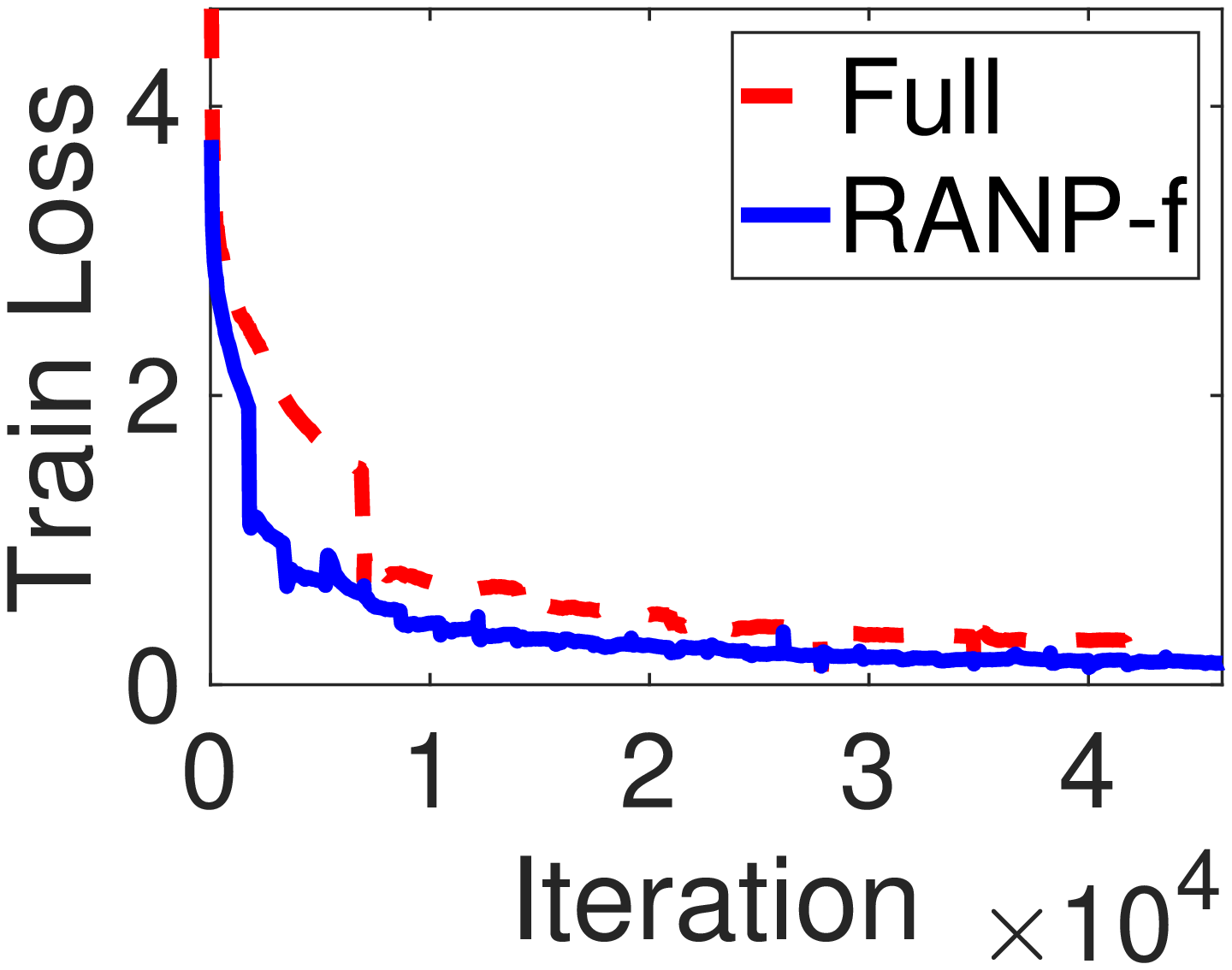}
  \vspace{-\baselineskip}
  \caption{}
  \label{fig:convergence_loss}
  \end{subfigure}
  \begin{subfigure}[b]{0.113\textwidth}
  \centering
  \includegraphics[width=\textwidth]{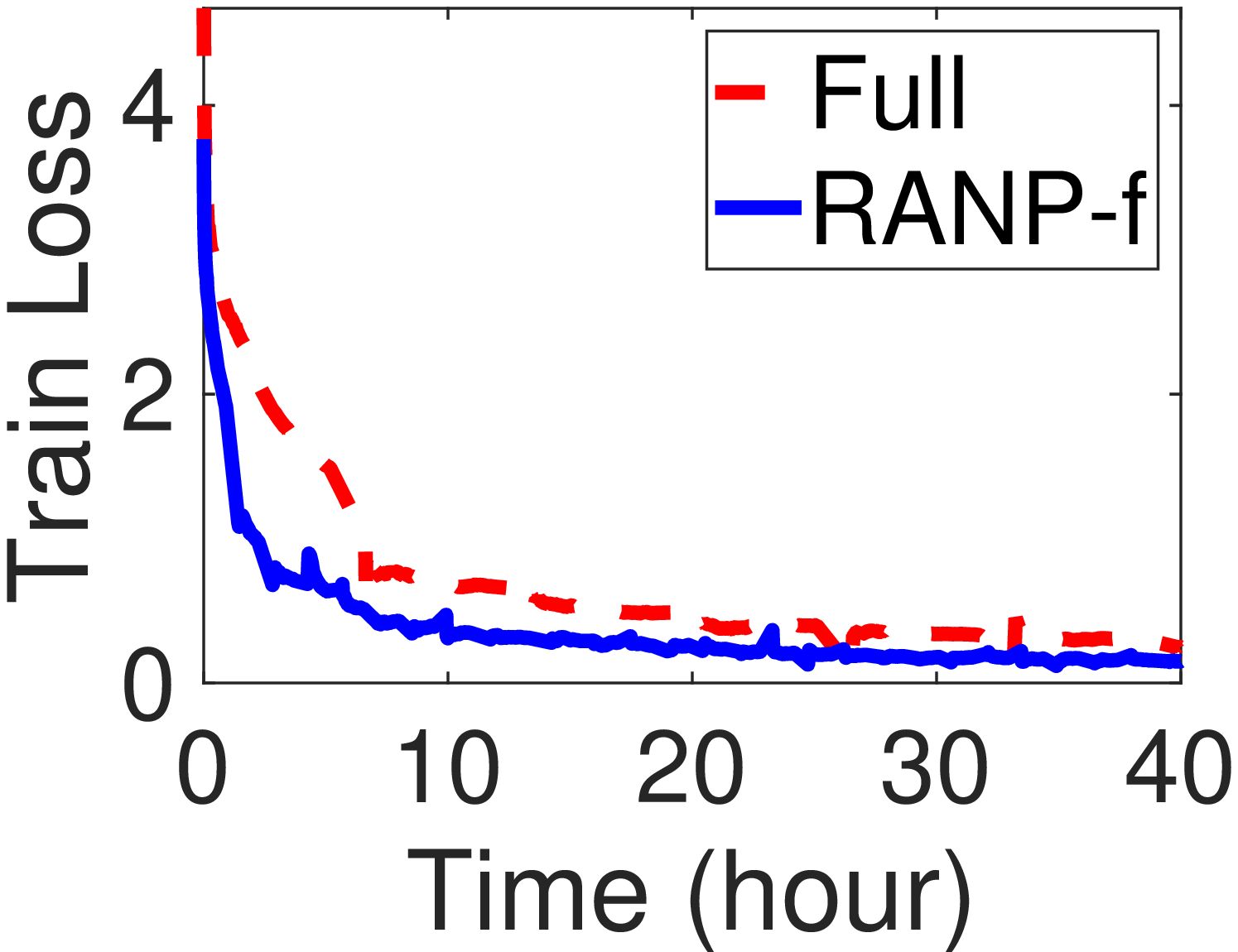}
  \vspace{-\baselineskip}
  \caption{}
  \label{fig:convergence_loss_time}
  \end{subfigure}
  \begin{subfigure}[b]{0.113\textwidth}
  \centering
  \includegraphics[width=\textwidth]{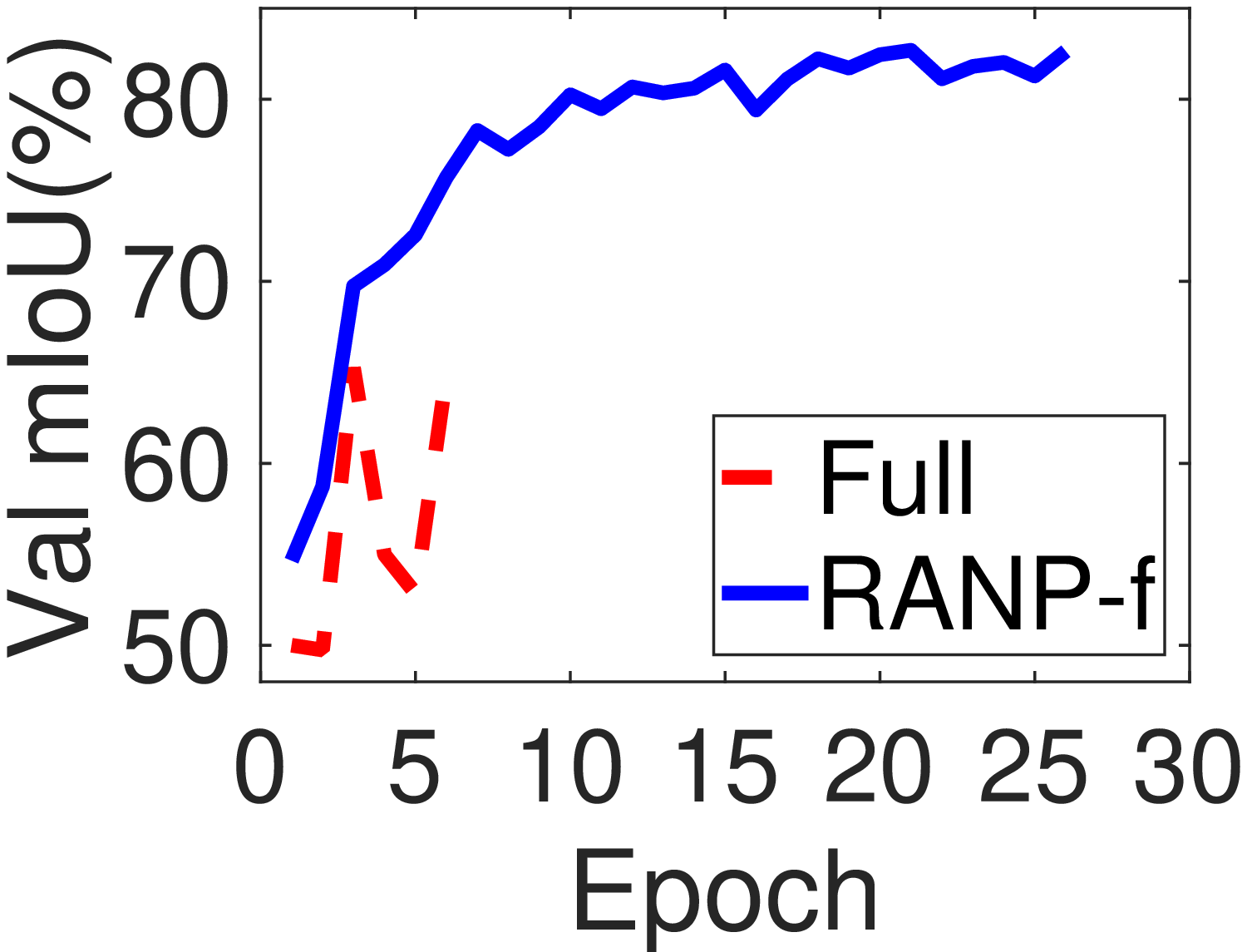}
  \vspace{-\baselineskip}
  \caption{}
  \label{fig:convergence_acc}
  \end{subfigure}
  \begin{subfigure}[b]{0.113\textwidth}
  \centering
  \includegraphics[width=\textwidth]{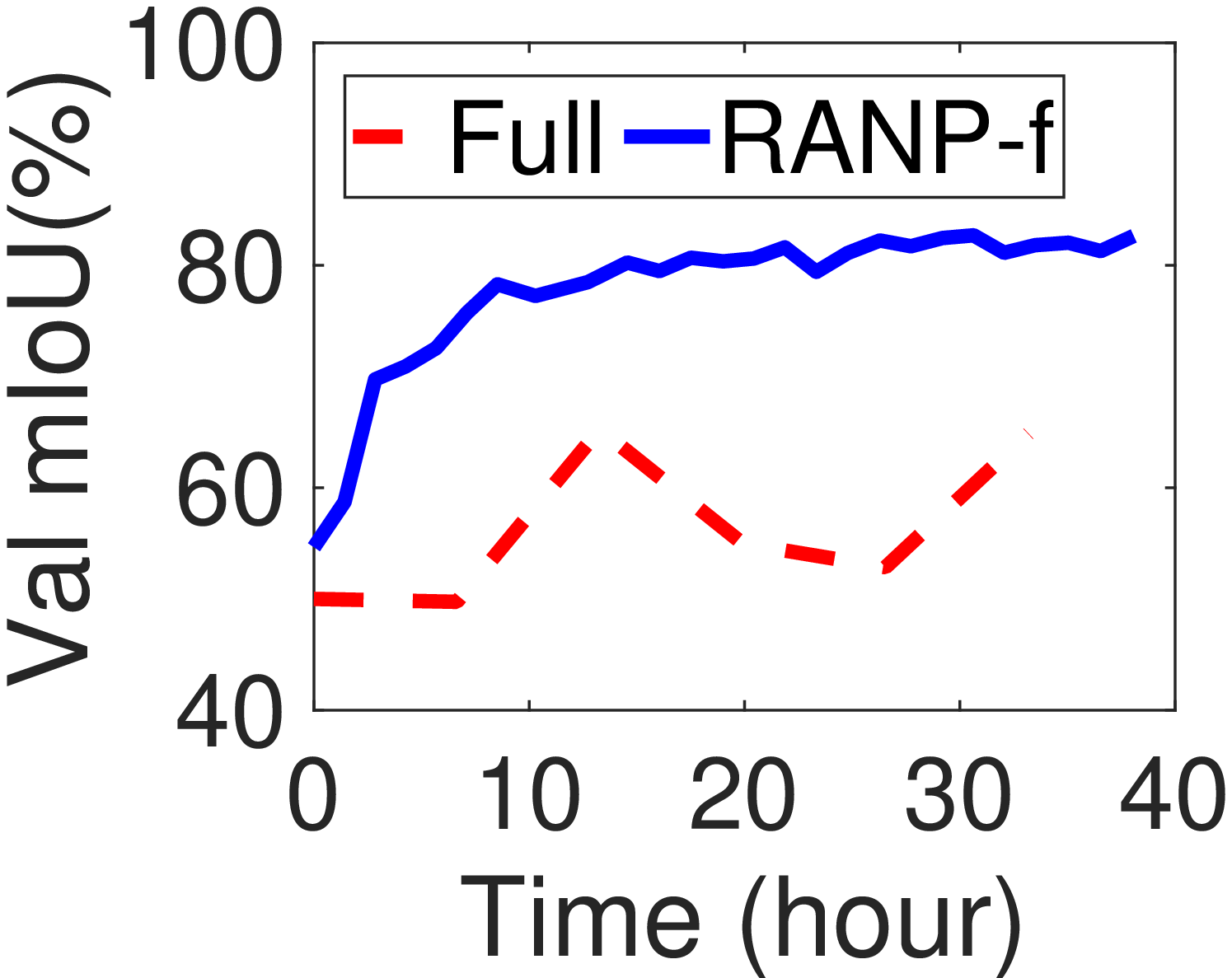}
  \vspace{-\baselineskip}
  \caption{}
  \label{fig:convergence_acc_time}
  \end{subfigure}
\vspace{-1mm}
\caption{ShapeNet: a faster convergence on a single GPU with 23-layer 3D-UNet and increased batch size due to the largely reduced resources by RANP-f.
Batch size is 1 for ``Full" and 4 for ``RANP-f".
Experiments run for 40 hours.}
\label{fig:convergence}
\end{center}
\vspace{-6mm}
\end{figure}

\section{Conclusion}
In this paper, we propose an effective resource aware neuron pruning method, RANP, for 3D CNNs.
RANP prunes a network at initialization by greatly reducing resources with negligible loss of accuracy.
Its resource aware reweighting scheme balances the neuron importance distribution in each layer and enhances the pruning capability of removing a high ratio, say 80\% on 3D-UNet, of neurons with minimal accuracy loss.
This advantage enables training deep 3D CNNs with a large batch size to improve accuracy and achieving lightweight training on one GPU.

Our experiments on 3D semantic segmentation using ShapeNet and BraTS'18 and video classification using UCF101 demonstrate the effectiveness of RANP by pruning 70\%-80\% neurons with minimal loss of accuracy.
Moreover, the transferred models pruned on a dataset and trained on another one are succeeded in maintaining high accuracy, indicating the high transferability of RANP.
Meanwhile, the largely reduced computational resources enable lightweight and fast training on one GPU with increased batch size.

\subsection*{Acknowledgement}
We would like to thank Ondrej Miksik for valuable discussions.
This work is supported by the Australian Centre for Robotic Vision (CE140100016) and Data61, CSIRO.
    \ifarxiv
        \setcounter{equation}{11}
        \setcounter{figure}{5}
        \setcounter{table}{5}
        \appendix
        
    \fi
\fi

{\small
\bibliographystyle{ieee}
\bibliography{main}
}

\end{document}